\documentclass[sigconf]{acmart} % , anonymous, review
\allowdisplaybreaks
\usepackage{algorithm}
\usepackage{algorithmic}
\usepackage{amsmath}
\usepackage{amsthm}
\usepackage{bm}
\usepackage{enumitem}
\usepackage{tabularx}
\usepackage{seqsplit}
\usepackage[bottom]{footmisc}
\newcommand{\hh}[1]{{\small\color{red}{\bf hh: #1}}}
\newcommand{\derek}[1]{{\small\color{teal}{\bf derek: #1}}}

\newcommand{\yd}[1]{{\small\color{orange}{\yd Yada: #1}}}
\newcommand{\vae}{\textsc{PoGeVon}}
\newtheorem*{theorem*}{Theorem}
\newtheorem{theorem}{Theorem}[section]

\newtheorem{definition}[theorem]{Definition}
\newtheorem{lemma}[theorem]{Lemma}
\newtheorem*{lemma*}{Lemma}
\newtheorem{proposition}[theorem]{Proposition}
\newtheorem*{proposition*}{Proposition}
\newtheorem{problem}{Problem}
\theoremstyle{remark}

\newcolumntype{Y}{>{\centering\arraybackslash}X}
\newcolumntype{A}{>{\centering}p{1.0cm}}
\newcolumntype{C}{>{\centering}p{3.5cm}}
\newcolumntype{Z}{>{\centering}p{3cm}}
\floatstyle{ruled}
\newfloat{listing}{tb}{lst}{}
\floatname{listing}{Listing}

\AtBeginDocument{%
  }

\copyrightyear{2023} 
\acmYear{2023} 
\setcopyright{rightsretained} 
\acmConference[KDD '23]{Proceedings of the 29th ACM SIGKDD Conference on Knowledge Discovery and Data Mining}{August 6--10, 2023}{Long Beach, CA, USA}
\acmBooktitle{Proceedings of the 29th ACM SIGKDD Conference on Knowledge Discovery and Data Mining (KDD '23), August 6--10, 2023, Long Beach, CA, USA}
\acmPrice{}
\acmDOI{10.1145/3580305.3599444}
\acmISBN{979-8-4007-0103-0/23/08}

\makeatletter
\gdef\@copyrightpermission{
  \begin{minipage}{0.3\columnwidth}
    \href{https://creativecommons.org/licenses/by/4.0/}{\includegraphics[width=0.90\textwidth]{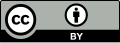}}
  \end{minipage}\hfill
  \begin{minipage}{0.7\columnwidth}
    \href{https://creativecommons.org/licenses/by/4.0/}{This work is licensed under a Creative Commons Attribution International 4.0 License.}
  \end{minipage}
}
\makeatother

\begin{document}

%%
%% The "title" command has an optional parameter,
%% allowing the author to define a "short title" to be used in page headers.
\title{Networked Time Series Imputation via Position-aware Graph Enhanced Variational Autoencoders}

%%
%% The "author" command and its associated commands are used to define
%% the authors and their affiliations.
%% Of note is the shared affiliation of the first two authors, and the
%% "authornote" and "authornotemark" commands
%% used to denote shared contribution to the research.

\author{Dingsu Wang}
\email{dingsuw2@illinois.edu}
\affiliation{%
  \institution{University of Illinois at Urbana-Champaign}
  \state{IL}
  \country{USA}
}
\author{Yuchen Yan}
\email{yucheny5@illinois.edu}
\affiliation{%
  \institution{University of Illinois at Urbana-Champaign}
  \state{IL}
  \country{USA}
}
\author{Ruizhong Qiu}
\email{rq5@illinois.edu}
\affiliation{%
  \institution{University of Illinois at Urbana-Champaign}
  \state{IL}
  \country{USA}
}
\author{Yada Zhu}
\email{yzhu@us.ibm.com}
\affiliation{%
  \institution{IBM Research}
  \state{NY}
  \country{USA}
}
\author{Kaiyu Guan}
\email{kaiyug@illinois.edu}
\affiliation{%
  \institution{University of Illinois at Urbana-Champaign}
  \state{IL}
  \country{USA}
}
\author{Andrew Margenot}
\email{margenot@illinois.edu}
\affiliation{%
  \institution{University of Illinois at Urbana-Champaign}
  \state{IL}
  \country{USA}
}
\author{Hanghang Tong}
\email{htong@illinois.edu}
\affiliation{%
  \institution{University of Illinois at Urbana-Champaign}
  \state{IL}
  \country{USA}
}

%%
%% By default, the full list of authors will be used in the page
%% headers. Often, this list is too long, and will overlap
%% other information printed in the page headers. This command allows
%% the author to define a more concise list
%% of authors' names for this purpose.
\renewcommand{\shortauthors}{Dingsu Wang et al.}

%%
% %% The abstract is a short summary of the work to be presented in the
% %% article.
% \begin{abstract}
%   A clear and well-documented \LaTeX\ document is presented as an
%   article formatted for publication by ACM in a conference proceedings
%   or journal publication. Based on the ``acmart'' document class, this
%   article presents and explains many of the common variations, as well
%   as many of the formatting elements an author may use in the
%   preparation of the documentation of their work.
% \end{abstract}

%%
%% The code below is generated by the tool at http://dl.acm.org/ccs.cfm.
%% Please copy and paste the code instead of the example below.
%%


%%
%% Keywords. The author(s) should pick words that accurately describe
%% the work being presented. Separate the keywords with commas.
%% A "teaser" image appears between the author and affiliation
%% information and the body of the document, and typically spans the
%% page.

\begin{abstract}
% \hh{can we have a better name, e.g., Networked Time Series Imputation based on XXX (XXX is a short phrase to summarize the essence of your method)?}

% \hh{let's think about how to further improve the current version: 1. can we have any additional analysis? 2. improve the writing, in both clarity and intuition. 3. experiments (additional datasets and/or baselines)?}
Multivariate time series (MTS) imputation is a widely studied problem in recent years. Existing methods can be divided into two main groups, including
% \hh{the categorization of these two groups might be a bit problematic: GNN itself belongs to deep models}
(1) deep recurrent or generative models that primarily focus on time series features, %while ignore graph information. 
and (2) graph neural networks (GNNs) based models that utilize the topological information from the inherent graph structure of MTS
% \yc{maybe the following half sentence can be deleted}
as relational inductive bias for imputation. Nevertheless, these methods either neglect topological information or assume the graph structure is fixed and accurately known.
% \yc{break this sentence}
% \hh{and accurately known? -- compared with the existing work such as net3, we handle dynamic and incomplete graph right?}
Thus, they fail to fully utilize the graph dynamics for precise imputation in more challenging MTS data such as \textit{networked time series}
% \hh{networked time series}
(\textit{NTS}), where the underlying graph is constantly changing and might have missing edges. In this paper, we propose a novel approach to overcome these limitations. First, we define the problem of imputation over NTS which contains missing values in both node time series features and graph structures. Then, we design a new model named \vae\
% \hh{how about \textwsc{PoGeVon} -- this can be spelled from the second half of your title, it mimics Pokemon, but with an emphasis on Graph (G) and Variational (V)?}, 
% \yc{need to be simplfied}
which leverages variational autoencoder (VAE) to predict missing values over both node time series features and graph structures. In particular, we propose a new node position embedding 
% \hh{a ... embeddings -- awkward}
based on random walk with restart (RWR) in the encoder with provable higher expressive power compared with message-passing based graph neural networks (GNNs).
% \yc{break this sentence}
We further design a decoder with 3-stage predictions from the perspective of multi-task learning to impute missing values in both time series and graph structures reciprocally.
% \hh{can we add a sentence regarding the novelty/contribution/key idea of our method on the decoder side?}.
Experiment results demonstrate the effectiveness of our model over baselines.
% \derek{Simplify sentences.}
\end{abstract}

%%
%% This command processes the author and affiliation and title
%% information and builds the first part of the formatted document.
\maketitle
\vspace{-0.12cm}
\section{Introduction}
Multivariate time series (MTS) data are common in many real-world applications, such as stock prediction~\cite{ding2015deep,xu2021hist}, traffic forecasting~\cite{li2017diffusion,yu2017spatio,zhang2020spatio} and pandemic analysis~\cite{kapoor2020examining,panagopoulos2021transfer}. However, these data are often incomplete and contain missing values
% \yc{miss some values} (contain certain amount of missing values)
due to reasons such as market close or monitoring sensor/system failure.
% \hh{other reasons? for stock prediction, it is probably not due to sensor failure?}.
Predicting the missing values, which is referred to as the MTS imputation task, plays an important role in these real-world applications.

%Recently, a  large amount of approaches emerge for MTS imputation \cite{fang2020time}. To name a few,.....

% \yc{Put this sentence after briefly introducing the literature or directly delete it because you have the``despite" paragraph. }
Recently, a large amount of approaches emerge for MTS imputation~\cite{fang2020time} in the literature. To name a few, BRITS~\cite{cao2018brits} is built upon
% \hh{upon}
bidirectional recurrent modules and GAIN~\cite{yoon2018gain} is one of the earliest works that use adversarial training for the task. However, many of them ignore the available relational information within the data and thus are less effective to predict missing values compared to those considering both spatial and temporal information. 
% \yc{Start from here.}
In order to tackle this problem, some recent works utilize GNNs or other similar algorithms to assist the imputation over MTS data. GRIN~\cite{cini2021filling} adopts a bidirectional recurrent model based on message passing neural networks~\cite{gilmer2017neural}. They perform a one-step propagation of the hidden representations on the graph to capture the spatial dependencies within the MTS data. SPIN~\cite{marisca2022learning} is a follow-up method which solves the error accumulation problem of GRIN in highly sparse data. It introduces a new attention mechanism to capture spatial-temporal information through inter-node and intra-node attentions. By stacking several attention blocks, the model simulates a diffusion process and can handle data with high missing rates.
% \hh{we should mention Net3 somewhere in this paragraph}
Recently, $\text{NET}^3$~\cite{jing2021network} generalizes the setting and  % steps forward and 
studies tensor time series data in which the underlying graph contains multiple modes. The authors utilize a tensor graph convolution network (GCNs) and a tensor recurrent neural network (RNNs) to handle the tensor graphs and time series respectively.

\begin{figure}[t]
\centering
\includegraphics[width=0.45\textwidth,trim = 2 2 2 2,clip]{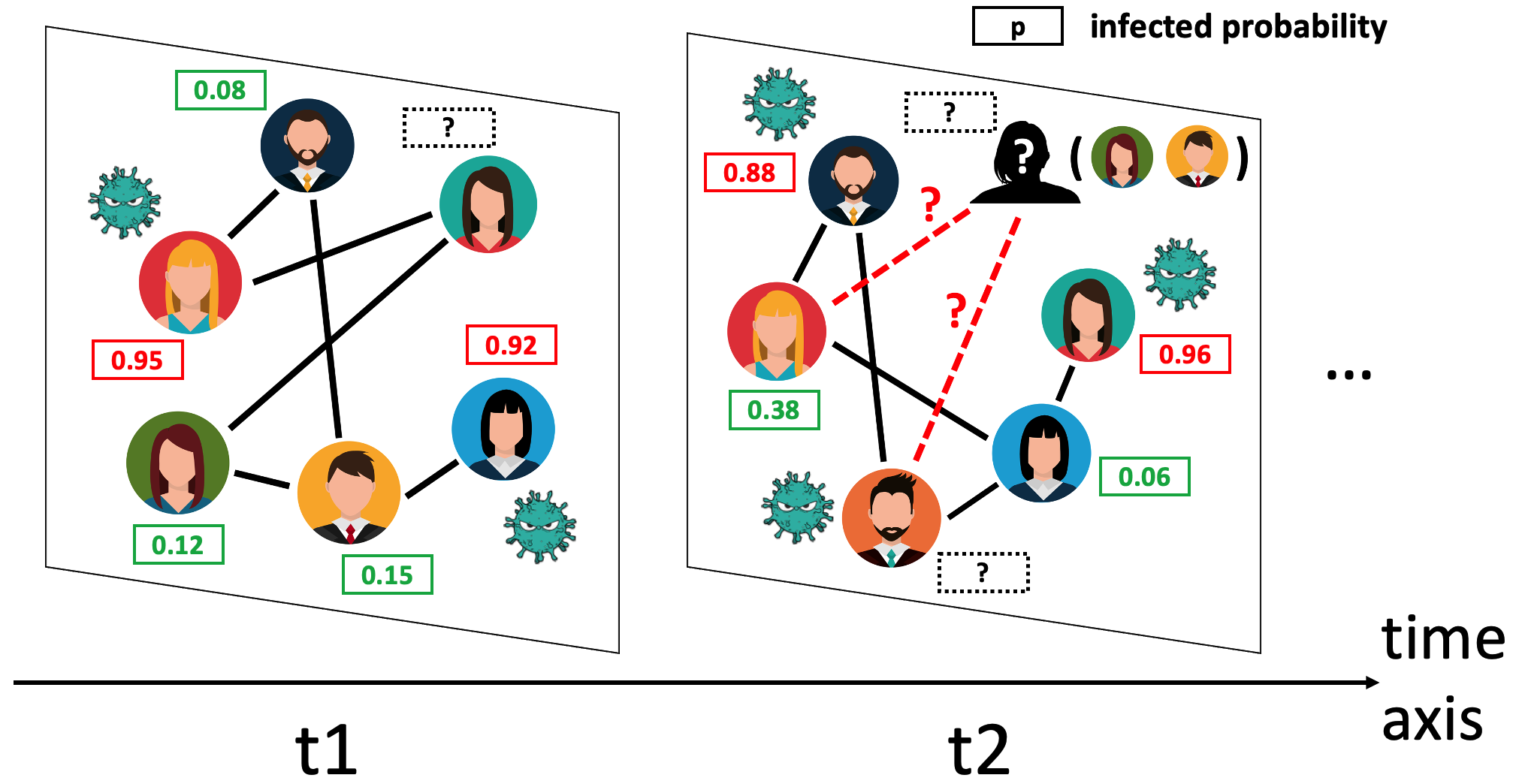} % Reduce the figure size so that it is slightly narrower than the column. Don't use precise values for figure width.This setup will avoid overfull boxes.
\vspace{-3mm}
\caption{An illustrative example of an interaction network during the COVID-19 pandemic where %doctors lose
% \yc{mark small virus}
some patients' infection status might not be available and we have no access to whom these people interact with, which represents a networked time series (NTS) with both missing node features and missing edges.
% \yc{mark the meaning of the value in the box.}
}
% \hh{the time series aspect of the NTS is not shown in the figure. consider to put a time series (to indicate infection status) for each individual, with dashed segment to indicate the missing values. we can still keep the virus symbol aside it.}}
\label{fig:nts}
\end{figure}

Despite the strong empirical performance of these methods on the MTS imputation problem, they rely on the assumption that the underlying graph is fixed and accurately known.
% \hh{what does this mean? better use the same phrase as in abstract: fixed and accurately known}.
However, the graph structure of an MTS may constantly change over time in real-world scenarios. Take epidemiological studies as an example, during the evolution of a pandemic, individuals like human beings or animals may move around and thus the graph that models the spread of disease is dynamic.
%If doctors lose some people's health reports, one could use these individuals' historical data and data from people he interacted with in last few days to make a prediction.
% For example, some streets within a district might be closed for a specific period of time due to construction. This clearly influence the connectivity between different blocks and thus resulting in the changing of traffic flows.
% Taking wireless mobile networks as an example, cell phone users might move around and make contacts with various people in different time periods, which result in a dynamic graphs.
% 
In literature, such time series data %with dynamic graph structures %from HH: for NTS, it is not necessarily that the graph must be dynamic
are referred to as \textit{networked time series}
% \hh{networked?}
(\textit{NTS})\footnote{In some research works~\cite{knight2016modelling}, NTS is also named as \textit{network time series}.}~\cite{jing2021network}.
% \cite{jing2021network}.
Given the nature of NTS data, the missing components can occur in both the node features and the graph structures (See an example in Figure~\ref{fig:nts}), which makes NTS imputation an essentially harder problem compared to MTS imputation.

%This paper will focus on solving the NTS imputation problem. 
In this paper, we first formally define the problem of NTS imputation. We point out that the key challenges of this problem are twofold:
% The key challenges of this problem are that: 
% \hh{shall we point out imputing the graph structure and imputing the time series are mutually beneficial and thus is essentially a multi-task learning problem?}
% \derek{challenge: problems}
(1) The graph that
% \hh{this sentence seems broken. check}
lies behind
% \hh{beind or underlying?}
time series data is evolving constantly, and contains missing edges. Therefore, algorithms should capture the graph dynamics and at the same time be able to restore the lost structures.
% \RQ{It might be better if starting this as a new sentence instead of ``and'', because ``Therefore'' is for the challenge (1)}
(2) The node feature time series also contains missing values, which requires the model to solve a general MTS imputation problem as well.
% missing values in both features and graphs may be correlated with each other. Hence, models should be aware of this hidden relations and impute missing values in a collaborative way.
To address these challenges, 
% \yc{Polish this part after polishing the whole paper.}
we formulate NTS imputation as a \textit{multi-task learning} problem and propose a novel model named \vae\ based on variational autoencoder (VAE)~\cite{kingma2013auto}. Our proposed model consists of two parts, including a recurrent encoder with node position embeddings based on random walk with restart (RWR)~\cite{tong2006fast} and a decoder with 3-stage predictions. The \textit{global} and \textit{local} structural information obtained from RWR with respect to a set of anchor nodes provides useful node representations.
% useful node representations especially when the graph is strongly connected.
Moreover, the 3-stage prediction module in the decoder is designed to impute missing features in time series
% \hh{what are the features here? are they the time series? not clear from the current writing}
and graph structures reciprocally:
% \hh{what is this?}
the first stage prediction fills the missing values for node features and then is used for the imputation over graph structures during the second stage, in return, the predicted graph structures are used in the third stage for node feature imputation. Finally, we replicate
% \hh{replicate?}\derek{maybe duplicate is more accurate? since the architectures are exactly the same} duplicate
the VAE model in \vae\ to handle the bidirectional dynamics in the NTS data.
% \yc{mark}
% Extensive experiments are conducted on various real-world time series datasetes in the setting of NTS imputation. Experimental results demonstrate that our model can obtain higher imputation accuracy in all these datasets.
% , especially in the context of predicting missing values in the future while learning from historical data
The main contributions of this paper can be summarized as:\vspace{-1mm}
\begin{itemize}[leftmargin=8mm]
    \item \textbf{Problem Definition}. To our best knowledge, we are the first to study the joint problem of MTS
    % multivariate time series
    imputation and graph imputation over networked time series data.
    \item \textbf{Novel Algorithm and Analysis.} We propose a novel imputation model based on VAE, which consists of an encoder with RWR-based node position embeddings, and a decoder with 3-stage predictions. We provide theoretical analysis of the expressive power of our position embeddings compared with message-passing based temporal GNNs, as well as the benefit of multi-task learning approach for NTS imputation problem from the perspective of information bottleneck.
    % \hh{maybe update this based on the new analysis}
    % \hh{analysis asepct is not mentioned here}
    \item \textbf{Empirical Evaluations.} We demonstrate the effectiveness of our method by outperforming powerful baselines for both MTS imputation and link prediction tasks on various real-world datasets.
\end{itemize}
% \hh{a short paragraph about the organization of the paper}
\vspace{-1mm}

The rest of the paper is organized as follows. Section~\ref{sec:preliminary} defines the imputation problem over NTS data. Section~\ref{sec:methodology} presents the proposed \vae\ model. Section~\ref{sec:experiment} shows the experiment results. Related works and conclusions are given in Section~\ref{sec:related} and Section~\ref{sec:conclusion} respectively.
\vspace{-0.12cm}
\section{Problem Definition}\label{sec:preliminary}

\begin{table}[!htb]
\setlength{\abovecaptionskip}{3pt}
\caption{Symbols and Notations.}
\label{tb:notation}
\begin{tabular}{c|c}
\toprule
\textbf{Symbol}& \textbf{Definition} \\
\midrule
$\mathcal{G}$ & sequence of graphs \\
$\mathcal{A}$ & tensor of graph adjacency sequence \\
$\mathcal{X}$ & tensor of multivariate time series \\
$\mathcal{M}$ & mask tensor of $\mathcal{X}$ \\
$\mathcal{R}$ & tensor of node position embeddings \\
$G_t$ & graph at time step $t$ \\
$\Tilde{G}_t$ & observed graph at time step $t$ \\
$\Tilde{\mathcal{G}}$ & observed sequence of graphs \\
$\Tilde{\mathcal{A}}$ & observed tensor of graph adjacency sequence \\
$\Tilde{\mathcal{X}}$ & observed multivariate time series \\
\midrule
$\mathbf{A}_t$ & adjacency matrix at time $t$ \\
$\mathbf{X}_t$ & node feature matrix at time $t$ \\
$\mathbf{M}_t$ & mask matrix at time $t$ \\
$\mathbf{R}_t$ & RWR position matrix at time $t$ \\
$\mathbf{r}_{t,i}$ & RWR position score of node $i$ at time $t$ \\
$\mathbf{e}_i$ & one-hot restart vector with value 1 at index $i$ \\
$\mathbf{D} = \text{diag}(\mathbf{d})$ & diagonal matrix of the degree vector $\mathbf{d}$ \\
$\mathbf{A}^\top$ & transpose of $\mathbf{A}$ \\
$\mathbf{Z}$ & latent node embedding matrix of VAE \\
\midrule
$H(X)$ & entropy of random variable $X$ \\
$I(X;Y)$ & mutual information between $X$ and $Y$ \\
$T$ & length of time series \\
$N$ & number of nodes \\
$D$ & number of features \\
$i,j,u,v$ & indices of nodes \\
$c$ & restart probability in RWR \\
$z$ & latent representations of VAE \\
$\theta, \gamma, \phi$ & parameters of neural networks \\
$\| \cdot \|_F$ & Frobenius norm \\
$\odot$ & Hadamard product \\
\bottomrule
\end{tabular}
\end{table}

Table~\ref{tb:notation} lists main symbols and notations used throughout this paper. 
Calligraphic letters denote tensors or graphs (e.g., $\mathcal{X}$, $\mathcal{G}$), bold uppercase letters are used for matrices (e.g., $\mathbf{A}$), bold lowercase letters are for vectors (e.g., $\mathbf{v}$). Uppercase letters (e.g., $T$) are used for scalars, and lowercase letters (e.g., $i$) are for indices. For matrices, we use $\mathbf{A}[i,j]$ to denote %the indexing and gets 
the value at the $i$-th row and $j$-th column.

% \hh{a few sentences on the name convention we used}

% \RQ{The definitions here seem a bit abrupt. It might be better if you add a few transition sentences between naming conventions and definitions to brief why you introduce these definitions.}

We first present some necessary preliminaries and then introduce the networked time series imputation problem in this section.
\begin{definition}[\textbf{Multivariate Time Series (MTS)}]
A multivariate time series $\mathcal{X} \in \mathbb{R}^{T \times N \times D}$ is a sequence of observations: $\{ \mathbf{X}_1, \mathbf{X}_2, ..., \mathbf{X}_T \}$, where each observation $\mathbf{X}_t \in \mathbb{R}^{N \times D}$ is a slice of $\mathcal{X}$ at time step $t$ that contains $N$ entities with $D$ features.
\end{definition}

\begin{definition}[\textbf{Networked Time Series (NTS)}]
Networked time series is an extension of multivariate time series, in which a sequence of graphs $\mathcal{G}(\mathcal{A}, \mathcal{X})=\{ G_1, G_2, ..., G_T \}$ is given, and $\mathcal{A}$ models the node interactions as time goes by.
% \yc{mark}
Each graph $G_t$ is represented as a weighted adjacency matrix $\mathbf{A}_t \in \mathbb{R}^{N \times N}$ with the node feature matrix $\mathbf{X}_t \in \mathbb{R}^{N \times D}$.
% \hh{can we make it clear that $\mathcal{X}$ is the MTS in def3.1, and $X_t$ is a slice of $\mathcal{X$}?}
\end{definition}

\begin{definition}[\textbf{Mask Tensor}]\label{def:mask}
A binary mask tensor $\mathcal{M}: \{ \mathbf{M}_1, \mathbf{M}_2, ..., \mathbf{M}_T \} \in \mathbb{R}^{T \times N \times D}$ serves as the indicator
% \hh{indicator?}
of missing values in MTS data, in which the value $\mathbf{M}_t[i,j]$ indicates the availability of each feature $j$ of entity $i$ at time step $t$: $\mathbf{M}_t[i,j]$ being 0 or 1 indicates the corresponding feature is missing or observed.
\end{definition}

% \hh{we should also mention the mask tensor for graph adjacency -- we can either mention it in def 2.3 or after def 2.3 (e.g., likewise, we can define a mask tensor for graph adjacency tentor, blabla}

% \yc{mark}
Given the nature of NTS data, its missing data can occur in two parts: (1) missing values in node feature time series, and (2) missing edges in graph structures. The former is similar to missing values in traditional MTS, while the latter is unique in NTS which demonstrates the underlying dynamics of a graph sequence. Therefore, we can also define mask tensor for graph adjacency sequence similar to Definition \ref{def:mask}. We formally define the partially observed NTS data and NTS imputation problem as follows:

\begin{definition}[\textbf{Partially Observed NTS}]
A partially observed NTS: $\mathcal{G}(\Tilde{\mathcal{A}}, \Tilde{\mathcal{X}})=\{ \Tilde{G}_1, \Tilde{G}_2, ..., \Tilde{G}_T \}$  consists of observed graph adjacency tensor $\Tilde{\mathcal{A}}$ and observed node feature tensor $\Tilde{\mathcal{X}}$. The value of $\Tilde{\mathbf{A}}_t[i,j]$ and $\Tilde{\mathbf{X}}_t[i,j]$ can be observed only if $\mathbf{M}^A_t[i,j]=1$ and $\mathbf{M}^X_t[i,j]=1$ where $\mathcal{M}^A$ and $\mathcal{M}^X$ are the mask tensors for graph adjacency structure and node features respectively.
% where $\Tilde{\mathcal{A}} = \mathcal{A} \odot \mathcal{M}_A$ and $\Tilde{\mathcal{X}} = \mathcal{X} \odot \mathcal{M}_X$.
\end{definition}

\begin{problem}[\textbf{NTS Imputation}]\label{def:nts-impute}
\begin{description}
   \item 
   \item[Given:] A partially observed NTS with graph sequence $\mathcal{G}(\Tilde{\mathcal{A}}, \Tilde{\mathcal{X}})=\{ \Tilde{G}_1, \Tilde{G}_2, ..., \Tilde{G}_T \}$;
   % in which $\Tilde{\mathcal{A}}$ is the observed tensor of the graph adjacency sequence, and $\Tilde{\mathcal{X}}$ is the observed feature tensor by applying mask tensor $\mathcal{M}$ over $\mathcal{X}$
   % \hh{X is what we want to output. why is it in 'input'? maybe we should first define 'partially observed NTS with the help of mask tensor in defition 2.4. then we give a rigorous defintion of nts imputation}.
   \item[Output:] The predicted graph adjacency tensor $\mathcal{A}$ and the tensor $\mathcal{X}$ of node feature time series.
\end{description}
\end{problem}

{\em Note.} For clarity, we use node features and node time series interchangeably, and same for the graph adjacency imputations and missing edges/links predictions.
\vspace{-0.12cm}
\section{Methodology}\label{sec:methodology}

% \RQ{It is a bit confusing to me why your method is called \vae{}. It might be better if you mention the full name here or in introduction.}

\begin{figure*}[ht]
\centering
\includegraphics[width=1.0\textwidth,trim = 2 2 2 2,clip]{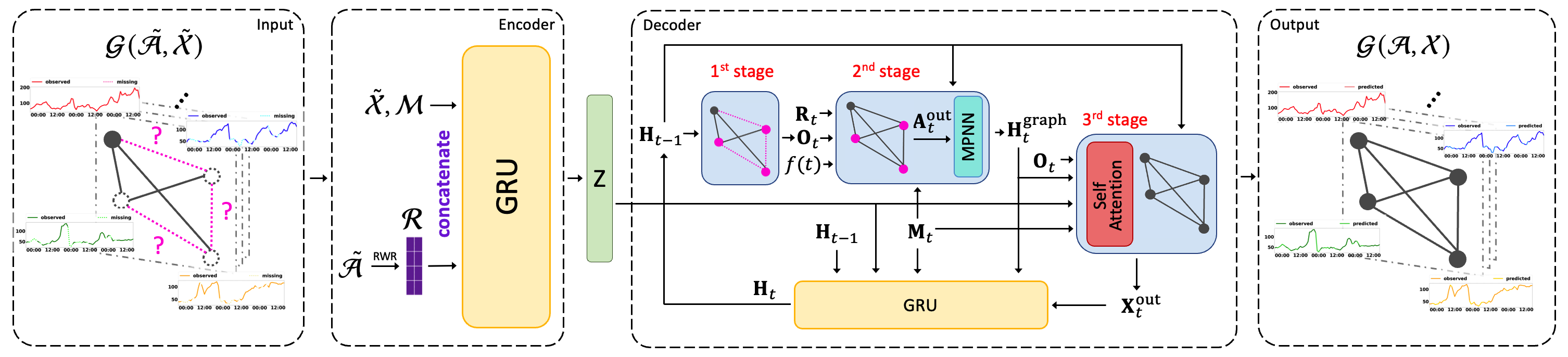}
\vspace{-7mm}
\caption{The model architecture of the proposed \vae.
% \hh{Derek: let's discuss this in our next in-person. i think we might further improve this figure. here are a few quick points: 1. can we re-design the layout so that it has four boxes: input, encoder, decoder and output. 2. for the icon of NTS, let's put a simple time series (like what you have in figure 1 or figure 2) beside each node, where the dashed part represents the missing node feature. as we move from the left/input to the right/output, the dashed part gradually shrink. 3. how do we initialize $H_{t-1}$? 4. how to indicate that we feed/update $H_t$ to $H_{t-1}$? 5. i still feel that generating R should be part of encoder}
}
% \yc{concate, mark each component, the current framework is not clear.}}
\label{fig:model.}
\vspace{-3mm}
\end{figure*}

In this section, we introduce our model named \underline{Po}sition-aware \underline{G}raph \underline{E}nhanced \underline{V}ariational Aut\underline{o}e\underline{n}coders (\vae) in detail. In order to predict the missing values in both the node features
% \hh{node feature, node time series: let's statement somewhere that we use them interchangeably. same for graph adjacency imputation, missing edge, link prediction}
and the graph structures, we design a novel \textit{variational autoencoder} (\textit{VAE}), whose detailed architecture is shown in Figure~\ref{fig:model.}. It consists of an encoder with node position embeddings based on \textit{random walk with restart} (\textit{RWR}), and a decoder with 3-stage predictions. 
% \yc{3-stage imputations or 3-stage predictions, be consistent with the introduction.}
We then replicate the VAE to handle bidirectional dynamics.
% (simplify)
% Specifically\yc{mark, followed by...}, we use two VAEs to handle sequential data in both the forward and backward directions similar to \cite{cao2018brits,cini2021filling}. 
We start in Subsection~\ref{sec:multi-task} to discuss the multi-task learning setting of NTS imputation problem, analyze its mutual benefit and implications to the encoder/decoder design. Then, we present the details of the proposed encoder (Subsection~\ref{sec:encoder}) and decoder (Subsection~\ref{sec:decoder}), followed by the training objective in Subsection~\ref{sec:objective} as well as the compelxity analysis in Subsection~\ref{sec:complexity}. 
%as well as its benefit, we will analyze the expressive power of our specially designed node position embeddings and present the whole \vae\ model in the following subsections.
% \yc{We start from introducing the encoder part and analyzing the expressiveness of our specially designed node position embeddings to present the whole XXXX framework...}

% \hh{we call sec3.2 as the framework, but this subsection is mostly for decoder right? maybe we can consider the following structure: 3.1: overall framework. briefly introduce our encoder-decoder framework (e.g., first paragraph of sec 3.2 plus figure 3). 3.2: encoder. 3.3: decoder. 3.4: objective and training.}

\begin{figure}[ht]
\includegraphics[width=0.48\textwidth,trim = 2 2 2 2,clip]{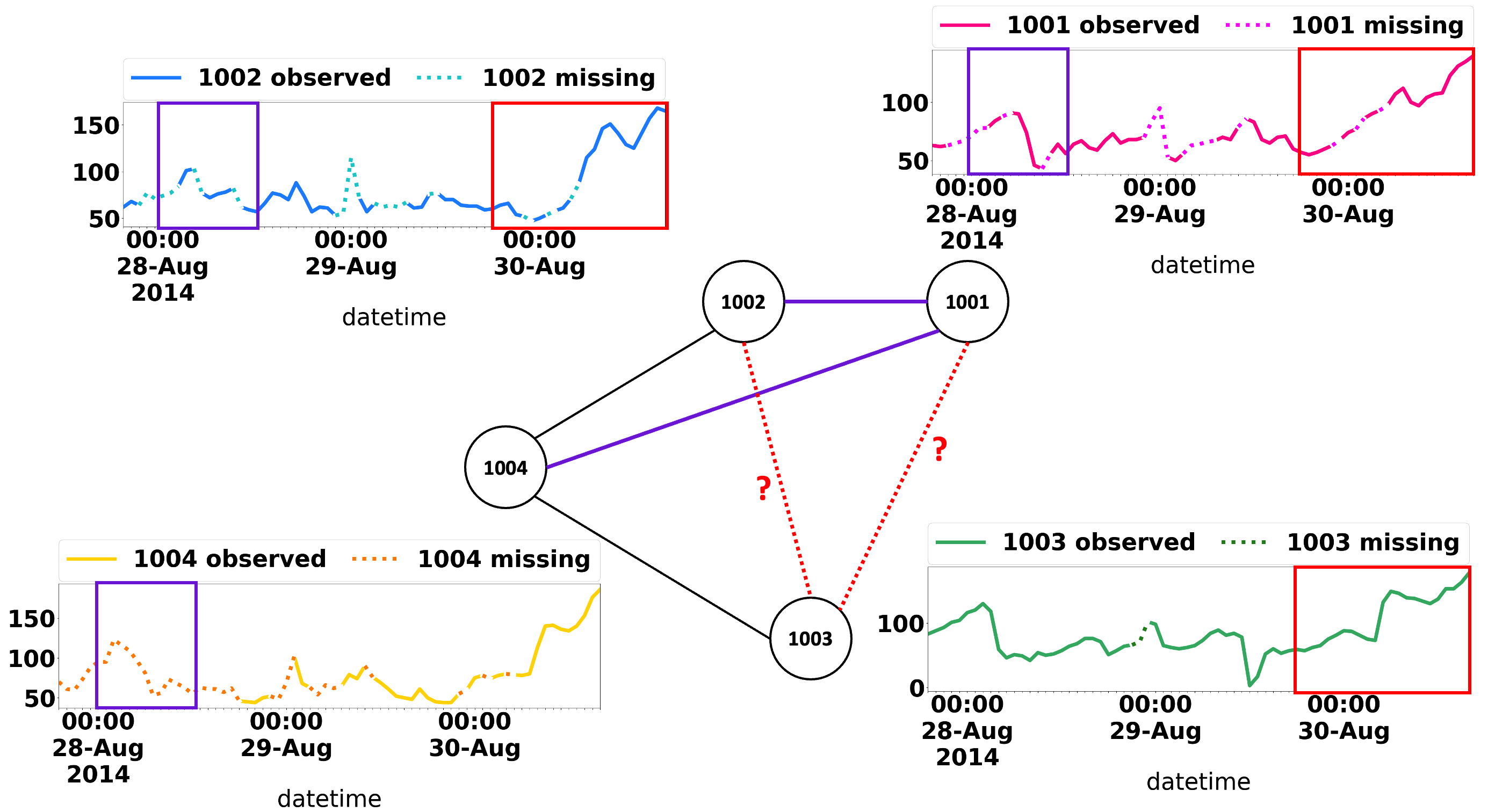}
\setlength{\belowcaptionskip}{-1pt}
\vspace{-0.7cm}
\caption{An illustrative example of mutual reinforcing effect between node feature imputation and graph structure imputation, based on 4 monitor stations in AQ36 dataset (See Section~\ref{sec:experiment} for the details of the dataset). Correlation between three time series (1001, 1002, and 1003, indicated by three red boxes) helps impute the missing edges between them (the two red dashed lines). Meanwhile, the edges between 1001, 1002 and 1004 (the two purple lines) helps impute time series/node features by capturing the lagged correlation between them (the three purple boxes). Best viewed in color.}
% \hh{Derek: 1. pls double check if my revisions are accurate. 2. if we use dotted lines to indicate missing values in time series, will that be more clear?}}
%, to capture the lagged  denote the correlated time series information that can help to establish missing edges, while purple boxes denote the information of trends in time series features from observed adjacent nodes that can be applied for missing feature predictions.}
\label{fig:ntsexp}
\end{figure}

\vspace{-1.5mm}
\subsection{Multi-Task Learning Framework}\label{sec:multi-task}
% \yc{Move all following contents to Section 3.}
% In this paper, we will focus on the problem of NTS imputation, 
% and especially, we assume the mask $\mathcal{M}$ and $\mathcal{M_A}$ are strongly correlated with each other\yc{not reasonable, this assumption is too strong.}: missing values and missing edges occur within the same set of nodes at each time step.
% and b
Because of the potential mutual benefit of predicting missing node features and edges, it is natural to formulate the NTS imputation as a multi-task learning problem which consists of the \textit{imputation}
% \textit{reconstruction}\hh{shall we use 'imputation' here? since later (right above eq2), we use 'reconstruction' to refer the imputation of both features and links}
task for node time series and the \textit{link prediction}
% \hh{is this something we can conduct any theoretic analysis, e.g., under certain condition, imputation of graph topology helps imputation of node feature and via versa? in addition, if we can have a toy example to illustrate such mutual reinforcing effect, that will help improve the intuition of the paper}
% \derek{graph imputation?}
% \yc{why supervised}
task for graph
% \yc{structures}
structures.
% \hh{1. we might consider to move this to section 3 (since we have not introduced encoder-decoder based solution for problem 1 yet. 2. we might also consider to turn the following into a theorem. but before that, there are a number of places that do not sound clear to me.}
Let us analyze the benefit of modeling NTS imputation as a multi-task learning problem from the perspective of \textit{information bottleneck} in unsupervised representation learning~\cite{tishby2000information,alemi2016deep}, and formulate the objective of NTS imputation as:
\begin{align}\label{IB}
    \max[I(\Tilde{\mathcal{A}}, \Tilde{\mathcal{X}}; z) - \beta I(z; \Tilde{\mathcal{G}}_{t:t+\Delta t})]
\end{align}
where $z$
% \hh{is the z the output of your encoder?} \derek{Yes, but in VAE, it usually represents the general class of the latent embeddings output by the encoder.}
is the latent representation
% \yc{what is bottleneck representation, add an explanation sentence.}
, $I(\cdot; \cdot)$
% \hh{the difference between $I(\cdot, \cdot)$ and $I(\cdot; \cdot)$}
is the mutual information, $\Tilde{\mathcal{G}}_{t:t+\Delta t}$ is the data sample which represents a sliding window of NTS data and $\beta$ is the Lagrange multiplier. This formulation closely relates to the objective of a $\beta$-VAE~\cite{higgins2016beta, % } as proved in~\cite{
alemi2016deep}.
% \hh{overall, there are quite a few very long sentences in this section. try to break each of them into a few shorter ones}
Here, the second term $\beta I(z; \Tilde{\mathcal{G}}_{t:t+\Delta t})$
% \yc{add concrete mathematical formula. same for the first term to make it clear.}
in Eq.~\eqref{IB}
% \hh{can we use eqref for equations? this will automatically add parathethis, and equations numbers will not be messed up with the table/figure numbers}
constraints the amount of identity information of each data sample that can transmit through the latent representation $z$.
% with the help of 
In $\beta$-VAE, this is upper bounded by minimizing the Kullback–Leibler divergence $\beta\cdot\mathbb{KL}[q_{\theta}(z|X)||p(z)]$~\cite{burgess2018understanding}.
% \RQ{This seems the first time where the abbreviation ``KL'' appears, so it might be clearer if you state the full name here.}
The first term $I(\Tilde{\mathcal{A}}, \Tilde{\mathcal{X}}; z)$ in Eq.~\eqref{IB} represents the reconstruction task of VAE which can be decomposed as~\cite{jing2021hdmi}:
% \hh{this kind of analysis is very helpful. let's elaborate this and make sure our statements are rigorous}
\begin{align}\label{II}
    I(\Tilde{\mathcal{A}}, \Tilde{\mathcal{X}}; z) = I(\Tilde{\mathcal{A}}; z) + I(\Tilde{\mathcal{X}}; z) - I(\Tilde{\mathcal{A}}; \Tilde{\mathcal{X}}; z)
\end{align}
where $I(\Tilde{\mathcal{A}}, \Tilde{\mathcal{X}}; z)$ represents the mutual information between the partially observed NTS $\Tilde{\mathcal{G}}$ (i.e., the joint distribution of $\Tilde{\mathcal{A}}$ and $\Tilde{\mathcal{X}}$) and $z$,
% \hh{i re-wrote it a bit. check if it is accurate}, 
% \hh{difference between $I(\Tilde{\mathcal{A}}, \Tilde{\mathcal{X}}; z)$ and $I(\Tilde{\mathcal{A}}; \Tilde{\mathcal{X}}; z)$}
while $I(\Tilde{\mathcal{A}}; \Tilde{\mathcal{X}}; z)$ is the High-order Mutual Information~\cite{mcgill1954multivariate,jing2021hdmi}, 
% \hh{is this also called high-order mi? see Baoyu's HDMI paper (www21) and the reference therein -- this might be an interesting angle to analyze MTL}
which measures the shared information among multiple different random variables (i.e., $\Tilde{\mathcal{A}}$, $\Tilde{\mathcal{X}}$, and $z$). It is worthy noting that when $\Tilde{\mathcal{A}}$ and $\Tilde{\mathcal{X}}$ are independent from each other (even given $z$), we have:
% \hh{not clear why this ($I(\Tilde{\mathcal{A}}; \Tilde{\mathcal{X}}; z) = 0$) leads to Eq3. what is the relationship between Eq2 and Eq3?}
\begin{align}\label{eq:uncorr}
\begin{split}
    & I(\Tilde{\mathcal{A}}, \Tilde{\mathcal{X}}; z) = H(\Tilde{\mathcal{A}}, \Tilde{\mathcal{X}}) - H(\Tilde{\mathcal{A}}, \Tilde{\mathcal{X}} | z) \\
    & = H(\Tilde{\mathcal{A}}) + H(\Tilde{\mathcal{X}}) - H(\Tilde{\mathcal{A}}|z) - H(\Tilde{\mathcal{X}}|z) = I(\Tilde{\mathcal{A}};z) + I(\Tilde{\mathcal{X}};z)
\end{split}
\end{align}
where $H(\cdot)$ is the entropy. Compared with Eq.~\eqref{II}, it is clear that $I(\Tilde{\mathcal{A}}; \Tilde{\mathcal{X}}; z)$ now equals to $0$. Under such circumstances, i.e., no correlation exists between features of any adjacent node pairs, the objective in Eq.~\eqref{IB} becomes modeling time series features and graph structures independently.
% similar to the setting of \cite{du2022disentangled,qian2020multi} in which the algorithms do not capture such correlation explicitly.
% \hh{why does MTL does not consider the correlations? i would imagine that MTL wants to leverage such correlation to boost the performance of different tasks?}
% \derek{Indeed, MTL should capture such correlation, I was just citing two papers in the field that directly predict labels/output for different tasks but not utilize the prediction for each other explicitly. I changed some of the terms/analysis and hope this is more clear.}
 However, in reality, this is often not the case. Figure~\ref{fig:ntsexp} demonstrates an illustrative example from the AQ36 dataset~\cite{zheng2015forecasting} in which NTS imputation problem occurs when monitor stations fail due to system errors and lose data as well as connections with each other. 
To maximize Eq.~\eqref{II}, we further decompose $I(\Tilde{\mathcal{A}}; \Tilde{\mathcal{X}}; z)$:
\begin{align}\label{HOMI}
I(\Tilde{\mathcal{A}}; \Tilde{\mathcal{X}}; z) & = I(\Tilde{\mathcal{A}}; z) - I(\Tilde{\mathcal{A}}; z | \Tilde{\mathcal{X}}) = I(\Tilde{\mathcal{X}}; z) - I(\Tilde{\mathcal{X}}; z | \Tilde{\mathcal{A}})
\end{align}
where the second equation holds because of symmetry~\cite{yeung1991new}. Combining Eq.~\eqref{II} and Eq.~\eqref{HOMI}, we can derive the objective term for the decoder as: 
\begin{align}\label{objective}
    % & I(\Tilde{\mathcal{A}}, \Tilde{\mathcal{X}}; z) = \underbrace{I(\Tilde{\mathcal{A}}; z)}_{\text{VAE decoder for $\Tilde{\mathcal{A}}$}} + \underbrace{I(\Tilde{\mathcal{X}}; z | \Tilde{\mathcal{A}})}_{\text{CVAE decoder for $\Tilde{\mathcal{X}}$}} \\
    & 2 \cdot I(\Tilde{\mathcal{A}}, \Tilde{\mathcal{X}}; z) = \underbrace{I(\Tilde{\mathcal{A}}; z) + I(\Tilde{\mathcal{X}}; z)}_{\text{VAE}} + \underbrace{I(\Tilde{\mathcal{X}}; z | \Tilde{\mathcal{A}}) + I(\Tilde{\mathcal{A}}; z | \Tilde{\mathcal{X}})}_{\text{Conditional VAE}}
    % 2 \cdot I(\Tilde{\mathcal{A}}, \Tilde{\mathcal{X}}; z) & = I(\Tilde{\mathcal{A}}; z) + I(\Tilde{\mathcal{X}}; z) \to \text{VAE decoder} \\
    % & + I(\Tilde{\mathcal{X}}; z | \Tilde{\mathcal{A}}) + I(\Tilde{\mathcal{A}}; z | \Tilde{\mathcal{X}}) \to\text{Conditional VAE decoder}
    % & 2 \cdot I(\Tilde{\mathcal{A}}, \Tilde{\mathcal{X}}; z) = 2 \cdot I(\Tilde{\mathcal{A}}; z) + 2 \cdot I(\Tilde{\mathcal{X}}; z) - 2 \cdot I(\Tilde{\mathcal{A}}; \Tilde{\mathcal{X}}; z) \\
    % & = I(\Tilde{\mathcal{A}}; z) + I(\Tilde{\mathcal{X}}; z) + I(\Tilde{\mathcal{A}}; z | \Tilde{\mathcal{X}}) + I(\Tilde{\mathcal{X}}; z | \Tilde{\mathcal{A}})\\
\end{align}
where the first two terms can be bounded by the objective for VAE decoder as in~\cite{alemi2016deep}. The last two terms represent the objective of conditional VAE (CVAE) since $I(\Tilde{\mathcal{X}}; z | \Tilde{\mathcal{A}}) = H(\Tilde{\mathcal{X}}|\Tilde{\mathcal{A}}) - H(\Tilde{\mathcal{X}}|\Tilde{\mathcal{A}}, z)$. The first term $H(\Tilde{\mathcal{X}}|\Tilde{\mathcal{A}})$ on the right hand can be dropped because it is independent from our model, and maximizing the second term $- H(\Tilde{\mathcal{X}}|\Tilde{\mathcal{A}}, z)$ is essentially the same as optimizing the decoder of CVAE with objective $\max p(\Tilde{\mathcal{X}}|\Tilde{\mathcal{A}})$. Similar analysis applies to $I(\Tilde{\mathcal{A}}; z | \Tilde{\mathcal{X}})$.
% \derek{Simplify this sentence.}
% \yc{delete the following part of this sentence.}similar to the derivation in \cite{alemi2016deep}.
Eq.~\eqref{objective} provies an important insight: we can use $\Tilde{\mathcal{A}}$ and $\Tilde{\mathcal{X}}$ as the conditions for each other's
% \hh{each other's}
predictions since imputation over one of them might be instructive for the other. % which denotes the different stages in our decoder. 
% \hh{let's add a short paragraph about (1) the main conclusion of our analysis, (2) what is the implications of such conclusion to the design of encoder and decoder. check if the following text makes sense and accurate.}

To summarize, our analysis reveals that (1) when the features of adjacent nodes are uncorrelated, we can impute the node time series and graph adjacency independently (Eq.~\eqref{eq:uncorr}); however, (2) in real applications, node features and graph structure are often correlated (e.g., Figure~\ref{fig:ntsexp}), and in such a scenario, there might be a mutual reinforcing effect between node feature imputation and graph adjacency imputation (Eq.~\eqref{HOMI}). Our analysis also provides novel and critical clues that can guide the design of the encoder-decoder framework for learning datasets with multi-modality such as NTS. For the encoder, Eq.~\eqref{objective} suggests that the latent representation $z$ (i.e., the output of the encoder) should encode both the graph adjacency information and the node feature information (i.e., the VAE part of Eq.~\eqref{objective}) as well as the mutual interaction between them (i.e., the CVAE part of Eq.~\eqref{objective}). For the decoder, we will present a three-stage prediction method so that the (imputed) graph structures and the (imputed) node features can be used as each other's condition respectively (i.e., the CVAE part of Eq.~\eqref{objective}).
 % This leads to the $\text{ELBO}^\text{new}$ as our objective function which is a lower bound to Eq. (\ref{IB}).
% \derek{Need double check. May need detailed derivation?}

\vspace{-2mm}
\subsection{Encoder}\label{sec:encoder}
The encoder aims to encode both the structural and the dynamic information of NTS data. Existing message-passing based GNNs typically only capture the \textit{local} information from close neighbors. However, long-distance information between nodes is important in NTS data since the graph is constantly evolving and interactions between nodes can occur at any time step. Therefore, to capture this long-distance \textit{global} information, we propose using position embeddings with random walk with restart (RWR)~\cite{tong2006fast,DBLP:conf/sigir/FuH21,yan2021bright,yan2022dissecting}.

% \yc{emphasize the specially designed position embedding from here.}

% \textcolor{red}{ encode information.
% Existing GNNs, close neighbors. Long-distance info important (dynamic, evolving scenario).
% To capture long-dist info, propose using position embedding.
% local: GNNs
% global: long-dist info
% high-order / homophily
% }

\vspace{-1mm}
\subsubsection{RWR-based Position Embeddings}\label{sec:rwr}
For a graph $G_t$ at time step $t$, the relative position vector
% \yc{vector}
for all
% \yc{all}
nodes w.r.t.
% \yc{w.r.t.}
one anchor node $i$ is computed by RWR as follows:
% \hh{replace p by another letter such as c (we also use p for encoder/decoder)}
% \yc{as following}:
\setlength\abovedisplayskip{5pt}
\setlength\belowdisplayskip{5pt}
\begin{equation}\label{rwr-equa}
    \mathbf{r}_{t,i} = (1-c)\mathbf{\hat{A}}_t\mathbf{r}_{t,i} + c\mathbf{e}_i
\end{equation}
where $\mathbf{\hat{A}}_t = (\mathbf{D}_t^{-1}\mathbf{A}_t)^\top$ is the normalized adjacency matrix of $G_t$, $\mathbf{e}_i \in \mathbb{R}^{N}$ is a one-hot vector which only contains nonzero value at position $i$ and $c$ is the restart probability.
% The restart probability $p$ is set to the commonly used 0.15 \yc{move to experiment}.
After reaching the stationary distribution, we concatenate the position scores $\mathbf{r}_{t,i} \in \mathbb{R}^{N}$
% \yc{or $\mathbf{r}_{t,i}$?}
of all the anchor nodes as the final position embeddings $\mathbf{R}_t \in \mathbb{R}^{N \times N}$,
% \hh{do we need to use all the N nodes as the anchors?} \derek{in practice, we don't and we mention it in the first paragraph of next subsection. We assume this to prove the proposition and theorem for the global information that can be captured by RWR-based position embeddings.}\hh{ok}
% \yc{use () for indexing}
where $N$ is the number of nodes.

%%from HH: i removed subsubsection title -- as the ELBO part is also 'theoretic analysis'.
%\subsubsection{Theoretical Analysis}
We next prove the expressive power of RWR-based position embeddings with following proposition and theorem.

% \RQ{The current statement of the proposition \ref{rwr} looks like a remark or an explanation. It might be clearer if the statement is specific and exact. For instance, it is unclear to me what is the specific definition of local/global information. Explanation about what this proposition implies could be put before or after the statement.}
\vspace{-1mm}
\begin{proposition}\label{rwr}
Random walk with restarts (RWR) captures information from close neighbors (local) and long-distance neighbors (global) in graph learning.
% both local (close neighbors) and global (long-distance neighbors) information in graph. 
% \hh{not sure if we need this as a proposition. maybe we can embedded this into the proof of theorem 4.2}
\end{proposition}

\begin{proof}
\vspace{-1mm}
See Appendix.
% \hh{since kdd allows unlimited appendix. let's move all the proof there}
% Based on Eq.~\eqref{rwr-equa}, the closed form solution for RWR can be derived as: $\mathbf{r}_i = p \cdot (\mathbf{I} - (1-p) \cdot \mathbf{\hat{A}}) ^ {-1} \mathbf{e}_i$, where $\mathbf{I}$ is the identity matrix. We could also solve this equation by power iterations: $(\mathbf{I} - (1-p) \cdot \mathbf{\hat{A}}) ^ {-1} \approx \sum_{t=0}^{\infty} ((1-p)\cdot \mathbf{\hat{A}})^t$.
% % \hh{i think the relationship is exact since you have the summation to infinite}.
% First of all, as the power term $t$ goes to infinity, the position embedding $\mathbf{R}$ can indeed capture \textit{global} information of graph. Second, the restart probability $p$ ensures nodes close to anchor nodes have larger values than those farther away, which encodes the \textit{local} information of graph.
\vspace{-1mm}
\end{proof}

% \RQ{It seems that the proof implicitly assumes that the graph is connected (or strongly connected if it is a directed graph), which ensures that the random walk on it is ergodic. Without ergodicity, RWR cannot capture global information. It might be clearer if assumptions are included in the statement of the proposition.}

% \derek{
% \begin{theorem}\label{theo}
% (Expressiveness) Given a temporal graph $\mathcal{G}$, if the temporal computation tree (TCT) \cite{souza2022provably} of two nodes $u$ and $v$ are different, then message passing temporal GNNs with RWR-based node position embeddings can distinguish two nodes $u$ and $v$.
% \end{theorem}

% \begin{proposition}\label{limitation}
% (Limitations) 
% \end{proposition}

% }

% \derek{
% By introducing TGN, we extend the setting to temporal graphs, however, we need to update all the statements and analysis in the framework section to align our methods with TGN.
% }

The benefit of RWR-based position embeddings in temporal graphs is summarized in Theorem~\ref{temporal-theo} from the perspective of message-passing
% \yc{message-passing}
based temporal graph networks (TGN)~\cite{rossi2020temporal}, which is a general class of GNNs designed for handling temporal graphs. % and shares similar design ideas of different components in \vae. 
It contains two main components: \textit{memory} (through RNNs) for capturing the dynamics of each node; \textit{aggregate and update} (through GNNs) for gathering topological information from neighbors.
\vspace{-1mm}
\begin{theorem}\label{temporal-theo}
Given a temporal graph $\mathcal{G}$, TGN with RWR-based node position embeddings $g_{\theta}$ has more expressive power than regular TGN $f_{\theta}$ in node representation learning: $\mathbb{D}(g(u),g(v)) \geq \mathbb{D}(f(u),f(v))$ where $\mathbb{D}(\cdot, \cdot)$ measures the expressiveness by counting the distinguishable node pairs $(u,v)$ in $\mathcal{G}$ based on node representations.
% \hh{do we need an equation similar to eq13 here?}
\end{theorem}
\begin{proof}
\vspace{-1mm}
See Appendix.
\vspace{-1mm}
\end{proof}

Finally, to capture the dynamic information in NTS data, we use a 2-layer gated recurrent unit (GRU)~\cite{chung2014empirical} as the encoder
% \hh{the encoder now includes both genrating R and GRU right?} \derek{Yes.}
to model $q_{\theta}(z|\Tilde{\mathcal{X}},\mathcal{M},\mathcal{R})$, where 
% $\Tilde{\mathcal{X}}$ is the observed node feature sequence with missing values, $\mathcal{M}$ is the corresponding mask tensor, $\Tilde{\mathcal{A}}$ is the adjacency matrix of incomplete graph and 
$z$ is the latent representation and $\mathcal{R}=\{ \mathbf{R}_1, ..., \mathbf{R}_T \}$ is the tensor of node position embeddings. For each $\mathbf{R}_t$, instead of treating all the nodes as anchor nodes, usually only a small subset of anchor nodes $|S| = L$ would be sufficient to distinguish nodes from each other
% \hh{what does this mean? -- we say 'distinguish X from Y'}
% \derek{Explanation for section 3.1.1 when using $\mathbf{R}_t \in \mathbb{R}^{N \times N}$}
in practice~\cite{you2019position}.
% \yc{move to subsection 3.2.}
% \yc{mention $\Tilde{\mathcal{X}}$ and $\Tilde{\mathcal{A}}$ in Section 3}
% We use embedding $\mathbf{R}[i,:]$
% % \yc{adjust all notations.}
% of each node $i$ calculated from $\Tilde{\mathcal{A}}$ as an additional position-aware features of nodes in the graph.\yc{This sentence is redundant.}
Masks $\mathcal{M}$ and position embeddings $\mathcal{R}$ are concatenated with the input $\Tilde{\mathcal{X}}$ at each time step before feeding into the GRU.

\vspace{-2mm}
\subsection{Decoder}\label{sec:decoder}
% \hh{let's refine figure 2: 1. the decoder seems overly complicated. maybe we will have three main boxes for decoder, denoting the three stages, plus a few extra essential components, but no need to give all the details in figure. 2. do we also handle missing nodes? 3. what are Ut and Vt at left bottom?}
We design the decoder as a GRU with 3-stage predictions. We use $\mathbf{H}_t$ to denote node embedding matrix at time step $t$ and $\mathcal{H}$ to denote node embedding tensor.
% Since both the node feature time series and the sequence of graph structures capture certain dynamics of the data, imputation over one of them might be instructive for the other. Therefore, 
Based on the analysis in Section~\ref{sec:multi-task}, we model the complementary relation between feature imputation $p_{\phi}(\Tilde{\mathcal{X}}|\Tilde{\mathcal{A}}, \mathcal{M}, z)$
% \yc{mark}
and network imputation $p_{\gamma}(\Tilde{\mathcal{A}}|\mathcal{M}, \mathcal{R}, z)$ at different prediction stages in the decoder as follows.

% \RQ{Here $p_{\phi}(\Tilde{\mathcal{X}}|\Tilde{\mathcal{A}}, \mathcal{M}, z)$ and $p_{\gamma}(\Tilde{\mathcal{A}}|\mathcal{M}, \mathcal{R}, z)$ share a name $p$. It might be clearer if they have different names.}
% \derek{$p$ is usually used as the notation for decoders in VAE. They are distinguished based on the subscripts, which denotes different parameters.}
\vspace{-1.5mm}
\subsubsection{First-stage Feature Prediction}\label{1-stage} In the first stage, we use a linear layer
% \hh{how do we get $H_0$?}
% \hh{1. let's an equation number for main equations. 2. in stage 2, we use $Linear()$ to denote a linear layer. we should use consistent formats}
to generate an initial prediction of the missing values in the time series:
% \[
% \mathbf{\hat{Y}}^1_t = \mathbf{H}_{t-1}\mathbf{W}_h + \mathbf{b}_h
% \]
\begin{align}
    \mathbf{\hat{Y}}_{1,t} = \text{Linear}(\mathbf{H}_{t-1})
\end{align}
where $\mathbf{H}_{t-1}$ is the hidden representation of each node from the previous time step and $\mathbf{H}_0$ is sampled from a normal distribution $N(0,1/\sqrt{d_h})$ where $d_h$ is the hidden dimension. Similar to~\cite{cini2021filling}, we then use a filler operator to replace the missing values in the input $\Tilde{\mathbf{X}}_t$ with $\mathbf{\hat{Y}}_{1,t}$
% \hh{is this $\mathbf{\hat{Y}}^{1}_t$}
to get the first-stage output $\mathbf{O}_t$:
\begin{align}\label{eq:filler}
    \mathbf{O}_t = \mathbf{M}_t \odot \Tilde{\mathbf{X}}_t + (1 - \mathbf{M}_t) \odot \mathbf{\hat{Y}}_{1,t}
\end{align}

\vspace{-1.5mm}
\subsubsection{Second-stage Link Prediction}\label{2-stage} Our second-stage prediction imputes the missing weighted edges within graphs. $\mathbf{O}_t$ is used with the mask $\mathbf{M}_t$, the position embedding $\mathbf{R}_t$ and $\mathbf{H}_{t-1}$ to get the embeddings of all nodes at timestep $t$ through a linear layer:
\begin{align}
    \mathbf{U}_t = \text{Linear}(\mathbf{O}_t \| \mathbf{M}_t \| \mathbf{R}_t \| \mathbf{H}_{t-1})
\end{align}
where $\|$ is concatenation. We directly use the hidden states from previous time step $\mathbf{H}_{t-1}$ as the embeddings
% \hh{do we really need Vt? can we avoid it?}$\mathbf{V}_t$ (i.e., $\mathbf{V}_t = \mathbf{H}_{t-1}$)
for those missing nodes
% \hh{do we handle this?}
% \derek{Yes, in Eq.(12)}
since no new features or graph structures of them are available at time step $t$.
% \yc{From here}Besides, compared to other sequential models such as Transformers \cite{vaswani2017attention}, RNNs in our decoder may suffer a lot when data contains missing values and has irregular spacing between observations \cite{shukla2021multi}, especially when the missing rate is high. Therefore, we need an effective way to encode time to better utilize temporal information to enhance the link predictions. We use the time encoding from \cite{xu2020inductive} to tackle this problem:\yc{too redundant.}
% \hh{move this after Eq.11}
In NTS, observations are usually obtained by irregular sampling and the imputation problem over them can occur at any future step in real world problems. Being able to handle such uncertainty and forecasting unseen graph structure/time series data in the future time step are two key characteristics of an NTS imputation model. Therefore, in order to capture the dynamics between different timestamps and enhance the expressiveness of \vae, we also encode the time information with learnable Fourier features based on Bochner's theorem~\cite{xu2019self,xu2020inductive}, whose properties are summarized in Proposition~\ref{time-embed}, as follows:
% \yc{ Based on Bochner's theorem \cite{xu2020inductive}, we encode the time information as following: Why the time encoder is introduced in the decoder subsection?}
% , which is derived based on Bochner's theorem:
% \hh{is $f(t)$ a vector or matrix? if the latter, shall we change it to $\mathbf{F}_t$? If not, we should introduce its matrix counterpart to be used in Eq12}
\begin{align}
    f(t) = \sqrt{\frac{1}{k}}[\cos(\mathbf{w}_1t),\sin(\mathbf{w}_1t),...,\cos(\mathbf{w}_kt),\sin(\mathbf{w}_kt)]
\end{align}
where $\mathbf{w}_1,...,\mathbf{w}_k$ are learnable parameters.
\begin{proposition}\label{time-embed}
    Time encoding function f(t) is invariant to time rescaling and generalizes to any future unseen timestamps. 
\end{proposition}
% \RQ{Similar with the RWR proposition, it might be better if you clearly state your definition of translation invariance (e.g., which types of translation) and generalization (e.g., near future or far future).}
\begin{proof}
\vspace{-1mm}
    See~\cite{xu2020inductive,kazemi2019time2vec}.
\vspace{-1mm}
\end{proof}
Then, we concatenate node embeddings with time encodings through broadcasting as the input of a two-layer multi-layer perceptron (MLP) to predict the missing edges:
\begin{align}
    \mathbf{A}^{\text{out}}_t = \text{MLP}(\mathbf{U}_t \| \mathbf{H}_{t-1} \| f(t))
\end{align}
% Afterwards, we fill the incomplete graph $\Tilde{\mathbf{A}}_t$ with the predicted edges $\mathbf{A}^{\text{miss}}_t$ and get the predicted graph $\mathbf{A}^{\text{out}}_t$. \hh{$\mathbf{A}^{\text{out}}_t$, $\mathbf{A}^{\text{miss}}_t$, and later you also have $\mathbf{A}^{\text{pred}}_t$. which is which? do we need all of them?}
% \yc{updated graph or predicted graph? consistent}
The next step is to enhance node embeddings with updated graph structures. The general
% \hh{class}
class of message-passing neural networks (MPNNs) \cite{gilmer2017neural} is used similar to the \textit{aggregate and update} step in TGN
% \hh{or MPNN?}
% \derek{It should be TGN which aligns our analysis/proof in subsection 3.1}
to capture the graph topological information, 
which can be defined as:
\begin{align}
    \mathbf{H}_{t}^{\text{graph}} = \text{MPNN}(\mathbf{U}_t, \mathbf{A}_t^{\text{out}})
\end{align}
whose detailed design can be found in Appendix.
% \hh{can we simply use sth like $H^{g} = \text{MPNN}(A^{out})$? in a similar format as for Linear, MLP, Attn? if we want, we can move these details to appendix.}
% \begin{align}
% \text{MPNN}(F_u, F_m, \mathbf{d}_{t,i}, \mathbf{A}) = F_u(\mathbf{d}_{t,i}, & \sum_{\substack{j \in \mathcal{N}(i)}} F_m(\mathbf{h}_{t,i}, \mathbf{d}_{t,j}, e_{i,j}))
% \end{align}
% \noindent where $F_u$ and $F_m$ are update and message functions with learnable parameters, $\mathbf{d}_{t,i}$ is the node representation for node $i$ at time step $t$, $e_{i,j}$ is the edge weight between node $i$ and $j$, and $\mathcal{N}(i)$ represents node $i$'s neighbors.We apply a two-layer MPNNs with skip connection to obtain the hidden node embeddings $\mathbf{H}_t^{out}$ based on the predicted graph $\mathbf{A}^{\text{out}}_t$:
% % \hh{can we change $F_{u,1}$ to $F_{u}^1$? same change for $F_{u,2}$, $F_{m,1}$, and $F_{m,2}$}
% \begin{align}
% \begin{split}
%     & \mathbf{H}_{1,t} = \text{MPNN}(F_u^1, F_m^1, \mathbf{U}_t, \mathbf{A}_t^{\text{out}}) \\
%     & \mathbf{H}_{2,t} = \text{MPNN}(F_u^2, F_m^2, \mathbf{H}_{t,1}, \mathbf{A}_t^{\text{out}}) \\
%     & \mathbf{H}_{t}^{\text{graph}} = \mathbf{H}_{1,t} \oplus \mathbf{H}_{2,t}
% \end{split}
% \end{align}
% where $\oplus$ is the element-wise addition.

\vspace{-1.5mm}
\subsubsection{Third-stage Feature Prediction}\label{3-stage} In the third-stage prediction, we utilize the structural information $\mathbf{H}_{t}^{\text{graph}}$ to make a fine-grained imputation again over node features time series. Aiming to enhance the semantics of the node representations, we apply a self attention layer~\cite{vaswani2017attention} to capture cross-node information
% \textit{local}\yc{avoid using the term local here.} information at time $t$. Aiming to enable our model to have a broader view of graph dynamics along the entire sequence, we apply a self attention layer from \cite{vaswani2017attention} to capture inter-node information 
in our third-stage prediction,
% , which can be formulated as:
% \hh{if this is the standard self-attention mechanism, we can describe it in a similar way as linear layer or MLP, and skip the detailed notations and equations.}
% \begin{align}
% \begin{split}
%     & \text{Attn}(\mathbf{h}_{t,i}) = \sum_{j\in \mathcal{N}(i)} \text{softmax}\big ( \frac{\mathbf{Q}_i\mathbf{K}_j^T}{\sqrt{d}} \big ) \mathbf{V}_j \\
%     \text{and} \quad & \mathbf{Q}_i = \mathbf{h}_{t,i}\mathbf{W}_Q, \mathbf{K}_j = \mathbf{h}_{t,j}\mathbf{W}_K, \mathbf{V}_j = \mathbf{h}_{t,j}\mathbf{W}_V
% \end{split}
% \end{align}
% where $\mathcal{N}(i)$ is the neighbor of node $i$, $d$ is the hidden dimension and $\mathbf{W}_Q,\mathbf{W}_K,\mathbf{W}_V$ are learnable parameters.
which helps to encode richer node interaction information that is not captured in $\mathbf{H}_{t}^{\text{graph}}$.
% \textit{global} node interaction\yc{avoid using the term global here, or maybe we can change into time-local/global and spatial local/global} information within the entire sequence through training.
The latent node representations
% \hh{Z? to be consistent with Eq13}
$\textbf{Z}$,
% \hh{need to be bold?},
previous hidden state $\mathbf{H}_{t-1}$, the structural representation $\mathbf{H}_{t}^{\text{graph}}$ and the first stage output $\mathbf{O}_t$ as well as the masks $\mathbf{M}_t$ are all concatenated and processed by a self attention layer with an MLP to get the final output imputation representations:
% \hh{change z to Z?}
\begin{align}
    \mathbf{H}_t^{\text{out}} = \text{MLP}(\text{Attn}(\mathbf{Z} \| \mathbf{H}_{t-1} \| \mathbf{H}_{t}^{\text{graph}} \| \mathbf{O}_t \| \mathbf{M}_t))
\end{align}
Then a two-layer MLP is used for the third-stage
% \hh{third-stage}
prediction:
\begin{align}
    \hat{\mathbf{Y}}_{2,t} = \text{MLP}(\mathbf{H}_t^{\text{out}} \| \mathbf{H}_{t-1} \| \mathbf{H}_{t}^{\text{graph}})
\end{align}
A filler operator similar to Eq.~\eqref{eq:filler}
% \hh{similar to Eq.9?}
is applied to get the imputation output $\mathbf{X}^{\text{out}}_t$ from $\hat{\mathbf{Y}}_{2,t}$.
% \hh{should this be $\mathbf{\hat{Y}}^{2}_t$}.
Finally, a single layer GRU is used similar to the \textit{memory} step in TGN to update hidden representations based on the latent node representation $\textbf{Z}$, the output of second-stage $\mathbf{X}^{\text{out}}_t$, the mask $\mathbf{M}_t$ and the structural representation $\mathbf{H}_t^{\mathcal{A}}$ for each node and move on to the next time step:
% \hh{change z to Z?}
\begin{align}
    \mathbf{H}_{t} = \text{GRU}(\mathbf{Z} \| \mathbf{X}^{\text{out}}_t \| \mathbf{M}_t \| \mathbf{H}_t^{\text{graph}})
\end{align}
% \subsection{Bidirectional Model}\yc{This subsection is too short, which may be deleted.}
% \subsubsection{Bidirectional Model} Similar to \cite{cini2021filling}, we extend our VAE model to bidirectional by duplicating the architecture to handle the forward and backward sequences. An MLP is used over the output hidden representations for all time steps from VAE to produce the final imputation:
% \[
% \hat{\mathcal{Y}} = \text{MLP}(\mathcal{H}^{fwd}_{out} || \mathcal{H}^{bwd}_{out})
% \]
% where $\mathcal{H}_{out}$ is the tensor for imputation representation from third-stage predictions, $fwd$ and $bwd$ stand for forward and backward direction respectively.

\subsubsection{Bidirectional Model} Similar to~\cite{cini2021filling}, we extend our VAE model to bidirectional by replicating the architecture to handle both the forward and backward sequences. An MLP is used over the output hidden representations from these two VAEs to produce the final imputation output $\hat{\mathcal{Y}}$:
% \hh{1. not very clear. 2. what is $\hat{\mathcal{Y}}$? 3. would it be better if we change fwd to f and bwd to b?}
\begin{align}\label{bidirect}
\hat{\mathcal{Y}} = \text{MLP}(\mathcal{H}_{f}^{\text{out}} \| \mathcal{H}_{b}^{\text{out}} \| \mathcal{H}_{f}^{\text{graph}} \| \mathcal{H}_{b}^{\text{graph}} \| \mathcal{H}_{f} \| \mathcal{H}_{b})
\end{align}
where $\mathcal{H}^{\text{out}}$ is the tensor of imputation representations from the final stage prediction,
% \hh{$\textrm{fwd}$ and $\textrm{bwd}$}
$f$ and $b$ stand for forward and backward directions respectively. Algorithm~\ref{alg:pogevon-algo} summarizes the detailed workflow of the proposed \vae.
% \hh{can we have an actual algorithm?}
\vspace{-0.1cm}
\begin{algorithm}[ht]
\setlength{\belowcaptionskip}{5pt}
\caption{\vae: Position-aware Graph Enhanced Variational Autoencoders
% \hh{can we also put the full name, e.g., \vae: xxxx}
}
\label{alg:pogevon-algo}
\begin{algorithmic}[1]
\REQUIRE A partially observed NTS:
% with graph sequence
$\mathcal{G}(\Tilde{\mathcal{A}}, \Tilde{\mathcal{X}})=\{ \Tilde{G}_1, \Tilde{G}_2, ..., \Tilde{G}_T \}$.
\ENSURE The predicted tensor $\mathcal{X}$ of node feature time series and the predicted graph adjacency tensor $\mathcal{A}$.
\STATE Generate node position embeddings $\mathcal{R}$ based on Eq. \eqref{rwr-equa}.
% random walk with restart.
\FOR{$e = 1,2,3,..., \text{num\_epochs}$}
\STATE Encode $\Tilde{\mathcal{X}}_{f},\mathcal{M}_{f},\mathcal{R}_{f}$ to get $z_{f}$ based on Section~\ref{sec:encoder}.
% Use encoder $q_{\theta}(z_{f}|\Tilde{\mathcal{X}}_{f},\mathcal{M}_{f},\mathcal{R}_{f})$ to get $z_{f}$ based on Section~\ref{sec:encoder}.
% \hh{refer to the related equations or subsection numbers}.
\FOR{$t = 1,2,3, ..., T$ (forward direction)}
% \hh{is this for forward direction? if so, mark it (as a comment). same for the backward direction}
\STATE Perform first-stage decoding based on Section~\ref{1-stage}.
\STATE Perform second-stage decoding based on Section~\ref{2-stage}.
\STATE Perform third-stage decoding based on Section~\ref{3-stage}.
\ENDFOR
\STATE Encode $\Tilde{\mathcal{X}}_{b},\mathcal{M}_{b},\mathcal{R}_{b}$ to get $z_{b}$ based on Section~\ref{sec:encoder}.
% Use encoder $q_{\theta}(z_{b}|\Tilde{\mathcal{X}}_{b},\mathcal{M}_{b},\mathcal{R}_{b})$ to get $z_{b}$ based on Section~\ref{sec:encoder}.
\FOR{$t = T, T-1, ..., 1$ (backward direction)}
\STATE Perform first-stage decoding based on Section~\ref{1-stage}.
\STATE Perform Second-stage decoding based on Section~\ref{2-stage}.
\STATE Perform third-stage decoding based on Section~\ref{3-stage}.
\ENDFOR
\STATE Generate final outputs $\hat{\mathcal{Y}}$ based on Eq.~\eqref{bidirect}.
\STATE Update parameters $\theta, \gamma, \phi$ by optimizing the loss in Eq.~\eqref{loss}.
\ENDFOR
\STATE Obtain the predicted tensor $\mathcal{X}$ of node feature time series based on Eq.~\eqref{eq:filler}  
by replacing missing values in $\hat{\mathcal{X}}$ with $\hat{\mathcal{Y}}$.
\STATE %Use Eq.~\eqref{eq:filler} to 
Obtain the predicted graph adjacency tensor $\mathcal{A}$ based on Eq.~\eqref{eq:filler} by replacing missing values in $\hat{\mathcal{A}}$ with $\mathcal{A}^{\text{out}}$.
\RETURN the predicted tensor time series $\mathcal{X}$ and the predicted tensor of graph adjacency $\mathcal{A}$.
\end{algorithmic}
\end{algorithm}
% \derek{To be updated.}

\vspace{-1.5mm}
\subsection{Objective and Training}\label{sec:objective}
The \textit{Evidence Lower Bound } (\textit{ELBO}) objective function of a vanilla conditional VAE~\cite{doersch2016tutorial,collier2020vaes} over missing data imputations can be defined as:
\begin{align}\label{cvae-elbo}
\begin{split}
\text{ELBO}(\theta, \phi) & = \mathbb{E}_q[\log p_{\phi}(\Tilde{\mathcal{X}}|z, \mathcal{M})] - \\  & \mathbb{KL}[q_{\theta}(z|\Tilde{\mathcal{X}},\mathcal{M}) || p_{\phi}(z)] \leq \log p_{\phi}(\Tilde{\mathcal{X}}|\mathcal{M})
\end{split}
\end{align}
Our goal is to learn a good generative model of both
% \hh{short name, e.g., observed node feature time series}
the observed multivariate node feature time series $\Tilde{\mathcal{X}}$ and
% \hh{observed graph adjacency}
the observed graph adjacency $\Tilde{\mathcal{A}}$. Thus, we can treat
% \hh{short name of R, e.g., position embedding}
the position embeddings $\mathcal{R}$ as an extra condition in addition to the mask $\mathcal{M}$ similar to~\cite{ivanov2018variational}. This is because, $\mathcal{M}$ and $\mathcal{R}$ are auxiliary covariates, and are given or can be generated
% \hh{but M is given right?}
through deterministic functions based on $\Tilde{\mathcal{A}}$ and $\Tilde{\mathcal{X}}$ respectively. Therefore, it is more natural to maximize $\log p (\Tilde{\mathcal{X}}, \Tilde{\mathcal{A}} | \mathcal{M}, \mathcal{R})$ as our objective, which is summarized in the following lemma.
% as well as the additional condition $\Tilde{\mathcal{A}}$.
% and based on which we present the following lemma for deriving the new ELBO of our model: \hh{what is the main conclusion of this lemma? e.g., do we mean the new ELBO is better than Eq. 19? how does that connect to the conclusion at the end of sec 2?}
\begin{lemma}
\vspace{-1mm}
% With the assumption that
% % \hh{one random variable is independent of another}
% the prior $p(z)$ is jointly independent of
% % \hh{which random variable? this part of the sentence is not very clear}
% condition $\mathcal{M}$ and $\mathcal{R}$
%With the assumption that 
Under the condition that $\mathcal{M}$ and $\mathcal{R}$ are jointly independent of the prior $p(z)$: $p(z) = p(z | \mathcal{M}, \mathcal{R})$, the new ELBO objective of the proposed \vae\ for the NTS imputation problem is: 
% \hh{consider to turn this into a lemma}
\begin{align}
\begin{split}
 & \text{ELBO}^\text{new}(\theta, \gamma, \phi)= \mathbb{E}_q[\log p_{\phi}(\Tilde{\mathcal{X}}|\Tilde{\mathcal{A}}, \mathcal{M}, z)]  \\ & + \mathbb{E}_q[\log p_{\gamma}(\Tilde{\mathcal{A}}|\mathcal{M}, \mathcal{R}, z) - \mathbb{KL}[q_{\theta}(z|\Tilde{\mathcal{X}}, \mathcal{M}, \mathcal{R}) || p_{\phi}(z)]]
 \end{split}
\end{align}
where $\gamma$ denotes parameters of the link prediction module.
\end{lemma}
\vspace{-2mm}
\begin{proof}
\vspace{-1mm}
The derivation of %the new ELBO of our method 
$\text{ELBO}^\text{new}$ can be formulated as:
\vspace{-1mm}
\begin{align*}
    & \log p (\Tilde{\mathcal{X}}, \Tilde{\mathcal{A}} | \mathcal{M}, \mathcal{R}) = \log \int p(\Tilde{\mathcal{X}}, \Tilde{\mathcal{A}} | \mathcal{M}, \mathcal{R}, z) p(z) dz \\[-1mm]
    & = \log \int p(\Tilde{\mathcal{X}} | \Tilde{\mathcal{A}}, \mathcal{M}, \mathcal{R}, z) p(\Tilde{\mathcal{A}}|\mathcal{M}, \mathcal{R}, z) p(z) dz \\
    % & \text{since $p(\mathcal{R}|\Tilde{\mathcal{A}})=1$,} \\
    & \text{since the node position embedding $\mathcal{R}$ can be generated from} \\[-1mm] & \text{the observed graph adjacency $\Tilde{\mathcal{A}}$,} \\[-1mm]
    & = \log \int p(\Tilde{\mathcal{X}} | \Tilde{\mathcal{A}}, \mathcal{M}, z) p(\Tilde{\mathcal{A}}|\mathcal{M}, \mathcal{R}, z) p(z) \frac{q(z|\Tilde{\mathcal{X}}, \mathcal{M}, \mathcal{R})}{q(z|\Tilde{\mathcal{X}}, \mathcal{M},\mathcal{R})} dz \\
    & = \log \mathbb{E}_q[p(\Tilde{\mathcal{X}} | \Tilde{\mathcal{A}}, \mathcal{M}, z) p(\Tilde{\mathcal{A}}|\mathcal{M}, \mathcal{R}, z) \frac{p(z)}{q(z|\Tilde{\mathcal{X}}, \mathcal{M},\mathcal{R})}] \\
    & \geq \mathbb{E}_q[\log p(\Tilde{\mathcal{X}} | \Tilde{\mathcal{A}}, \mathcal{M}, z)] + \mathbb{E}_q[\log p(\Tilde{\mathcal{A}}|\mathcal{M}, \mathcal{R}, z)]
    \\ & - \mathbb{KL}[q(z|\Tilde{\mathcal{X}}, \mathcal{M}, \mathcal{R}) || p(z)] \qedhere
\end{align*}
% \yc{mark}
\vspace{-6mm}
\end{proof}
\noindent This lemma generalizes the ELBO in Eq.~\eqref{cvae-elbo} to the multi-task learning setting which ensures the learning objective of the proposed \vae\ is consistent with our analysis in Section~\ref{sec:multi-task}. That is, $\text{ELBO}^\text{new}$ corresponds to Eq. \eqref{IB} by modeling dependencies between the observed node time series $\mathcal{X}$ and observed graph adjacency $\mathcal{A}$ similar to Eq. \eqref{objective}.
% \hh{1-2 sentences to remind what this analysis is, e.g., That is, xxxx}

% The derivation of $\text{ELBO}^\text{new}$ can be found in the Appendix \ref{appd:elbo}.
We use a similar strategy as in ~\cite{collier2020vaes,nazabal2020handling} to maximize $\text{ELBO}^\text{new}$ by training our model over observed data and infer missing ones based on $p(\mathcal{X}|\Tilde{\mathcal{X}}) \approx \int p(\mathcal{X}|z)q(z|\Tilde{\mathcal{X}})dz$. We (1) use the mean absolute error (MAE) as the error function for the feature imputation and (2) use the Frobenius norm between the predicted adjacency matrices and the observed adjacency matrices
% \yc{mark, observed}
as the link prediction loss.
% : $L_{\mathcal{A}} = \|\Tilde{\mathcal{A}} - \mathcal{A}^{\text{out}}\|_F$.
% to approximate $\mathbb{E}_q[\log p(\mathcal{A}|\Tilde{\mathcal{A}}, \mathcal{M}, z)]$ in $\text{ELBO}^\text{new}$.
The model is trained by minimizing the following loss function which is composed of errors of all three stages:
\begin{align}\label{loss}
    & \mathcal{L} =
    \underbrace{L(\hat{\mathcal{Y}}_{t:t+\Delta t}, \Tilde{\mathcal{X}}_{t:t+\Delta t}, \mathcal{M}_{t:t+\Delta t}) + \beta \cdot \mathbb{KL}_{f} + \beta \cdot \mathbb{KL}_{b}}_{\text{First and third terms in $\text{ELBO}^\text{new}$}} \\
    & + \underbrace{L(\mathcal{O}_{f,t:t+\Delta t}, \Tilde{\mathcal{X}}_{t:t+\Delta t}, \mathcal{M}_{t:t+\Delta t}) + L(\mathcal{O}_{b,t:t+\Delta t}, \Tilde{\mathcal{X}}_{t:t+\Delta t}, \mathcal{M}_{t:t+\Delta t})}_{\text{Error for the 1$^{\text{st}}$ stage prediction}}\notag \\
    & + \underbrace{\gamma\cdot\|\Tilde{\mathcal{A}}_{f,t:t+\Delta t} - \mathcal{A}^{\text{out}}_{f,t:t+\Delta t}\|_F + \gamma\cdot\|\Tilde{\mathcal{A}}_{b,t:t+\Delta t} - \mathcal{A}^{\text{out}}_{b,t:t+\Delta t}\|_F}_{\text{Error for the 2$^{\text{nd}}$ stage prediction (i.e., second term in $\text{ELBO}^\text{new}$)}}\notag \\
    & + \underbrace{L(\mathcal{X}^{\text{out}}_{f,t:t+\Delta t}, \Tilde{\mathcal{X}}_{t:t+\Delta t}, \mathcal{M}_{t:t+\Delta t}) + L(\mathcal{X}^{\text{out}}_{b,t:t+\Delta t}, \Tilde{\mathcal{X}}_{t:t+\Delta t}, \mathcal{M}_{t:t+\Delta t})}_{\text{Error for the 3$^{\text{rd}}$ stage prediction}} \notag
\end{align}
where
% the first line correspond to the $\text{ELBO}^\text{new}$, the second line is for first-stage predictions, while the third and fourth lines guide decoder's second-stage predictions and third-stage predictions respectively.
$\beta$ is the weight for KL divergence similar to~\cite{higgins2016beta} and $\gamma$ is the weight for the 2$^{\text{nd}}$ stage prediction. The element wise error function $L(\mathcal{X}^{\text{pred}}, \mathcal{X}^{\text{label}}, \mathcal{M})$ outputs the average error by calculating the inner product between mask tensor $\mathcal{M}$ and $|\mathcal{X}^{\text{label}} - \mathcal{X}^{\text{pred}}|$. The loss $\mathcal{L}$ is optimized through each sample in the dataset which is a sliding window $(t:t+\Delta t)$ of NTS data (i.e., $\Tilde{\mathcal{G}}_{t:t+\Delta t}$).
% \yc{(1) This sentence has two calculate, not clear. (2) What is the fwd and bwd, you should clearly state the forward and backward.}

\vspace{-0.15cm}
\subsection{Complexity Analysis}\label{sec:complexity}
The computational complexity of \vae\ can be analyzed through the following aspects. First, calculating the position embedding $\mathcal{R}$ has the complexity $\mathcal{O}(T \cdot \bar{E} \cdot \log \frac{1}{\epsilon})$ \cite{wang2020personalized} where $\bar{E}$ is the average number of edges and $\epsilon$ is the absolute error bound for the power iteration of RWR. Second, with a standard bidirectional VAE based on GRU, MPNN increases the complexity by $\mathcal{O}(\bar{E})$ with sparse matrix multiplications at each time step. Third, the self-attention used in the third-stage decoder has the complexity $\mathcal{O}(N^2)$. There are several ways to reduce the overall time complexity. For example, %However, this can be amortized since 
most of the computations can be parallelized. One computational bottleneck lies in the computation of self-attention. The existing techniques for efficient attentions~\cite{tay2022efficient} can be readily applied in the proposed \vae, such as Linformer~\cite{wang2020linformer} which uses low-rank projections to make the cost of the attention mechanism $\mathcal{O}(N)$ and Reformer~\cite{kitaev2020reformer} which applies locality sensitive hashing to reduce the complexity of attention to $\mathcal{O}(N\cdot \log N)$. %Analysis of these alternatives compared to \vae\ can be found in Appendix.
% We provide analysis of these alternatives compared with \vae\ in the Appendix.

\vspace{-0.12cm}
\section{Experiment}\label{sec:experiment}

We apply the proposed \vae\ to the networked time series imputation task, and evaluate it in the following aspects:
\begin{itemize}
    \item $\mathcal{Q}1.$ How effective is \vae\ for networked time series imputation?
    % \derek{Mark.}
    % \yc{separate the tasks for \mathcal{X} and \mathcal{A}}
    \item $\mathcal{Q}2.$ To what extent does our method benefit from
different components of the model?
\end{itemize}

\vspace{-3mm}
\subsection{Experimental Setup}
\subsubsection{Datasets} We evaluate the proposed \vae\ model on five real-world datasets, and the statistics of all the datasets are listed in Table~\ref{table:data}.

\vspace{-3mm}
\begin{table}[ht]
\begin{center}
\setlength{\abovecaptionskip}{3pt}
\caption{Statistics of the datasets. Entity numbers of PeMS* datasets refer to the original number of sensors/stations in the corresponding dataset and only part of them are used to build the graphs.}
\begin{tabularx}{\columnwidth}{l|A|A|Y|Y} 
 \toprule
 Dataset & \# of entity & \# of nodes & average \# of edges & time length  \\ 
 \specialrule{1pt}{1pt}{3pt}
 COVID-19 & 50 & 50 & 1344.75 & 346 \\ 
 \midrule
 AQ36 & 36 & 36 & 341.57 & 8759 \\ 
 \midrule
 PeMS-BA & 1632 & 64 & 675.45 & 25920 \\
 \midrule
 PeMS-LA & 2383 & 64 & 1095.54 & 25920 \\
 \midrule
 PeMS-SD & 674 & 64 & 1295.11 & 25920 \\
 \bottomrule
\end{tabularx}
% \derek{I will try to see if there are other datasets that we can use (such as those for diffusion history/source localization research).
% Besides, I'm thinking to add one more experiment for different masking rates/scenarios, and make it a bar (or line) plot}
\label{table:data}
\end{center}
\vspace{-3mm}
\end{table}
% \yc{Appendix}.

% \begin{itemize}
%     \item \textbf{AQ36}:
%     % \yc{need discussion.}
%     A dataset of AQI values of PM2.5 pollutant captured by 36 sensors in Beijing \cite{cao2018brits,cini2021filling}.
%     \item \textbf{PeMS-BA}: A dataset of traffic flows from 64 sensors in Bay Area based on the Caltrans Performance Measurement System (PeMS) \cite{chen2001freeway}.
%     \item \textbf{PeMS-LA}: A dataset of traffic flows from 64 sensors in Los Angeles based on PeMS.
%     \item \textbf{PeMS-SD}: A dataset of traffic flows from 64 sensors in San Diego based on PeMS.
% \end{itemize}
% \vspace{-15pt}
% \derek{Mark. May need to move generation details to appendix due to limited space.}
\begin{itemize}[leftmargin=9mm]
    \item \textbf{COVID-19}: A dataset of COVID-19 infection cases and deaths in 50 states in USA from 01/21/2020 to 12/31/2020~\cite{katragadda2022examining}. Similar to~\cite{kapoor2020examining}, we choose infection cases of states as the time series data $\mathbf{X}$ and use mobility of people across different states to model the spatial relationship $\mathbf{A}$ between them. Then, we apply a Radial Basis Function (RBF) $f(u,v,t) = \exp(-\frac{||x^u_t - x^v_t||^2}{2\sigma^2})$ ~\cite{chen2021z} to capture the dynamics and generate the graph sequence. Finally, we simulate the missing edges in the NTS imputation problem by masking edges when one of its end nodes contains missing features.
    % \hh{not clear. do you mean 'one of its nodes has missing features'?}. 
    Specifically, an edge weight $w^{u,v}_t$ between nodes $u$ and $v$ at time $t$ can be defined as:
    \vspace{-0.5mm}
    \begin{equation}\label{data:construct}
      w^{u,v}_t =
        \begin{cases}
          w^{u,v} & \text{if $\mathbf{A}[u,v] \neq 0$ and $f(u,v,t) > k$} \\ 
          & \text{\qquad and $m^u_t = 1$, $m^v_t = 1$.} \\
          0 & \text{otherwise.}
        \end{cases}     
    % \vspace{-0.5mm}
    \end{equation}
    where $k$ is the positive threshold for graph dynamics and we choose $k=0.3$ for COVID-19 dataset. We randomly mask out $25\%$ of the node features in this dataset, and split %each of them along 
    the time axis to $70\%$ for training, $10\%$ for validation and $20\%$ for test respectively.
    
    \item \textbf{AQ36}: A dataset of AQI values of different air pollutants collected from various monitor stations over 43 cities in China~\cite{zheng2015forecasting}. 
    % Many previous works on MTS imputation use a subset of this dataset which only includes 36 stations \cite{cao2018brits,cini2021filling}. Due to the fact that a sequence of graphs needed in the NTS imputation may require large amount of memory, we only consider the reduced version of the dataset with only 36 nodes (\textbf{AQ36}), which has around $25\%$ missing values.
    Following~\cite{cao2018brits,cini2021filling}, we use the reduced version of the dataset which contains 36 nodes (\textbf{AQ36}) and pick the last four months as the test data.
    % Previous methods use the $\text{3}^\text{rd}$, $\text{6}^\text{th}$, $\text{9}^\text{th}$ and $\text{12}^\text{th}$ as the test data. Since our purpose is to predict the missing values in the future, we use the last four months as the test data in our experiments. \yc{remove, simplify.}
    To construct the static graph $G(\mathbf{A},\mathbf{X})$, we use the thresholded Gaussian kernel from~\cite{shuman2013emerging} to get the pairwise distances $\mathbf{A}[u,v]$ 
    % \yc{notation consistency.}
    between stations $u$ and $v$ as the edge weight. The graph sequence is constructed using the similar method as Eq. \eqref{data:construct} over normalized time series features and the threshold $k$ is set to 0.8. We use the same mask setting as~\cite{yi2016st} which simulates the true missing data distribution.

    % Then, we apply a Radial Basis Function (RBF) $f(u,v,t) = \exp(-\frac{||x^u_t - x^v_t||^2}{2\sigma^2})$ similar to \cite{chen2021z} to capture the dynamics and generate the graph sequence. Finally, we simulate the missing edges in NTS imputation problem by masking edges when one of its nodes' values is missing. Specifically, an edge weight $w^{u,v}_t$ between node $u$ and $v$ at time $t$ can be defined as:
    
    % \begin{equation}
    %   w^{u,v}_t =
    %     \begin{cases}
    %       w^{u,v} & \text{if $\mathbf{A}^{u,v} \neq 0$ and $f(u,v,t) > k$} \\ 
    %       & \text{\qquad and $m^u_t = 1$, $m^v_t = 1$.} \\
    %       0 & \text{otherwise.}
    %     \end{cases}       
    % \end{equation}
    % where $k$ is the positive threshold for graph dynamics and we choose $k=0.8$ in our experiments.
    
    \item \textbf{PeMS-BA/PeMS-LA/PeMS-SD}
    Three datasets contain traffic statistics based on the Caltrans Performance Measurement System (PeMS)~\cite{chen2001freeway}, which cover the freeway system in major areas of California. We collect 5-minute interval traffic flow data from 3 different stations 4, 7 and 11 between 01/01/2022 and 03/31/2022, which represent the traffic information from Bay Area, Los Angeles and San Diego respectively. For each dataset, we pick 64 sensors with the largest feature variance, and use their latitude and longitude values to calculate pairwise distances to build the static graph. We only keep edges with weight within certain threshold, and we use 15 miles for \textbf{PeMS-BA}/\textbf{PeMS-LA} and 10 miles for \textbf{PeMS-SD}. The graph sequence is constructed using the similar method as the AQ36 dataset, and the threshold $k$ is set to 0.8. We use similar masking settings as COVID-19 dataset.
    % We randomly mask out $25\%$ of the features in all these dataset, and split each of them along the time axis to $70\%$ for training, $10\%$ for validation and $20\%$ for test respectively.
\end{itemize}
\vspace{-0.5mm}

The missing rate of AQ36's time series features is about $13.24\%$, while for COVID-19 dataset and all the traffic datasets, the time series features have $25\%$ missing values. 
% Following the original papers,
Based on Eq. \eqref{data:construct}, the missing rates of edges for AQ36 is $28.06\%$, for COVID-19 is $43.23\%$, and for PEMS-BA/PEMS-LA/PEMS-SD are $43.75\%/43.74\%/43.71\%$ respectively.

To be consistent with the dataset settings in previous works such as GRIN~\cite{cini2021filling}, we use the following window length to train the models: (i) 14 for COVID-19 dataset which corresponds to 2 weeks, (ii) 36 for AQ36 dataset which corresponds to 1.5 days and (iii) 24 for all the traffic datasets which corresponds to 2 hours of data.

\begin{table*}[ht]
\setlength{\abovecaptionskip}{3pt}
\caption{Performance comparison over COVID-19 and AQ36 datasets. Smaller is better.}
\scalebox{0.71}{
\begin{tabular}{c | c c c | c c c} 
 \toprule
 & \multicolumn{3}{c|}{\Large \textbf{COVID-19}} & \multicolumn{3}{c}{\Large \textbf{AQ36}} \\
 \Large{\textbf{Models}} & MAE & MSE & MRE & MAE & MSE & MRE \\ [0.3ex] 
 \specialrule{1pt}{1pt}{3pt}
 Mean & 3.081 $\pm\ 0.000$ & 10.707 $\pm\ 0.000$ & 0.284 $\pm\ 0.000$ & 62.299 $\pm\ 0.000$ & 6525.709 $\pm\ \phantom{00}0.000$ & 0.835 $\pm\ 0.000$  \\
 \midrule
 MF & 0.276 $\pm\ 0.026$ & \phantom{0}0.165 $\pm\ 0.025$ & 0.026 $\pm\ 0.002$ & 39.582 $\pm\ 0.189$ & 4545.596 $\pm\ \phantom{0}61.411$ & 0.531 $\pm\ 0.002$ \\
 \midrule
 MICE & 0.077 $\pm\ 0.005$ & \phantom{0}0.013 $\pm\ 0.002$ & 0.007 $\pm\ 0.000$ & 38.889 $\pm\ 0.268$ & 4314.435 $\pm\ \phantom{0}20.617$ & 0.521 $\pm\ 0.003$ \\
 \midrule
 BRITS & 0.386 $\pm\ 0.006$ & \phantom{0}0.293 $\pm\ 0.009$ & 0.036 $\pm\ 0.001$ & 23.393 $\pm\ 0.802$ & 1276.226 $\pm\ 102.916$ & 0.314 $\pm\ 0.011$ \\
 \midrule
 rGAIN & 0.579 $\pm\ 0.069$ & \phantom{0}0.571 $\pm\ 0.106$ & 0.055 $\pm\ 0.006$ & 25.032 $\pm\ 1.426$ & 1358.134 $\pm\ 152.361$ & 0.335 $\pm\ 0.019$ \\
 \midrule
 SAITS & 0.466 $\pm\ 0.010$ & \phantom{0}0.366 $\pm\ 0.019$ & 0.043 $\pm\ 0.001$ & 51.097 $\pm\ 0.625$ & 5026.475 $\pm\ 75.120$ & 0.685 $\pm\ 0.008$ \\
 \midrule
 TimesNet & 0.028 $\pm\ 0.002$ & \phantom{0}0.002 $\pm\ 0.000$ & 0.003 $\pm\ 0.000$ & 40.700 $\pm\ 0.278$ & 3383.554 $\pm\ 49.499$ & 0.545 $\pm\ 0.004$ \\
 \midrule
 GRIN & 0.319 $\pm\ 0.038$ & \phantom{0}0.165 $\pm\ 0.040$ & 0.029 $\pm\ 0.004$ & 29.420 $\pm\ 0.231$ & 2050.726 $\pm\ \phantom{0}56.028$ & 0.394 $\pm\ 0.003$ \\
 \midrule
 $\text{NET}^3$ & 0.547 $\pm\ 0.004$ & \phantom{0}0.682 $\pm\ 0.006$ & 0.051 $\pm\ 0.000$ & 34.755 $\pm\ 0.497$ & 2473.718 $\pm\ \phantom{0}37.461$ & 0.466 $\pm\ 0.007$ \\
 \midrule
 \vae\ & \textbf{0.007} $\pm\ \textbf{0.001}$ & \phantom{0}\textbf{0.000} $\pm\ \textbf{0.000}$ & \textbf{0.001} $\pm\ \textbf{0.000}$ & \textbf{19.494} $\pm\ \textbf{1.101}$ & \textbf{1213.474} $\pm\ \textbf{125.529}$ & \textbf{0.261} $\pm\ \textbf{0.015}$ \\
 \bottomrule
\end{tabular}}
\label{table:result1}
\vspace{-2mm}
\end{table*}

\begin{table*}[ht]
\setlength{\abovecaptionskip}{3pt}
\caption{Performance comparison over PeMS-BA, PeMS-LA and PeMS-SD datasets. Smaller is better.}
\scalebox{0.71}{
\begin{tabular}{c | c c c | c c c | c c c} 
\toprule
 & \multicolumn{3}{c|}{\Large \textbf{PeMS-BA}} & \multicolumn{3}{c}{\Large \textbf{PeMS-LA}} & \multicolumn{3}{c}{\Large \textbf{PeMS-SD}} \\
 \Large{\textbf{Models}} & MAE & MSE & MRE & MAE & MSE & MRE & MAE & MSE & MRE \\ [0.3ex] 
 \specialrule{1pt}{1pt}{3pt}
 Mean & 192.047 $\pm\ 0.000$ & 47504.159 $\pm\ \phantom{00}0.000$ & 0.474 $\pm\ 0.000$ & 216.681 $\pm\ 0.000$ & 62664.657 $\pm\ \phantom{00}0.000$ & 0.406 $\pm\ 0.000$ & 208.192 $\pm\ 0.000$ & 55780.002 $\pm\ \phantom{00}0.000$ & 0.529 $\pm\ 0.000$ \\ 
 \midrule
 MF & \phantom{0}57.265 $\pm\ 1.148$ & \phantom{0}8091.407 $\pm\ 185.123$ & 0.141 $\pm\ 0.003$ & \phantom{0}77.339 $\pm\ 0.699$ & 15202.678 $\pm\ 156.348$ & 0.145 $\pm\ 0.001$ & \phantom{0}45.811 $\pm\ 0.318$ & \phantom{0}6044.345 $\pm\ \phantom{0}72.976$ & 0.117 $\pm\ 0.001$ \\
 \midrule
 MICE & \phantom{0}50.861 $\pm\ 0.765$ & \phantom{0}6724.148 $\pm\ 109.829$ & 0.126 $\pm\ 0.002$ & \phantom{0}64.018 $\pm\ 1.015$ & 10822.355 $\pm\ 405.410$ & 0.120 $\pm\ 0.002$ & \phantom{0}38.978 $\pm\ 1.036$ & \phantom{0}4771.186 $\pm\ \phantom{0}92.335$ & 0.100 $\pm\ 0.003$ \\
 \midrule
 BRITS & \phantom{0}30.274 $\pm\ 0.095$ & \phantom{0}2942.411 $\pm\ \phantom{0}16.511$ & 0.075 $\pm\ 0.000$ & \phantom{0}36.921 $\pm\ 0.133$ & \phantom{0}3681.595 $\pm\ \phantom{0}21.635$ & 0.069 $\pm\ 0.000$ & \phantom{0}21.232 $\pm\ 0.059$ & \phantom{0}1563.234 $\pm\ \phantom{0}28.309$ & 0.054 $\pm\ 0.000$ \\
 \midrule
 rGAIN & \phantom{0}38.862 $\pm\ 0.752$ & \phantom{0}3422.914 $\pm\ \phantom{0}61.281$ & 0.096 $\pm\ 0.002$ & \phantom{0}49.611 $\pm\ 1.083$ & \phantom{0}5533.964 $\pm\ 234.335$ & 0.093 $\pm\ 0.002$ & \phantom{0}33.212 $\pm\ 1.475$ & \phantom{0}2341.466 $\pm\ \phantom{0}98.314$ & 0.085 $\pm\ 0.004$ \\
 \midrule
 SAITS & \phantom{0}46.567 $\pm\ 0.530$ & \phantom{0}5412.574 $\pm\ 161.132$ & 0.115 $\pm\ 0.001$ & \phantom{0}61.896 $\pm\ 0.892$ & \phantom{0}10998.854 $\pm\ 204.345$ & 0.116 $\pm\ 0.002$ & \phantom{0}34.117 $\pm\ 0.886$ & \phantom{0}4101.397 $\pm\ 152.141$ & 0.087 $\pm\ 0.002$ \\
 \midrule
 TimesNet & \phantom{0}25.859 $\pm\ 0.115$ & \phantom{0}1676.843 $\pm\ \phantom{0}16.144$ & 0.064 $\pm\ 0.000$ & \phantom{0}27.452 $\pm\ 0.114$ & \phantom{0}2058.227 $\pm\ 6.213$ & 0.052 $\pm\ 0.000$ & \phantom{0}21.583 $\pm\ 0.085$ & \phantom{0}1284.300 $\pm\ \phantom{0}21.839$ & 0.055 $\pm\ 0.000$ \\
 \midrule
 GRIN & \phantom{0}30.057 $\pm\ 1.073$ & \phantom{0}1922.072 $\pm\ \phantom{0}74.327$ & 0.074 $\pm\ 0.003$ & \phantom{0}47.835 $\pm\ 2.059$ & \phantom{0}4561.512 $\pm\ 298.533$ & 0.090 $\pm\ 0.004$ & \phantom{0}41.001 $\pm\ 1.543$ & \phantom{0}3000.012 $\pm\ 201.018$ & 0.105 $\pm\ 0.004$  \\
 \midrule
 $\text{NET}^3$ & \phantom{0}35.671 $\pm\ 0.111$ & \phantom{0}2735.574 $\pm\ \phantom{00}6.138$ & 0.009 $\pm\ 0.000$ & \phantom{0}37.652 $\pm\ 0.113$ & \phantom{0}3416.784 $\pm\ \phantom{00}6.765$ & 0.071 $\pm\ 0.000$ & \phantom{0}34.111 $\pm\ 0.184$ & \phantom{0}2487.581 $\pm\ \phantom{00}9.798$ & 0.087 $\pm\ 0.000$ \\
 \midrule
 \vae\ & \phantom{0}\textbf{22.194} $\pm\ \textbf{0.046}$ & \phantom{0}\textbf{1248.681} $\pm\ \phantom{00}\textbf{4.297}$ & \textbf{0.055} $\pm\ \textbf{0.000}$ & \phantom{0}\textbf{23.905} $\pm\ \textbf{0.245}$ & \phantom{0}\textbf{1714.962} $\pm\ \phantom{0}\textbf{31.035}$ & \textbf{0.045} $\pm\ \textbf{0.000}$ & \phantom{0}\textbf{18.990} $\pm\ \textbf{0.112}$ & \phantom{00}\textbf{951.559} $\pm\ \phantom{00}\textbf{8.264}$ & \textbf{0.048} $\pm\ \textbf{0.000}$ \\
 \bottomrule
\end{tabular}}
\label{table:result2}
\vspace{-2mm}
\end{table*}

\vspace{-1mm}
\subsubsection{Baselines}

We compare the proposed \vae\ model with following baselines for the time series imputation task.
% \yc{too long}
% \yc{before to appendix.}
All the methods are trained with NVIDIA Tesla V100 SXM2 GPU.

% \hh{for tables, can we use the same font for std as the other numbers? right now, it looks wired. we can resize the entire table to make it fit into the page width}
\begin{enumerate}
    \item \textbf{Mean}. Impute with node level feature average along the sequence.
    \item \textbf{Matrix Factorization (MF)}. Matrix factorization of the incomplete matrix with rank 10. 
    \item \textbf{MICE}~\cite{white2011multiple}. Multiple imputation by chained equations. The algorithm fills the missing values iteratively until convergence. We use 10 nearest features and set the maximum iterations to 100.
    \item \textbf{BRITS}~\cite{cao2018brits}. BRITS has the similar bidirectional recurrent models as ours for time series imputation. It learns to impute only based on the time series features and does not consider the spatial information of the underlying graphs.
    \item \textbf{rGAIN}~\cite{cini2021filling}. A GAN based imputation model which is similar to SSGAN~\cite{miao2021generative}. % and is introduced in~\cite{cini2021filling}. 
    rGAIN can be regarded as an extension of GAIN~\cite{yoon2018gain} with bidirectional encoder and decoder.
    \item \textbf{SAITS}~\cite{du2023saits}. SAITS is a self-attention based methods with a weighted combination of two diagonally-masked self-attention blocks, which is trained by a joint optimization approach on imputation and reconstruction.
    \item \textbf{TimesNet}~\cite{wu2022timesnet}. TimesNet transforms the 1D time series into 2D space and present the intraperiod- and interperiod-variations simultaneously. Its inception-block is able to discover multiple periods and capture temporal 2D-variations from the transformed data.
    \item \textbf{GRIN}~\cite{cini2021filling}.  GRIN is a state-of-the-art model for MTS imputation with the relational information from a static and accurately known graph, which uses MPNN to build a spatio-temporal recurrent module and solves the problem in a bidirectional way.
    % \yc{footnote to appendixs}.
    \item \textbf{$\text{NET}^3$}~\cite{jing2021network}.  $\text{NET}^3$ is a recent work focusing on tensor time series learning and assumes that the tensor graphs are fixed and accurately known. %Results in their paper demonstrate its promising ability to handle the task of missing value recovery.
    % \hh{mention that it requires the graph info to be fixed and accurately known}
\end{enumerate}
NTS imputation (i.e., Problem~\ref{def:nts-impute}) also aims to solve the link prediction problem. We compare the performance of our method with following baselines:
\begin{enumerate}
    \item \textbf{VGAE}~\cite{kipf2016variational}. Vanilla variational graph autoencoder is the first work that brings VAE to graph learning, and has competitive performance on link prediction task over static graphs.
    \item \textbf{VGRNN}~\cite{hajiramezanali2019variational}. Variational graph recurrent neural networks extends VGAE to handle temporal information with the help of RNNs, and is a powerful baseline for the link prediction task on dynamic graphs.
\end{enumerate}
% vanilla variational graph autoencoders (VGAE)  and variational graph recurrent neural networks (VGRNN) \cite{hajiramezanali2019variational} over the link prediction task, which are powerful baselines for link prediction task for static graphs and dynamic graphs respectively.
% \yc{baseline consistency.}
\vspace{-1mm}
\subsubsection{Metrics}
We use \textit{mean absolute error} (MAE), \textit{mean squared error} (MSE) and \textit{mean relative error} (MRE) to evaluate the imputation performance of all models over missing features. For
% \yc{mark, ask Prof. link prediction or graph completion?}
the link prediction task, we use the Frobenius norm as the metric since all the edges are weighted. All the experiments are run with 5 different random seeds and the results are presented as mean $\pm$ standard deviation (std).
% Since we are the first to consider the missing edges problem in NTS data, we leave the comparison of graph reconstruction in the future.
% \vspace{-0.1mm}

\vspace{-3mm}
\subsection{Time Series Imputation Task
% Results and Analysis
}
Empirical results from Table~\ref{table:result1} and Table~\ref{table:result2} demonstrate that the proposed \vae\ outperforms all the baselines over the time series missing values prediction task in the NTS imputation problem. In particular, \vae\ achieves more than $10\%$ improvement on all the datasets compared with the best baselines. Especially, \vae\ has significant improvements over all the baselines over COVID-19 dataset where other neural network based models except TimesNet have even worse performance than traditional time series imputation methods such as MF and MICE.
% \hh{is the groundture basically a (near) straight line? if so, we may reconsider to put this figure -- the reviewer s might say: ha, this is a easy case and even in this easy case, your method (red line) is not fitting the groundtruth so well. if we want to put a figure, we want to show a non-trivial case that visually, our method clearly fits well with the groundtruth and it does better than baselines.}The prediction results of different baselines over COVID-19 dataset with $70\%$ missing rates over test data can be found in Figure \ref{fig:covid}.
% % and \vae\ has the predictions closest to the ground truth.
% \begin{figure}[ht]
% \includegraphics[width=0.49\textwidth]{imgs/covid_prediction.png}
% \caption{Different models' predictions of log-normalized COVID-19 infection cases in Illinois and New York from 10/15/2020 to 12/31/2020. Best viewed in color.}
% \label{fig:covid}
% \end{figure}
% It is clear that \vae\ can achieve better predictions results compared with other baselines.
% \hh{It is. It's is an informal expression and we should avoid it in papers. same changes in other places}
It is worth noting that, although equipped with modules to handle topological information from graphs, GRIN and $\text{NET}^3$ are less competitive than \vae\ when the graph is constantly changing and contains missing edges.
%\hh{do we have any insight on the possible reason for that?}\derek{Based on the experiments, it seems when graph contains missing edges, these two models will degrade to recurrent models and has similar performance as baselines without any spatial information such as BRITS and rGAIN. It is might due to the reason that missing edges bring extra noise (instead of useful information) to the model and they cannot handle such noise really well, which can also be demonstrated in the visualization we have in appendix.}%hh: thanks for the explanation. the current wording is probably sufficient.
%In PEMS04 dataset, GRIN and $\text{NET}^3$ can only achieves \yc{achieve} similar performance with BRITS and rGAIN which only models node time series without any graph information. Even worse, 
% \hh{consider to drop this line}For example, although being able to well reconstruct observed data, $\text{NET}^3$ suffers a lot from fluctuations as shown in Figure \ref{fig:covid}.
On the AQ36 dataset and the PeMS-SD dataset, they 
% \yc{even, need discussion}
bear worse performance compared to BRITS and rGAIN, which do not leverage any topological information. \vae\ outperforms BRITS and rGAIN by at least $12.92\%$ and $10.55\%$ on these two datasets respectively, which further indicates the effectiveness of our method. Although TimesNet is the strongest model over most of the datasets except AQ36, there still exists a large gap between its performance and \vae\ even with much more parameters (triple number of parameters of PoGeVon). The main reason \vae\ fluctuates (with a large std) on AQ36 dataset compared with traffic datasets is that AQ36 has fewer training samples (time steps), which brings more uncertainty for the model and results in larger differences of performances using different random seeds. 
% \vspace{-0.1mm}

\vspace{-2mm}
\subsection{Link Prediction Task}
\vspace{-2mm}
% \derek{Mark.}
% \yc{Move the link prediction subsection to the main content.}
% As for the link prediction task, the results are listed in Table \ref{table:linkpred} in the appendix. And it's obvious that \vae\ can out perform VGAE and VGRNN over all the traffic dataset. While for AQ36 dataset, \vae\ is less competitive due to the limited number of training data.
\begin{table}[ht]
\setlength{\abovecaptionskip}{3pt}
\caption{Performance comparison of the link prediction task in NTS imputation. Smaller is better.}
\begin{tabularx}{\columnwidth}{c|Y|Y|Y|Y} 
\toprule
Models & AQ36 & PeMS-BA & PeMS-LA & PeMS-SD  \\ 
\specialrule{1pt}{1pt}{3pt}
VGAE & \small{134.42} \footnotesize{$\pm\ 0.11$} & \small{431.94} \footnotesize{$\pm\ 2.31$} & \small{404.17} \footnotesize{$\pm\ 1.76$} & \small{399.78} \footnotesize{$\pm\ 1.29$} \\ 
\midrule
VGRNN & \small{133.92} \footnotesize{$\pm\ 0.29$} & \small{428.82} \footnotesize{$\pm\ 0.01$} & \small{402.30} \footnotesize{$\pm\ 0.90$} & \small{398.81} \footnotesize{$\pm\ 0.01$} \\ 
\midrule
\vae\  & \small{\textbf{95.42}} \footnotesize{$\pm\ \textbf{1.80}$} & \small{\textbf{148.44}} \footnotesize{$\pm\ \textbf{0.31}$} & \small{\textbf{168.05}} \footnotesize{$\pm\ \textbf{0.31}$} & \small{\textbf{185.86}} \footnotesize{$\pm\ \textbf{0.15}$} \\
\bottomrule
\end{tabularx}
\label{table:linkpred}
\vspace{-2mm}
\end{table}
VGAE and VGRNN were originally designed for link prediction over unweighted graphs. However, all the graphs are weighted in our NTS imputation settings, and thus, we modify these models correspondingly and apply the same Frobenius loss function we use in \vae\ to train them. All the results are listed in Table~\ref{table:linkpred}. Both baselines have relatively worse performance compared to \vae\ in all the datasets, and even using RNNs, VGRNN only gains minor improvement over VGAE. This indicates that both VGAE and VGRNN may not be able to handle the link prediction task over weighted dynamic graphs very well.
% \vspace{-0.1mm}

\subsection{Ablation Studies}
\vspace{-3mm}
\begin{table}[ht]
\setlength{\abovecaptionskip}{3pt}
\caption{Ablation study of \vae\ over AQ36 dataset on time series feature imputation. Smaller is better.}
\begin{tabularx}{\columnwidth}{c|Y|Y|Y} 
\toprule
Models & MAE & MSE & MRE  \\ 
\specialrule{1pt}{1pt}{3pt}
\vae\  & \small{\textbf{19.49}} \footnotesize{$\pm\ \textbf{1.10}$} & \small{\textbf{1213.47}} \footnotesize{$\pm\ \textbf{125.53}$} & \small{\textbf{0.26}} \footnotesize{$\pm\ \textbf{0.02}$} \\
\midrule
change RWR to SPD & \small{21.98} \footnotesize{$\pm\ 1.55$} & \small{1309.55} \footnotesize{$\pm\ 199.24$} & \small{0.33} \footnotesize{$\pm\ 0.02$} \\ 
\midrule
change RWR to RWPE Embeddings & \small{23.75} \footnotesize{$\pm 0.85$} & \small{1597.67} \footnotesize{$\pm 210.77$} & \small{0.32} \footnotesize{$\pm 0.01$} \\ 
\midrule
change RWR to PGNN Embeddings & \small{24.46} \footnotesize{$\pm 2.59 $} & \small{1625.19} \footnotesize{$\pm 393.25$} & \small{0.33} \footnotesize{$\pm 0.04$} \\ 
\midrule
% Remove 1-stage decoder & \small{60.87} \footnotesize{$\pm 0.50$} & \small{6517.95} \footnotesize{$\pm 88.11$} & \small{0.82} \footnotesize{$\pm 0.01$} \\ 
% \midrule
w/o link prediction in 2$^{\text{nd}}$ stage & \small{28.71} \footnotesize{$\pm\ 3.38$} & \small{2130.46} \footnotesize{$\pm\ 417.45$} & \small{0.38} \footnotesize{$\pm\ 0.05$} \\ 
\midrule
w/o self-attention in 3$^{\text{rd}}$ stage & \small{23.40} \footnotesize{$\pm\ 1.00$} & \small{1576.06} \footnotesize{$\pm\ 194.45$} & \small{0.31} \footnotesize{$\pm\ 0.01$} \\ 
\bottomrule
\end{tabularx}
\label{table:ablation}
\vspace{-3mm}
\end{table}

To evaluate the effectiveness of different components of our proposed method, we compare \vae\ with following variants: (1) Replace RWR node position embeddings with the shortest path distance (SPD) based node embeddings by calculating the distance between each node with anchor nodes. (2) Replace RWR node position embeddings with the RWPE node position embeddings from~\cite{dwivedi2021graph}. (3) Replace RWR node position embeddings with the PGNN node position embeddings from~\cite{you2019position}. (4) Remove the link prediction module in the 2$^{\text{nd}}$ stage prediction. (5) Remove the self-attention module in the 3$^{\text{rd}}$ stage prediction by replacing it with a linear layer. The results of the ablation study over AQ36 dataset are shown in Table~\ref{table:ablation}. As we can see, the proposed method \vae\ indeed performs the best which corroborates the necessity of all these components in the model.
% \vspace{-0.1mm}

\vspace{-1mm}
\subsubsection{Sensitivity Analysis}
% \noindent \textbf{Sensitivity Analysis.}
We conduct sensitivity analysis to study the effect brought by increasing the masking rates. We consider the following mask rates: $15\%$, $25\%$, $35\%$, $45\%$. In order to keep a reasonable edge missing rate, for each edge with either end node being masked, they have $70\%$ of chance being masked instead of using the setting from Eq. \eqref{data:construct}. 
\begin{figure}[ht]
\includegraphics[width=0.48\textwidth,trim = 2 2 2 2,clip]{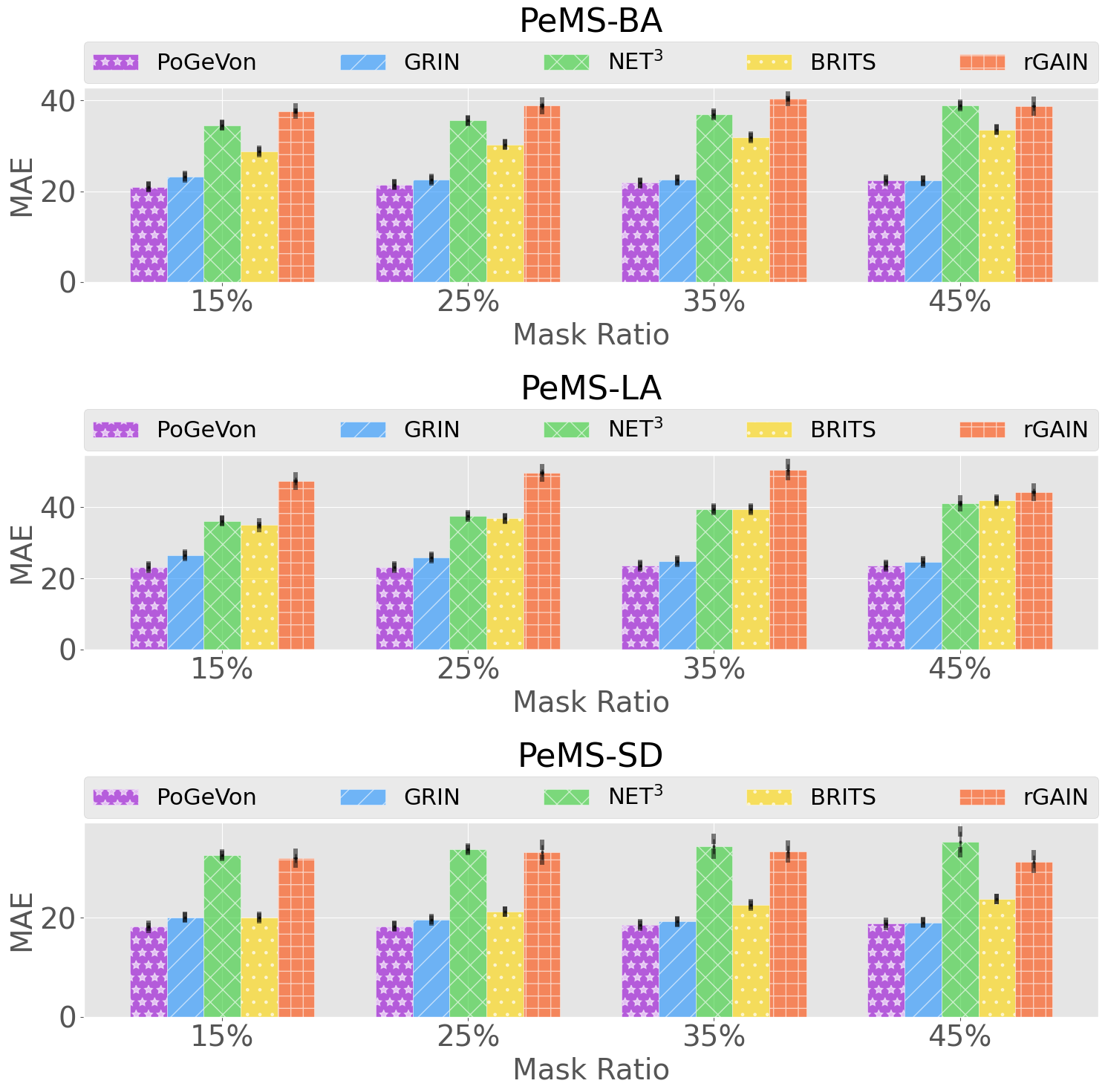}
\vspace{-7mm}
\caption{Sensitivity analysis for time series imputation with different masking rates on the traffic dataset. Lower is better. Best viewed in color.}
\label{fig:sensitivity}
\end{figure}
The results are shown in Figure \ref{fig:sensitivity}, in which the error bar demonstrates the standard deviation of MAE over 5 runs with different random seeds. The proposed \vae\ consistently outperforms all the baselines in these settings which further demonstrates the effectiveness and robustness of our method.
% \vspace{-0.1mm}
% \vspace{-0.12cm}
\section{Related Work}\label{sec:related}
In this section, we review the related works which can be categorized into two groups, including (1) multivariate time series imputation and (2) GNNs with relative position encodings.

\noindent \textbf{Multivariate Time Series Imputation.}
% \yc{subsection or just textit and textbf}
In addition to traditional methods such as ARIMA~\cite{box2015time} and K-Nearest Neighbors (KNN)~\cite{chen2000nearest}, deep learning models are widely adopted in recent years to solve the MTS imputation problem. BRITS~\cite{cao2018brits} is one of the most representative methods which uses bidirectional RNNs. There also exist
a wide range of
methods using deep generative models such as generative adversarial nets (GAN)~\cite{goodfellow2014generative} and VAE~\cite{kingma2013auto}. GAIN~\cite{yoon2018gain} is one of the earliest methods that use GAN to impute missing data, and later \cite{luo2018multivariate} applies GAN to the multivariate time series setting based on 2-stage imputation. $\text{E}^2\text{GAN}$~\cite{luo2019e2gan} is an end-to-end GAN and uses the noised 
% \hh{noisy?} \derek{it should be noised which is used by the original paper of $\text{E}^2\text{GAN}$}
compression and reconstruction strategy to generate more reasonable imputed values compared to previous works. SSGAN~\cite{miao2021generative}
proposes a novel method based on
GAN to handle missing data in partially labeled time series data. VAE is used in GP-VAE~\cite{fortuin2020gp} to solve the MTS imputation task with Gaussian process as the prior.

Other works handle MTS imputation problem from the perspective of spatial-temporal modeling, which takes the advantage of entities relations from the underlying graph. \cite{cai2015fast} is the first trial of using matrix factorization algorithm to recover missing values over MTS data with graph structures. More recently, GNNs have been used to capture the topological information in the MTS data. GRIN~\cite{cini2021filling} proposes a novel bidirectional message passing RNN with a spatial decoder to handle both the spatial and temporal information. SPIN~\cite{marisca2022learning} uses sparse spatiotemporal attention to capture inter-node and intra-node information for predicting missing values in MTS. $\text{NET}^3$~\cite{jing2021network} generalizes the problem to tensor time series where multiple modes of relation dependencies exist in the time series. It introduces a tensor GCN~\cite{kipf2016semi} to handle the tensor graphs and then proposes a tensor RNN to incorporate the temporal dynamics. One common limitation of all these methods is that they either ignore the topological information from graph or assume the graph is fixed and accurately known.
% \hh{1-2 sentences on the difference between our work and these works, e.g., all these work either ignore the graph information or assume the graph is fixed and accuractely known}
% It should be noted that MTS imputation with spatial-temporal modeling is a simplified version of NTS imputation problem since NTS data contains graph dynamics, while NTS can be regarded as a special case of tensor time series in which only one mode of the tensor graph is considered.

\noindent \textbf{GNNs with Relative Position Encodings.}
The expressive power of message-passing based GNNs has been proved to be bounded by 1-Weisfeiler-Lehman test (1-WL test) in~\cite{xu2018powerful}. Many follow-up works have been done to improve the expressive power of GNNs which go beyond 1-WL test, and position-aware graph neural networks (P-GNNs)~\cite{you2019position} is one of them. P-GNNs randomly picks sets of anchor nodes and learn a non-linear distance-weighted aggregation scheme over these anchor sets for each node. This relative position encodings for nodes are proved to be more expressive than regular GNNs.
% Based on P-GNNs, GraphReach \cite{nishad2020graphreach} selects anchor nodes through reachability estimations and enhances the semantic information of position-aware embeddings.
Distance Encoding~\cite{li2020distance} uses graph-distance measures between nodes as extra features and proves that it can distinguish node sets in most regular graphs in which message-passing based GNNs would fail.
% Recently,
\cite{dwivedi2021graph} proposes a novel module for learnable structural and positional encodings (LSPE) along with GNNs and Transformers~\cite{vaswani2017attention}, which generates more expressive node embeddings.
Recently, PEG~\cite{wang2022equivariant} is introduced for imposing permutation equivariance and stability to position encodings, which uses separate channels for node features and position features.
Compared with these existing methods, our proposed RWR-based position embedding %is more elegant and 
could capture more topological information from the entire graph, as our analysis in Section~\ref{sec:rwr} shows.
% capture more topological information compared with P-GNNs and GraphReach, and is more elegant since it does not require much computation as Distance Encoding and LSPE for each node.
% \hh{1-2 sentences on our new contributions compared with these works}
\vspace{-2mm}
\section{Conclusion}\label{sec:conclusion}
In this paper, we focus on solving networked time series imputation problem, which has two main challenges: (1) the graph is dynamic and missing edges exist, and (2) the node features time series contain missing values. To tackle these challenges, we propose \vae , a novel VAE model utilizing specially designed RWR-based position embeddings in the encoder. For the decoder, we design a 3-stage predictions to impute missing values in both features and structures complementarily.
% \yc{too long, break this sentence.}.
Experiments on a variety of real-world datasets show that \vae\ consistently outperforms strong baseline methods for the NTS imputation problem.
% \vspace{-0.15cm}
\section{Acknowledgements}
This work is supported by NSF (1947135, %hh-career-new
and 2134079 %MoDL
),
the NSF Program on Fairness in AI in collaboration with Amazon (1939725), % FAI
DARPA (HR001121C0165), %DARPA INCAS
NIFA (2020-67021-32799), % NSF-NIFA AIFarm@UIUC
DHS (\seqsplit{17STQAC00001-06-00}), %DHS COE privacy
ARO (W911NF2110088), %durip
the C3.ai Digital Transformation Institute, %C3.AI
MIT-IBM Watson AI Lab, %for papers with Yada Zhu
and IBM-Illinois Discovery Accelerator Institute. %IIDAI
The content of the information in this document does not necessarily reflect the position or the policy of the Government or Amazon, and no official endorsement should be inferred.  The U.S. Government is authorized to reproduce and distribute reprints for Government purposes notwithstanding any copyright notation here on.

%%
%% The next two lines define the bibliography style to be used, and
%% the bibliography file.
\bibliographystyle{ACM-Reference-Format}
\balance
\bibliography{references}

%%% -*-BibTeX-*-
%%% Do NOT edit. File created by BibTeX with style
%%% ACM-Reference-Format-Journals [18-Jan-2012].

\begin{thebibliography}{83}

%%% ====================================================================
%%% NOTE TO THE USER: you can override these defaults by providing
%%% customized versions of any of these macros before the \bibliography
%%% command.  Each of them MUST provide its own final punctuation,
%%% except for \shownote{}, \showDOI{}, and \showURL{}.  The latter two
%%% do not use final punctuation, in order to avoid confusing it with
%%% the Web address.
%%%
%%% To suppress output of a particular field, define its macro to expand
%%% to an empty string, or better, \unskip, like this:
%%%
%%% \newcommand{\showDOI}[1]{\unskip}   % LaTeX syntax
%%%
%%% \def \showDOI #1{\unskip}           % plain TeX syntax
%%%
%%% ====================================================================

\ifx \showCODEN    \undefined \def \showCODEN     #1{\unskip}     \fi
\ifx \showDOI      \undefined \def \showDOI       #1{#1}\fi
\ifx \showISBNx    \undefined \def \showISBNx     #1{\unskip}     \fi
\ifx \showISBNxiii \undefined \def \showISBNxiii  #1{\unskip}     \fi
\ifx \showISSN     \undefined \def \showISSN      #1{\unskip}     \fi
\ifx \showLCCN     \undefined \def \showLCCN      #1{\unskip}     \fi
\ifx \shownote     \undefined \def \shownote      #1{#1}          \fi
\ifx \showarticletitle \undefined \def \showarticletitle #1{#1}   \fi
\ifx \showURL      \undefined \def \showURL       {\relax}        \fi
% The following commands are used for tagged output and should be
% invisible to TeX
\providecommand\bibfield[2]{#2}
\providecommand\bibinfo[2]{#2}
\providecommand\natexlab[1]{#1}
\providecommand\showeprint[2][]{arXiv:#2}

\bibitem[Alemi et~al\mbox{.}(2016)]%
        {alemi2016deep}
\bibfield{author}{\bibinfo{person}{Alexander~A Alemi}, \bibinfo{person}{Ian
  Fischer}, \bibinfo{person}{Joshua~V Dillon}, {and} \bibinfo{person}{Kevin
  Murphy}.} \bibinfo{year}{2016}\natexlab{}.
\newblock \showarticletitle{Deep variational information bottleneck}.
\newblock \bibinfo{journal}{\emph{arXiv preprint arXiv:1612.00410}}
  (\bibinfo{year}{2016}).
\newblock


\bibitem[Box et~al\mbox{.}(2015)]%
        {box2015time}
\bibfield{author}{\bibinfo{person}{George~EP Box}, \bibinfo{person}{Gwilym~M
  Jenkins}, \bibinfo{person}{Gregory~C Reinsel}, {and} \bibinfo{person}{Greta~M
  Ljung}.} \bibinfo{year}{2015}\natexlab{}.
\newblock \bibinfo{booktitle}{\emph{Time series analysis: forecasting and
  control}}.
\newblock \bibinfo{publisher}{John Wiley \& Sons}.
\newblock


\bibitem[Burgess et~al\mbox{.}(2018)]%
        {burgess2018understanding}
\bibfield{author}{\bibinfo{person}{Christopher~P Burgess},
  \bibinfo{person}{Irina Higgins}, \bibinfo{person}{Arka Pal},
  \bibinfo{person}{Loic Matthey}, \bibinfo{person}{Nick Watters},
  \bibinfo{person}{Guillaume Desjardins}, {and} \bibinfo{person}{Alexander
  Lerchner}.} \bibinfo{year}{2018}\natexlab{}.
\newblock \showarticletitle{Understanding disentangling in $\beta$-VAE}.
\newblock \bibinfo{journal}{\emph{arXiv preprint arXiv:1804.03599}}
  (\bibinfo{year}{2018}).
\newblock


\bibitem[Cai et~al\mbox{.}(2015)]%
        {cai2015fast}
\bibfield{author}{\bibinfo{person}{Yongjie Cai}, \bibinfo{person}{Hanghang
  Tong}, \bibinfo{person}{Wei Fan}, {and} \bibinfo{person}{Ping Ji}.}
  \bibinfo{year}{2015}\natexlab{}.
\newblock \showarticletitle{Fast mining of a network of coevolving time
  series}. In \bibinfo{booktitle}{\emph{Proceedings of the 2015 SIAM
  International Conference on Data Mining}}. SIAM, \bibinfo{pages}{298--306}.
\newblock


\bibitem[Cao et~al\mbox{.}(2018)]%
        {cao2018brits}
\bibfield{author}{\bibinfo{person}{Wei Cao}, \bibinfo{person}{Dong Wang},
  \bibinfo{person}{Jian Li}, \bibinfo{person}{Hao Zhou}, \bibinfo{person}{Lei
  Li}, {and} \bibinfo{person}{Yitan Li}.} \bibinfo{year}{2018}\natexlab{}.
\newblock \showarticletitle{Brits: Bidirectional recurrent imputation for time
  series}.
\newblock \bibinfo{journal}{\emph{Advances in neural information processing
  systems}}  \bibinfo{volume}{31} (\bibinfo{year}{2018}).
\newblock


\bibitem[Chen et~al\mbox{.}(2001)]%
        {chen2001freeway}
\bibfield{author}{\bibinfo{person}{Chao Chen}, \bibinfo{person}{Karl Petty},
  \bibinfo{person}{Alexander Skabardonis}, \bibinfo{person}{Pravin Varaiya},
  {and} \bibinfo{person}{Zhanfeng Jia}.} \bibinfo{year}{2001}\natexlab{}.
\newblock \showarticletitle{Freeway performance measurement system: mining loop
  detector data}.
\newblock \bibinfo{journal}{\emph{Transportation Research Record}}
  \bibinfo{volume}{1748}, \bibinfo{number}{1} (\bibinfo{year}{2001}),
  \bibinfo{pages}{96--102}.
\newblock


\bibitem[Chen and Shao(2000)]%
        {chen2000nearest}
\bibfield{author}{\bibinfo{person}{Jiahua Chen} {and} \bibinfo{person}{Jun
  Shao}.} \bibinfo{year}{2000}\natexlab{}.
\newblock \showarticletitle{Nearest neighbor imputation for survey data}.
\newblock \bibinfo{journal}{\emph{Journal of official statistics}}
  \bibinfo{volume}{16}, \bibinfo{number}{2} (\bibinfo{year}{2000}),
  \bibinfo{pages}{113}.
\newblock


\bibitem[Chen et~al\mbox{.}(2021)]%
        {chen2021z}
\bibfield{author}{\bibinfo{person}{Yuzhou Chen}, \bibinfo{person}{Ignacio
  Segovia}, {and} \bibinfo{person}{Yulia~R Gel}.}
  \bibinfo{year}{2021}\natexlab{}.
\newblock \showarticletitle{Z-GCNETs: time zigzags at graph convolutional
  networks for time series forecasting}. In
  \bibinfo{booktitle}{\emph{International Conference on Machine Learning}}.
  PMLR, \bibinfo{pages}{1684--1694}.
\newblock


\bibitem[Chien et~al\mbox{.}(2020)]%
        {chien2020adaptive}
\bibfield{author}{\bibinfo{person}{Eli Chien}, \bibinfo{person}{Jianhao Peng},
  \bibinfo{person}{Pan Li}, {and} \bibinfo{person}{Olgica Milenkovic}.}
  \bibinfo{year}{2020}\natexlab{}.
\newblock \showarticletitle{Adaptive universal generalized pagerank graph
  neural network}.
\newblock \bibinfo{journal}{\emph{arXiv preprint arXiv:2006.07988}}
  (\bibinfo{year}{2020}).
\newblock


\bibitem[Chung et~al\mbox{.}(2014)]%
        {chung2014empirical}
\bibfield{author}{\bibinfo{person}{Junyoung Chung}, \bibinfo{person}{Caglar
  Gulcehre}, \bibinfo{person}{KyungHyun Cho}, {and} \bibinfo{person}{Yoshua
  Bengio}.} \bibinfo{year}{2014}\natexlab{}.
\newblock \showarticletitle{Empirical evaluation of gated recurrent neural
  networks on sequence modeling}.
\newblock \bibinfo{journal}{\emph{arXiv preprint arXiv:1412.3555}}
  (\bibinfo{year}{2014}).
\newblock


\bibitem[Cini et~al\mbox{.}(2021)]%
        {cini2021filling}
\bibfield{author}{\bibinfo{person}{Andrea Cini}, \bibinfo{person}{Ivan
  Marisca}, {and} \bibinfo{person}{Cesare Alippi}.}
  \bibinfo{year}{2021}\natexlab{}.
\newblock \showarticletitle{Filling the g\_ap\_s: Multivariate time series
  imputation by graph neural networks}.
\newblock \bibinfo{journal}{\emph{arXiv preprint arXiv:2108.00298}}
  (\bibinfo{year}{2021}).
\newblock


\bibitem[Collier et~al\mbox{.}(2020)]%
        {collier2020vaes}
\bibfield{author}{\bibinfo{person}{Mark Collier}, \bibinfo{person}{Alfredo
  Nazabal}, {and} \bibinfo{person}{Christopher~KI Williams}.}
  \bibinfo{year}{2020}\natexlab{}.
\newblock \showarticletitle{VAEs in the presence of missing data}.
\newblock \bibinfo{journal}{\emph{arXiv preprint arXiv:2006.05301}}
  (\bibinfo{year}{2020}).
\newblock


\bibitem[Ding et~al\mbox{.}(2015)]%
        {ding2015deep}
\bibfield{author}{\bibinfo{person}{Xiao Ding}, \bibinfo{person}{Yue Zhang},
  \bibinfo{person}{Ting Liu}, {and} \bibinfo{person}{Junwen Duan}.}
  \bibinfo{year}{2015}\natexlab{}.
\newblock \showarticletitle{Deep learning for event-driven stock prediction}.
  In \bibinfo{booktitle}{\emph{Twenty-fourth international joint conference on
  artificial intelligence}}.
\newblock


\bibitem[Doersch(2016)]%
        {doersch2016tutorial}
\bibfield{author}{\bibinfo{person}{Carl Doersch}.}
  \bibinfo{year}{2016}\natexlab{}.
\newblock \showarticletitle{Tutorial on variational autoencoders}.
\newblock \bibinfo{journal}{\emph{arXiv preprint arXiv:1606.05908}}
  (\bibinfo{year}{2016}).
\newblock


\bibitem[Du et~al\mbox{.}(2023)]%
        {du2023saits}
\bibfield{author}{\bibinfo{person}{Wenjie Du}, \bibinfo{person}{David
  C{\^o}t{\'e}}, {and} \bibinfo{person}{Yan Liu}.}
  \bibinfo{year}{2023}\natexlab{}.
\newblock \showarticletitle{Saits: Self-attention-based imputation for time
  series}.
\newblock \bibinfo{journal}{\emph{Expert Systems with Applications}}
  \bibinfo{volume}{219} (\bibinfo{year}{2023}), \bibinfo{pages}{119619}.
\newblock


\bibitem[Dwivedi et~al\mbox{.}(2021)]%
        {dwivedi2021graph}
\bibfield{author}{\bibinfo{person}{Vijay~Prakash Dwivedi},
  \bibinfo{person}{Anh~Tuan Luu}, \bibinfo{person}{Thomas Laurent},
  \bibinfo{person}{Yoshua Bengio}, {and} \bibinfo{person}{Xavier Bresson}.}
  \bibinfo{year}{2021}\natexlab{}.
\newblock \showarticletitle{Graph neural networks with learnable structural and
  positional representations}.
\newblock \bibinfo{journal}{\emph{arXiv preprint arXiv:2110.07875}}
  (\bibinfo{year}{2021}).
\newblock


\bibitem[Fang and Wang(2020)]%
        {fang2020time}
\bibfield{author}{\bibinfo{person}{Chenguang Fang} {and} \bibinfo{person}{Chen
  Wang}.} \bibinfo{year}{2020}\natexlab{}.
\newblock \showarticletitle{Time series data imputation: A survey on deep
  learning approaches}.
\newblock \bibinfo{journal}{\emph{arXiv preprint arXiv:2011.11347}}
  (\bibinfo{year}{2020}).
\newblock


\bibitem[Fortuin et~al\mbox{.}(2020)]%
        {fortuin2020gp}
\bibfield{author}{\bibinfo{person}{Vincent Fortuin}, \bibinfo{person}{Dmitry
  Baranchuk}, \bibinfo{person}{Gunnar R{\"a}tsch}, {and}
  \bibinfo{person}{Stephan Mandt}.} \bibinfo{year}{2020}\natexlab{}.
\newblock \showarticletitle{Gp-vae: Deep probabilistic time series imputation}.
  In \bibinfo{booktitle}{\emph{International conference on artificial
  intelligence and statistics}}. PMLR, \bibinfo{pages}{1651--1661}.
\newblock


\bibitem[Fu et~al\mbox{.}(2022)]%
        {DBLP:conf/kdd/FuFMTH22}
\bibfield{author}{\bibinfo{person}{Dongqi Fu}, \bibinfo{person}{Liri Fang},
  \bibinfo{person}{Ross Maciejewski}, \bibinfo{person}{Vetle~I. Torvik}, {and}
  \bibinfo{person}{Jingrui He}.} \bibinfo{year}{2022}\natexlab{}.
\newblock \showarticletitle{Meta-Learned Metrics over Multi-Evolution Temporal
  Graphs}. In \bibinfo{booktitle}{\emph{{KDD} 2022}}.
\newblock


\bibitem[Fu and He(2021)]%
        {DBLP:conf/sigir/FuH21}
\bibfield{author}{\bibinfo{person}{Dongqi Fu} {and} \bibinfo{person}{Jingrui
  He}.} \bibinfo{year}{2021}\natexlab{}.
\newblock \showarticletitle{{SDG:} {A} Simplified and Dynamic Graph Neural
  Network}. In \bibinfo{booktitle}{\emph{{SIGIR} 2021}}.
\newblock


\bibitem[Fu et~al\mbox{.}(2020)]%
        {DBLP:conf/kdd/FuZH20}
\bibfield{author}{\bibinfo{person}{Dongqi Fu}, \bibinfo{person}{Dawei Zhou},
  {and} \bibinfo{person}{Jingrui He}.} \bibinfo{year}{2020}\natexlab{}.
\newblock \showarticletitle{Local Motif Clustering on Time-Evolving Graphs}. In
  \bibinfo{booktitle}{\emph{{KDD} 2020}}.
\newblock


\bibitem[Gilmer et~al\mbox{.}(2017)]%
        {gilmer2017neural}
\bibfield{author}{\bibinfo{person}{Justin Gilmer}, \bibinfo{person}{Samuel~S
  Schoenholz}, \bibinfo{person}{Patrick~F Riley}, \bibinfo{person}{Oriol
  Vinyals}, {and} \bibinfo{person}{George~E Dahl}.}
  \bibinfo{year}{2017}\natexlab{}.
\newblock \showarticletitle{Neural message passing for quantum chemistry}. In
  \bibinfo{booktitle}{\emph{International conference on machine learning}}.
  PMLR, \bibinfo{pages}{1263--1272}.
\newblock


\bibitem[Goodfellow et~al\mbox{.}(2014)]%
        {goodfellow2014generative}
\bibfield{author}{\bibinfo{person}{Ian Goodfellow}, \bibinfo{person}{Jean
  Pouget-Abadie}, \bibinfo{person}{Mehdi Mirza}, \bibinfo{person}{Bing Xu},
  \bibinfo{person}{David Warde-Farley}, \bibinfo{person}{Sherjil Ozair},
  \bibinfo{person}{Aaron Courville}, {and} \bibinfo{person}{Yoshua Bengio}.}
  \bibinfo{year}{2014}\natexlab{}.
\newblock \showarticletitle{Generative adversarial nets}.
\newblock \bibinfo{journal}{\emph{Advances in neural information processing
  systems}}  \bibinfo{volume}{27} (\bibinfo{year}{2014}).
\newblock


\bibitem[Hajiramezanali et~al\mbox{.}(2019)]%
        {hajiramezanali2019variational}
\bibfield{author}{\bibinfo{person}{Ehsan Hajiramezanali},
  \bibinfo{person}{Arman Hasanzadeh}, \bibinfo{person}{Krishna Narayanan},
  \bibinfo{person}{Nick Duffield}, \bibinfo{person}{Mingyuan Zhou}, {and}
  \bibinfo{person}{Xiaoning Qian}.} \bibinfo{year}{2019}\natexlab{}.
\newblock \showarticletitle{Variational graph recurrent neural networks}.
\newblock \bibinfo{journal}{\emph{Advances in neural information processing
  systems}}  \bibinfo{volume}{32} (\bibinfo{year}{2019}).
\newblock


\bibitem[Higgins et~al\mbox{.}(2016)]%
        {higgins2016beta}
\bibfield{author}{\bibinfo{person}{Irina Higgins}, \bibinfo{person}{Loic
  Matthey}, \bibinfo{person}{Arka Pal}, \bibinfo{person}{Christopher Burgess},
  \bibinfo{person}{Xavier Glorot}, \bibinfo{person}{Matthew Botvinick},
  \bibinfo{person}{Shakir Mohamed}, {and} \bibinfo{person}{Alexander
  Lerchner}.} \bibinfo{year}{2016}\natexlab{}.
\newblock \showarticletitle{beta-vae: Learning basic visual concepts with a
  constrained variational framework}.
\newblock  (\bibinfo{year}{2016}).
\newblock


\bibitem[Ivanov et~al\mbox{.}(2018)]%
        {ivanov2018variational}
\bibfield{author}{\bibinfo{person}{Oleg Ivanov}, \bibinfo{person}{Michael
  Figurnov}, {and} \bibinfo{person}{Dmitry Vetrov}.}
  \bibinfo{year}{2018}\natexlab{}.
\newblock \showarticletitle{Variational autoencoder with arbitrary
  conditioning}.
\newblock \bibinfo{journal}{\emph{arXiv preprint arXiv:1806.02382}}
  (\bibinfo{year}{2018}).
\newblock


\bibitem[Jin et~al\mbox{.}(2020)]%
        {jin2020recurrent}
\bibfield{author}{\bibinfo{person}{Woojeong Jin}, \bibinfo{person}{Meng Qu},
  \bibinfo{person}{Xisen Jin}, {and} \bibinfo{person}{Xiang Ren}.}
  \bibinfo{year}{2020}\natexlab{}.
\newblock \showarticletitle{Recurrent Event Network: Autoregressive Structure
  Inferenceover Temporal Knowledge Graphs}. In
  \bibinfo{booktitle}{\emph{Proceedings of the 2020 Conference on Empirical
  Methods in Natural Language Processing (EMNLP)}}.
  \bibinfo{pages}{6669--6683}.
\newblock


\bibitem[Jing et~al\mbox{.}(2021a)]%
        {jing2021hdmi}
\bibfield{author}{\bibinfo{person}{Baoyu Jing}, \bibinfo{person}{Chanyoung
  Park}, {and} \bibinfo{person}{Hanghang Tong}.}
  \bibinfo{year}{2021}\natexlab{a}.
\newblock \showarticletitle{Hdmi: High-order deep multiplex infomax}. In
  \bibinfo{booktitle}{\emph{Proceedings of the Web Conference 2021}}.
  \bibinfo{pages}{2414--2424}.
\newblock


\bibitem[Jing et~al\mbox{.}(2021b)]%
        {jing2021network}
\bibfield{author}{\bibinfo{person}{Baoyu Jing}, \bibinfo{person}{Hanghang
  Tong}, {and} \bibinfo{person}{Yada Zhu}.} \bibinfo{year}{2021}\natexlab{b}.
\newblock \showarticletitle{Network of tensor time series}. In
  \bibinfo{booktitle}{\emph{Proceedings of the Web Conference 2021}}.
  \bibinfo{pages}{2425--2437}.
\newblock


\bibitem[Jing et~al\mbox{.}(2022)]%
        {jing2022retrieval}
\bibfield{author}{\bibinfo{person}{Baoyu Jing}, \bibinfo{person}{Si Zhang},
  \bibinfo{person}{Yada Zhu}, \bibinfo{person}{Bin Peng},
  \bibinfo{person}{Kaiyu Guan}, \bibinfo{person}{Andrew Margenot}, {and}
  \bibinfo{person}{Hanghang Tong}.} \bibinfo{year}{2022}\natexlab{}.
\newblock \showarticletitle{Retrieval Based Time Series Forecasting}.
\newblock \bibinfo{journal}{\emph{arXiv preprint arXiv:2209.13525}}
  (\bibinfo{year}{2022}).
\newblock


\bibitem[Kapoor et~al\mbox{.}(2020)]%
        {kapoor2020examining}
\bibfield{author}{\bibinfo{person}{Amol Kapoor}, \bibinfo{person}{Xue Ben},
  \bibinfo{person}{Luyang Liu}, \bibinfo{person}{Bryan Perozzi},
  \bibinfo{person}{Matt Barnes}, \bibinfo{person}{Martin Blais}, {and}
  \bibinfo{person}{Shawn O'Banion}.} \bibinfo{year}{2020}\natexlab{}.
\newblock \showarticletitle{Examining covid-19 forecasting using
  spatio-temporal graph neural networks}.
\newblock \bibinfo{journal}{\emph{arXiv preprint arXiv:2007.03113}}
  (\bibinfo{year}{2020}).
\newblock


\bibitem[Katragadda et~al\mbox{.}(2022)]%
        {katragadda2022examining}
\bibfield{author}{\bibinfo{person}{Satya Katragadda},
  \bibinfo{person}{Ravi~Teja Bhupatiraju}, \bibinfo{person}{Vijay Raghavan},
  \bibinfo{person}{Ziad Ashkar}, {and} \bibinfo{person}{Raju Gottumukkala}.}
  \bibinfo{year}{2022}\natexlab{}.
\newblock \showarticletitle{Examining the COVID-19 case growth rate due to
  visitor vs. local mobility in the United States using machine learning}.
\newblock \bibinfo{journal}{\emph{Scientific Reports}} \bibinfo{volume}{12},
  \bibinfo{number}{1} (\bibinfo{year}{2022}), \bibinfo{pages}{1--12}.
\newblock


\bibitem[Kazemi et~al\mbox{.}(2019)]%
        {kazemi2019time2vec}
\bibfield{author}{\bibinfo{person}{Seyed~Mehran Kazemi},
  \bibinfo{person}{Rishab Goel}, \bibinfo{person}{Sepehr Eghbali},
  \bibinfo{person}{Janahan Ramanan}, \bibinfo{person}{Jaspreet Sahota},
  \bibinfo{person}{Sanjay Thakur}, \bibinfo{person}{Stella Wu},
  \bibinfo{person}{Cathal Smyth}, \bibinfo{person}{Pascal Poupart}, {and}
  \bibinfo{person}{Marcus Brubaker}.} \bibinfo{year}{2019}\natexlab{}.
\newblock \showarticletitle{Time2vec: Learning a vector representation of
  time}.
\newblock \bibinfo{journal}{\emph{arXiv preprint arXiv:1907.05321}}
  (\bibinfo{year}{2019}).
\newblock


\bibitem[Kingma and Ba(2014)]%
        {kingma2014adam}
\bibfield{author}{\bibinfo{person}{Diederik~P Kingma} {and}
  \bibinfo{person}{Jimmy Ba}.} \bibinfo{year}{2014}\natexlab{}.
\newblock \showarticletitle{Adam: A method for stochastic optimization}.
\newblock \bibinfo{journal}{\emph{arXiv preprint arXiv:1412.6980}}
  (\bibinfo{year}{2014}).
\newblock


\bibitem[Kingma and Welling(2013)]%
        {kingma2013auto}
\bibfield{author}{\bibinfo{person}{Diederik~P Kingma} {and}
  \bibinfo{person}{Max Welling}.} \bibinfo{year}{2013}\natexlab{}.
\newblock \showarticletitle{Auto-encoding variational bayes}.
\newblock \bibinfo{journal}{\emph{arXiv preprint arXiv:1312.6114}}
  (\bibinfo{year}{2013}).
\newblock


\bibitem[Kipf and Welling(2016a)]%
        {kipf2016semi}
\bibfield{author}{\bibinfo{person}{Thomas~N Kipf} {and} \bibinfo{person}{Max
  Welling}.} \bibinfo{year}{2016}\natexlab{a}.
\newblock \showarticletitle{Semi-supervised classification with graph
  convolutional networks}.
\newblock \bibinfo{journal}{\emph{arXiv preprint arXiv:1609.02907}}
  (\bibinfo{year}{2016}).
\newblock


\bibitem[Kipf and Welling(2016b)]%
        {kipf2016variational}
\bibfield{author}{\bibinfo{person}{Thomas~N Kipf} {and} \bibinfo{person}{Max
  Welling}.} \bibinfo{year}{2016}\natexlab{b}.
\newblock \showarticletitle{Variational graph auto-encoders}.
\newblock \bibinfo{journal}{\emph{arXiv preprint arXiv:1611.07308}}
  (\bibinfo{year}{2016}).
\newblock


\bibitem[Kitaev et~al\mbox{.}(2020)]%
        {kitaev2020reformer}
\bibfield{author}{\bibinfo{person}{Nikita Kitaev}, \bibinfo{person}{{\L}ukasz
  Kaiser}, {and} \bibinfo{person}{Anselm Levskaya}.}
  \bibinfo{year}{2020}\natexlab{}.
\newblock \showarticletitle{Reformer: The efficient transformer}.
\newblock \bibinfo{journal}{\emph{arXiv preprint arXiv:2001.04451}}
  (\bibinfo{year}{2020}).
\newblock


\bibitem[Knight et~al\mbox{.}(2016)]%
        {knight2016modelling}
\bibfield{author}{\bibinfo{person}{MI Knight}, \bibinfo{person}{MA Nunes},
  {and} \bibinfo{person}{GP Nason}.} \bibinfo{year}{2016}\natexlab{}.
\newblock \showarticletitle{Modelling, detrending and decorrelation of network
  time series}.
\newblock \bibinfo{journal}{\emph{arXiv preprint arXiv:1603.03221}}
  (\bibinfo{year}{2016}).
\newblock


\bibitem[Li et~al\mbox{.}(2020a)]%
        {li2020distance}
\bibfield{author}{\bibinfo{person}{Pan Li}, \bibinfo{person}{Yanbang Wang},
  \bibinfo{person}{Hongwei Wang}, {and} \bibinfo{person}{Jure Leskovec}.}
  \bibinfo{year}{2020}\natexlab{a}.
\newblock \showarticletitle{Distance encoding: Design provably more powerful
  neural networks for graph representation learning}.
\newblock \bibinfo{journal}{\emph{Advances in Neural Information Processing
  Systems}}  \bibinfo{volume}{33} (\bibinfo{year}{2020}),
  \bibinfo{pages}{4465--4478}.
\newblock


\bibitem[Li et~al\mbox{.}(2018)]%
        {li2018deeper}
\bibfield{author}{\bibinfo{person}{Qimai Li}, \bibinfo{person}{Zhichao Han},
  {and} \bibinfo{person}{Xiao-Ming Wu}.} \bibinfo{year}{2018}\natexlab{}.
\newblock \showarticletitle{Deeper insights into graph convolutional networks
  for semi-supervised learning}. In \bibinfo{booktitle}{\emph{Thirty-Second
  AAAI conference on artificial intelligence}}.
\newblock


\bibitem[Li et~al\mbox{.}(2020b)]%
        {li2020dynamic}
\bibfield{author}{\bibinfo{person}{Xiaohan Li}, \bibinfo{person}{Mengqi Zhang},
  \bibinfo{person}{Shu Wu}, \bibinfo{person}{Zheng Liu}, \bibinfo{person}{Liang
  Wang}, {and} \bibinfo{person}{S~Yu Philip}.}
  \bibinfo{year}{2020}\natexlab{b}.
\newblock \showarticletitle{Dynamic graph collaborative filtering}. In
  \bibinfo{booktitle}{\emph{2020 IEEE International Conference on Data Mining
  (ICDM)}}. IEEE, \bibinfo{pages}{322--331}.
\newblock


\bibitem[Li et~al\mbox{.}(2017)]%
        {li2017diffusion}
\bibfield{author}{\bibinfo{person}{Yaguang Li}, \bibinfo{person}{Rose Yu},
  \bibinfo{person}{Cyrus Shahabi}, {and} \bibinfo{person}{Yan Liu}.}
  \bibinfo{year}{2017}\natexlab{}.
\newblock \showarticletitle{Diffusion convolutional recurrent neural network:
  Data-driven traffic forecasting}.
\newblock \bibinfo{journal}{\emph{arXiv preprint arXiv:1707.01926}}
  (\bibinfo{year}{2017}).
\newblock


\bibitem[Lim et~al\mbox{.}(2021)]%
        {lim2021large}
\bibfield{author}{\bibinfo{person}{Derek Lim}, \bibinfo{person}{Felix Hohne},
  \bibinfo{person}{Xiuyu Li}, \bibinfo{person}{Sijia~Linda Huang},
  \bibinfo{person}{Vaishnavi Gupta}, \bibinfo{person}{Omkar Bhalerao}, {and}
  \bibinfo{person}{Ser~Nam Lim}.} \bibinfo{year}{2021}\natexlab{}.
\newblock \showarticletitle{Large scale learning on non-homophilous graphs: New
  benchmarks and strong simple methods}.
\newblock \bibinfo{journal}{\emph{Advances in Neural Information Processing
  Systems}}  \bibinfo{volume}{34} (\bibinfo{year}{2021}),
  \bibinfo{pages}{20887--20902}.
\newblock


\bibitem[Loshchilov and Hutter(2016)]%
        {loshchilov2016sgdr}
\bibfield{author}{\bibinfo{person}{Ilya Loshchilov} {and}
  \bibinfo{person}{Frank Hutter}.} \bibinfo{year}{2016}\natexlab{}.
\newblock \showarticletitle{Sgdr: Stochastic gradient descent with warm
  restarts}.
\newblock \bibinfo{journal}{\emph{arXiv preprint arXiv:1608.03983}}
  (\bibinfo{year}{2016}).
\newblock


\bibitem[Luo et~al\mbox{.}(2018)]%
        {luo2018multivariate}
\bibfield{author}{\bibinfo{person}{Yonghong Luo}, \bibinfo{person}{Xiangrui
  Cai}, \bibinfo{person}{Ying Zhang}, \bibinfo{person}{Jun Xu},
  {et~al\mbox{.}}} \bibinfo{year}{2018}\natexlab{}.
\newblock \showarticletitle{Multivariate time series imputation with generative
  adversarial networks}.
\newblock \bibinfo{journal}{\emph{Advances in neural information processing
  systems}}  \bibinfo{volume}{31} (\bibinfo{year}{2018}).
\newblock


\bibitem[Luo et~al\mbox{.}(2019)]%
        {luo2019e2gan}
\bibfield{author}{\bibinfo{person}{Yonghong Luo}, \bibinfo{person}{Ying Zhang},
  \bibinfo{person}{Xiangrui Cai}, {and} \bibinfo{person}{Xiaojie Yuan}.}
  \bibinfo{year}{2019}\natexlab{}.
\newblock \showarticletitle{E2gan: End-to-end generative adversarial network
  for multivariate time series imputation}. In
  \bibinfo{booktitle}{\emph{Proceedings of the 28th international joint
  conference on artificial intelligence}}. AAAI Press,
  \bibinfo{pages}{3094--3100}.
\newblock


\bibitem[Marisca et~al\mbox{.}(2022)]%
        {marisca2022learning}
\bibfield{author}{\bibinfo{person}{Ivan Marisca}, \bibinfo{person}{Andrea
  Cini}, {and} \bibinfo{person}{Cesare Alippi}.}
  \bibinfo{year}{2022}\natexlab{}.
\newblock \showarticletitle{Learning to Reconstruct Missing Data from
  Spatiotemporal Graphs with Sparse Observations}.
\newblock \bibinfo{journal}{\emph{arXiv preprint arXiv:2205.13479}}
  (\bibinfo{year}{2022}).
\newblock


\bibitem[McGill(1954)]%
        {mcgill1954multivariate}
\bibfield{author}{\bibinfo{person}{William McGill}.}
  \bibinfo{year}{1954}\natexlab{}.
\newblock \showarticletitle{Multivariate information transmission}.
\newblock \bibinfo{journal}{\emph{Transactions of the IRE Professional Group on
  Information Theory}} \bibinfo{volume}{4}, \bibinfo{number}{4}
  (\bibinfo{year}{1954}), \bibinfo{pages}{93--111}.
\newblock


\bibitem[Miao et~al\mbox{.}(2021)]%
        {miao2021generative}
\bibfield{author}{\bibinfo{person}{Xiaoye Miao}, \bibinfo{person}{Yangyang Wu},
  \bibinfo{person}{Jun Wang}, \bibinfo{person}{Yunjun Gao},
  \bibinfo{person}{Xudong Mao}, {and} \bibinfo{person}{Jianwei Yin}.}
  \bibinfo{year}{2021}\natexlab{}.
\newblock \showarticletitle{Generative semi-supervised learning for
  multivariate time series imputation}. In
  \bibinfo{booktitle}{\emph{Proceedings of the AAAI conference on artificial
  intelligence}}, Vol.~\bibinfo{volume}{35}. \bibinfo{pages}{8983--8991}.
\newblock


\bibitem[Nazabal et~al\mbox{.}(2020)]%
        {nazabal2020handling}
\bibfield{author}{\bibinfo{person}{Alfredo Nazabal}, \bibinfo{person}{Pablo~M
  Olmos}, \bibinfo{person}{Zoubin Ghahramani}, {and} \bibinfo{person}{Isabel
  Valera}.} \bibinfo{year}{2020}\natexlab{}.
\newblock \showarticletitle{Handling incomplete heterogeneous data using vaes}.
\newblock \bibinfo{journal}{\emph{Pattern Recognition}}  \bibinfo{volume}{107}
  (\bibinfo{year}{2020}), \bibinfo{pages}{107501}.
\newblock


\bibitem[Panagopoulos et~al\mbox{.}(2021)]%
        {panagopoulos2021transfer}
\bibfield{author}{\bibinfo{person}{George Panagopoulos},
  \bibinfo{person}{Giannis Nikolentzos}, {and} \bibinfo{person}{Michalis
  Vazirgiannis}.} \bibinfo{year}{2021}\natexlab{}.
\newblock \showarticletitle{Transfer graph neural networks for pandemic
  forecasting}. In \bibinfo{booktitle}{\emph{Proceedings of the AAAI Conference
  on Artificial Intelligence}}, Vol.~\bibinfo{volume}{35}.
  \bibinfo{pages}{4838--4845}.
\newblock


\bibitem[Paszke et~al\mbox{.}(2019)]%
        {paszke2019pytorch}
\bibfield{author}{\bibinfo{person}{Adam Paszke}, \bibinfo{person}{Sam Gross},
  \bibinfo{person}{Francisco Massa}, \bibinfo{person}{Adam Lerer},
  \bibinfo{person}{James Bradbury}, \bibinfo{person}{Gregory Chanan},
  \bibinfo{person}{Trevor Killeen}, \bibinfo{person}{Zeming Lin},
  \bibinfo{person}{Natalia Gimelshein}, \bibinfo{person}{Luca Antiga},
  {et~al\mbox{.}}} \bibinfo{year}{2019}\natexlab{}.
\newblock \showarticletitle{Pytorch: An imperative style, high-performance deep
  learning library}.
\newblock \bibinfo{journal}{\emph{Advances in neural information processing
  systems}}  \bibinfo{volume}{32} (\bibinfo{year}{2019}).
\newblock


\bibitem[Peng et~al\mbox{.}(2020)]%
        {peng2020graph}
\bibfield{author}{\bibinfo{person}{Zhen Peng}, \bibinfo{person}{Wenbing Huang},
  \bibinfo{person}{Minnan Luo}, \bibinfo{person}{Qinghua Zheng},
  \bibinfo{person}{Yu Rong}, \bibinfo{person}{Tingyang Xu}, {and}
  \bibinfo{person}{Junzhou Huang}.} \bibinfo{year}{2020}\natexlab{}.
\newblock \showarticletitle{Graph representation learning via graphical mutual
  information maximization}. In \bibinfo{booktitle}{\emph{Proceedings of The
  Web Conference 2020}}. \bibinfo{pages}{259--270}.
\newblock


\bibitem[Rossi et~al\mbox{.}(2020)]%
        {rossi2020temporal}
\bibfield{author}{\bibinfo{person}{Emanuele Rossi}, \bibinfo{person}{Ben
  Chamberlain}, \bibinfo{person}{Fabrizio Frasca}, \bibinfo{person}{Davide
  Eynard}, \bibinfo{person}{Federico Monti}, {and} \bibinfo{person}{Michael
  Bronstein}.} \bibinfo{year}{2020}\natexlab{}.
\newblock \showarticletitle{Temporal graph networks for deep learning on
  dynamic graphs}.
\newblock \bibinfo{journal}{\emph{arXiv preprint arXiv:2006.10637}}
  (\bibinfo{year}{2020}).
\newblock


\bibitem[Rubinsteyn and Feldman(2016)]%
        {fancyimpute}
\bibfield{author}{\bibinfo{person}{Alex Rubinsteyn} {and}
  \bibinfo{person}{Sergey Feldman}.} \bibinfo{year}{2016}\natexlab{}.
\newblock \bibinfo{title}{fancyimpute: An Imputation Library for Python}.
\newblock \bibinfo{howpublished}{https://github.com/iskandr/fancyimpute}.
\newblock


\bibitem[Shuman et~al\mbox{.}(2013)]%
        {shuman2013emerging}
\bibfield{author}{\bibinfo{person}{David~I Shuman}, \bibinfo{person}{Sunil~K
  Narang}, \bibinfo{person}{Pascal Frossard}, \bibinfo{person}{Antonio Ortega},
  {and} \bibinfo{person}{Pierre Vandergheynst}.}
  \bibinfo{year}{2013}\natexlab{}.
\newblock \showarticletitle{The emerging field of signal processing on graphs:
  Extending high-dimensional data analysis to networks and other irregular
  domains}.
\newblock \bibinfo{journal}{\emph{IEEE signal processing magazine}}
  \bibinfo{volume}{30}, \bibinfo{number}{3} (\bibinfo{year}{2013}),
  \bibinfo{pages}{83--98}.
\newblock


\bibitem[Tay et~al\mbox{.}(2022)]%
        {tay2022efficient}
\bibfield{author}{\bibinfo{person}{Yi Tay}, \bibinfo{person}{Mostafa Dehghani},
  \bibinfo{person}{Dara Bahri}, {and} \bibinfo{person}{Donald Metzler}.}
  \bibinfo{year}{2022}\natexlab{}.
\newblock \showarticletitle{Efficient transformers: A survey}.
\newblock \bibinfo{journal}{\emph{Comput. Surveys}} \bibinfo{volume}{55},
  \bibinfo{number}{6} (\bibinfo{year}{2022}), \bibinfo{pages}{1--28}.
\newblock


\bibitem[Tishby et~al\mbox{.}(2000)]%
        {tishby2000information}
\bibfield{author}{\bibinfo{person}{Naftali Tishby}, \bibinfo{person}{Fernando~C
  Pereira}, {and} \bibinfo{person}{William Bialek}.}
  \bibinfo{year}{2000}\natexlab{}.
\newblock \showarticletitle{The information bottleneck method}.
\newblock \bibinfo{journal}{\emph{arXiv preprint physics/0004057}}
  (\bibinfo{year}{2000}).
\newblock


\bibitem[Tong et~al\mbox{.}(2006)]%
        {tong2006fast}
\bibfield{author}{\bibinfo{person}{Hanghang Tong}, \bibinfo{person}{Christos
  Faloutsos}, {and} \bibinfo{person}{Jia-Yu Pan}.}
  \bibinfo{year}{2006}\natexlab{}.
\newblock \showarticletitle{Fast random walk with restart and its
  applications}. In \bibinfo{booktitle}{\emph{Sixth international conference on
  data mining (ICDM'06)}}. IEEE, \bibinfo{pages}{613--622}.
\newblock


\bibitem[Vaswani et~al\mbox{.}(2017)]%
        {vaswani2017attention}
\bibfield{author}{\bibinfo{person}{Ashish Vaswani}, \bibinfo{person}{Noam
  Shazeer}, \bibinfo{person}{Niki Parmar}, \bibinfo{person}{Jakob Uszkoreit},
  \bibinfo{person}{Llion Jones}, \bibinfo{person}{Aidan~N Gomez},
  \bibinfo{person}{{\L}ukasz Kaiser}, {and} \bibinfo{person}{Illia
  Polosukhin}.} \bibinfo{year}{2017}\natexlab{}.
\newblock \showarticletitle{Attention is all you need}.
\newblock \bibinfo{journal}{\emph{Advances in neural information processing
  systems}}  \bibinfo{volume}{30} (\bibinfo{year}{2017}).
\newblock


\bibitem[Wang et~al\mbox{.}(2020b)]%
        {wang2020personalized}
\bibfield{author}{\bibinfo{person}{Hanzhi Wang}, \bibinfo{person}{Zhewei Wei},
  \bibinfo{person}{Junhao Gan}, \bibinfo{person}{Sibo Wang}, {and}
  \bibinfo{person}{Zengfeng Huang}.} \bibinfo{year}{2020}\natexlab{b}.
\newblock \showarticletitle{Personalized pagerank to a target node, revisited}.
  In \bibinfo{booktitle}{\emph{Proceedings of the 26th ACM SIGKDD International
  Conference on Knowledge Discovery \& Data Mining}}.
  \bibinfo{pages}{657--667}.
\newblock


\bibitem[Wang et~al\mbox{.}(2022)]%
        {wang2022equivariant}
\bibfield{author}{\bibinfo{person}{Haorui Wang}, \bibinfo{person}{Haoteng Yin},
  \bibinfo{person}{Muhan Zhang}, {and} \bibinfo{person}{Pan Li}.}
  \bibinfo{year}{2022}\natexlab{}.
\newblock \showarticletitle{Equivariant and stable positional encoding for more
  powerful graph neural networks}.
\newblock \bibinfo{journal}{\emph{arXiv preprint arXiv:2203.00199}}
  (\bibinfo{year}{2022}).
\newblock


\bibitem[Wang et~al\mbox{.}(2020a)]%
        {wang2020linformer}
\bibfield{author}{\bibinfo{person}{Sinong Wang}, \bibinfo{person}{Belinda~Z
  Li}, \bibinfo{person}{Madian Khabsa}, \bibinfo{person}{Han Fang}, {and}
  \bibinfo{person}{Hao Ma}.} \bibinfo{year}{2020}\natexlab{a}.
\newblock \showarticletitle{Linformer: Self-attention with linear complexity}.
\newblock \bibinfo{journal}{\emph{arXiv preprint arXiv:2006.04768}}
  (\bibinfo{year}{2020}).
\newblock


\bibitem[White et~al\mbox{.}(2011)]%
        {white2011multiple}
\bibfield{author}{\bibinfo{person}{Ian~R White}, \bibinfo{person}{Patrick
  Royston}, {and} \bibinfo{person}{Angela~M Wood}.}
  \bibinfo{year}{2011}\natexlab{}.
\newblock \showarticletitle{Multiple imputation using chained equations: issues
  and guidance for practice}.
\newblock \bibinfo{journal}{\emph{Statistics in medicine}}
  \bibinfo{volume}{30}, \bibinfo{number}{4} (\bibinfo{year}{2011}),
  \bibinfo{pages}{377--399}.
\newblock


\bibitem[Wu et~al\mbox{.}(2022)]%
        {wu2022timesnet}
\bibfield{author}{\bibinfo{person}{Haixu Wu}, \bibinfo{person}{Tengge Hu},
  \bibinfo{person}{Yong Liu}, \bibinfo{person}{Hang Zhou},
  \bibinfo{person}{Jianmin Wang}, {and} \bibinfo{person}{Mingsheng Long}.}
  \bibinfo{year}{2022}\natexlab{}.
\newblock \showarticletitle{TimesNet: Temporal 2D-Variation Modeling for
  General Time Series Analysis}.
\newblock \bibinfo{journal}{\emph{arXiv preprint arXiv:2210.02186}}
  (\bibinfo{year}{2022}).
\newblock


\bibitem[Wu et~al\mbox{.}(2020)]%
        {wu2020connecting}
\bibfield{author}{\bibinfo{person}{Zonghan Wu}, \bibinfo{person}{Shirui Pan},
  \bibinfo{person}{Guodong Long}, \bibinfo{person}{Jing Jiang},
  \bibinfo{person}{Xiaojun Chang}, {and} \bibinfo{person}{Chengqi Zhang}.}
  \bibinfo{year}{2020}\natexlab{}.
\newblock \showarticletitle{Connecting the dots: Multivariate time series
  forecasting with graph neural networks}. In
  \bibinfo{booktitle}{\emph{Proceedings of the 26th ACM SIGKDD international
  conference on knowledge discovery \& data mining}}.
  \bibinfo{pages}{753--763}.
\newblock


\bibitem[Xu et~al\mbox{.}(2019)]%
        {xu2019self}
\bibfield{author}{\bibinfo{person}{Da Xu}, \bibinfo{person}{Chuanwei Ruan},
  \bibinfo{person}{Evren Korpeoglu}, \bibinfo{person}{Sushant Kumar}, {and}
  \bibinfo{person}{Kannan Achan}.} \bibinfo{year}{2019}\natexlab{}.
\newblock \showarticletitle{Self-attention with functional time representation
  learning}.
\newblock \bibinfo{journal}{\emph{Advances in neural information processing
  systems}}  \bibinfo{volume}{32} (\bibinfo{year}{2019}).
\newblock


\bibitem[Xu et~al\mbox{.}(2020)]%
        {xu2020inductive}
\bibfield{author}{\bibinfo{person}{Da Xu}, \bibinfo{person}{Chuanwei Ruan},
  \bibinfo{person}{Evren Korpeoglu}, \bibinfo{person}{Sushant Kumar}, {and}
  \bibinfo{person}{Kannan Achan}.} \bibinfo{year}{2020}\natexlab{}.
\newblock \showarticletitle{Inductive representation learning on temporal
  graphs}.
\newblock \bibinfo{journal}{\emph{arXiv preprint arXiv:2002.07962}}
  (\bibinfo{year}{2020}).
\newblock


\bibitem[Xu et~al\mbox{.}(2018)]%
        {xu2018powerful}
\bibfield{author}{\bibinfo{person}{Keyulu Xu}, \bibinfo{person}{Weihua Hu},
  \bibinfo{person}{Jure Leskovec}, {and} \bibinfo{person}{Stefanie Jegelka}.}
  \bibinfo{year}{2018}\natexlab{}.
\newblock \showarticletitle{How powerful are graph neural networks?}
\newblock \bibinfo{journal}{\emph{arXiv preprint arXiv:1810.00826}}
  (\bibinfo{year}{2018}).
\newblock


\bibitem[Xu et~al\mbox{.}(2021)]%
        {xu2021hist}
\bibfield{author}{\bibinfo{person}{Wentao Xu}, \bibinfo{person}{Weiqing Liu},
  \bibinfo{person}{Lewen Wang}, \bibinfo{person}{Yingce Xia},
  \bibinfo{person}{Jiang Bian}, \bibinfo{person}{Jian Yin}, {and}
  \bibinfo{person}{Tie-Yan Liu}.} \bibinfo{year}{2021}\natexlab{}.
\newblock \showarticletitle{HIST: A Graph-based Framework for Stock Trend
  Forecasting via Mining Concept-Oriented Shared Information}.
\newblock \bibinfo{journal}{\emph{arXiv preprint arXiv:2110.13716}}
  (\bibinfo{year}{2021}).
\newblock


\bibitem[Yan et~al\mbox{.}(2021a)]%
        {yan2021dynamic}
\bibfield{author}{\bibinfo{person}{Yuchen Yan}, \bibinfo{person}{Lihui Liu},
  \bibinfo{person}{Yikun Ban}, \bibinfo{person}{Baoyu Jing}, {and}
  \bibinfo{person}{Hanghang Tong}.} \bibinfo{year}{2021}\natexlab{a}.
\newblock \showarticletitle{Dynamic knowledge graph alignment}. In
  \bibinfo{booktitle}{\emph{Proceedings of the AAAI Conference on Artificial
  Intelligence}}, Vol.~\bibinfo{volume}{35}. \bibinfo{pages}{4564--4572}.
\newblock


\bibitem[Yan et~al\mbox{.}(2021b)]%
        {yan2021bright}
\bibfield{author}{\bibinfo{person}{Yuchen Yan}, \bibinfo{person}{Si Zhang},
  {and} \bibinfo{person}{Hanghang Tong}.} \bibinfo{year}{2021}\natexlab{b}.
\newblock \showarticletitle{Bright: A bridging algorithm for network
  alignment}. In \bibinfo{booktitle}{\emph{Proceedings of the Web Conference
  2021}}. \bibinfo{pages}{3907--3917}.
\newblock


\bibitem[Yan et~al\mbox{.}(2022)]%
        {yan2022dissecting}
\bibfield{author}{\bibinfo{person}{Yuchen Yan}, \bibinfo{person}{Qinghai Zhou},
  \bibinfo{person}{Jinning Li}, \bibinfo{person}{Tarek Abdelzaher}, {and}
  \bibinfo{person}{Hanghang Tong}.} \bibinfo{year}{2022}\natexlab{}.
\newblock \showarticletitle{Dissecting cross-layer dependency inference on
  multi-layered inter-dependent networks}. In
  \bibinfo{booktitle}{\emph{Proceedings of the 31st ACM International
  Conference on Information \& Knowledge Management}}.
  \bibinfo{pages}{2341--2351}.
\newblock


\bibitem[Yeung(1991)]%
        {yeung1991new}
\bibfield{author}{\bibinfo{person}{Raymond~W Yeung}.}
  \bibinfo{year}{1991}\natexlab{}.
\newblock \showarticletitle{A new outlook on Shannon's information measures}.
\newblock \bibinfo{journal}{\emph{IEEE transactions on information theory}}
  \bibinfo{volume}{37}, \bibinfo{number}{3} (\bibinfo{year}{1991}),
  \bibinfo{pages}{466--474}.
\newblock


\bibitem[Yi et~al\mbox{.}(2016)]%
        {yi2016st}
\bibfield{author}{\bibinfo{person}{Xiuwen Yi}, \bibinfo{person}{Yu Zheng},
  \bibinfo{person}{Junbo Zhang}, {and} \bibinfo{person}{Tianrui Li}.}
  \bibinfo{year}{2016}\natexlab{}.
\newblock \showarticletitle{ST-MVL: filling missing values in geo-sensory time
  series data}. In \bibinfo{booktitle}{\emph{Proceedings of the 25th
  International Joint Conference on Artificial Intelligence}}.
\newblock


\bibitem[Yoon et~al\mbox{.}(2018)]%
        {yoon2018gain}
\bibfield{author}{\bibinfo{person}{Jinsung Yoon}, \bibinfo{person}{James
  Jordon}, {and} \bibinfo{person}{Mihaela Schaar}.}
  \bibinfo{year}{2018}\natexlab{}.
\newblock \showarticletitle{Gain: Missing data imputation using generative
  adversarial nets}. In \bibinfo{booktitle}{\emph{International conference on
  machine learning}}. PMLR, \bibinfo{pages}{5689--5698}.
\newblock


\bibitem[You et~al\mbox{.}(2019)]%
        {you2019position}
\bibfield{author}{\bibinfo{person}{Jiaxuan You}, \bibinfo{person}{Rex Ying},
  {and} \bibinfo{person}{Jure Leskovec}.} \bibinfo{year}{2019}\natexlab{}.
\newblock \showarticletitle{Position-aware graph neural networks}. In
  \bibinfo{booktitle}{\emph{International conference on machine learning}}.
  PMLR, \bibinfo{pages}{7134--7143}.
\newblock


\bibitem[Yu et~al\mbox{.}(2017)]%
        {yu2017spatio}
\bibfield{author}{\bibinfo{person}{Bing Yu}, \bibinfo{person}{Haoteng Yin},
  {and} \bibinfo{person}{Zhanxing Zhu}.} \bibinfo{year}{2017}\natexlab{}.
\newblock \showarticletitle{Spatio-temporal graph convolutional networks: A
  deep learning framework for traffic forecasting}.
\newblock \bibinfo{journal}{\emph{arXiv preprint arXiv:1709.04875}}
  (\bibinfo{year}{2017}).
\newblock


\bibitem[Zhang et~al\mbox{.}(2020)]%
        {zhang2020spatio}
\bibfield{author}{\bibinfo{person}{Qi Zhang}, \bibinfo{person}{Jianlong Chang},
  \bibinfo{person}{Gaofeng Meng}, \bibinfo{person}{Shiming Xiang}, {and}
  \bibinfo{person}{Chunhong Pan}.} \bibinfo{year}{2020}\natexlab{}.
\newblock \showarticletitle{Spatio-temporal graph structure learning for
  traffic forecasting}. In \bibinfo{booktitle}{\emph{Proceedings of the AAAI
  Conference on Artificial Intelligence}}, Vol.~\bibinfo{volume}{34}.
  \bibinfo{pages}{1177--1185}.
\newblock


\bibitem[Zheng et~al\mbox{.}(2022)]%
        {zheng2022graph}
\bibfield{author}{\bibinfo{person}{Xin Zheng}, \bibinfo{person}{Yixin Liu},
  \bibinfo{person}{Shirui Pan}, \bibinfo{person}{Miao Zhang},
  \bibinfo{person}{Di Jin}, {and} \bibinfo{person}{Philip~S Yu}.}
  \bibinfo{year}{2022}\natexlab{}.
\newblock \showarticletitle{Graph neural networks for graphs with heterophily:
  A survey}.
\newblock \bibinfo{journal}{\emph{arXiv preprint arXiv:2202.07082}}
  (\bibinfo{year}{2022}).
\newblock


\bibitem[Zheng et~al\mbox{.}(2015)]%
        {zheng2015forecasting}
\bibfield{author}{\bibinfo{person}{Yu Zheng}, \bibinfo{person}{Xiuwen Yi},
  \bibinfo{person}{Ming Li}, \bibinfo{person}{Ruiyuan Li},
  \bibinfo{person}{Zhangqing Shan}, \bibinfo{person}{Eric Chang}, {and}
  \bibinfo{person}{Tianrui Li}.} \bibinfo{year}{2015}\natexlab{}.
\newblock \showarticletitle{Forecasting fine-grained air quality based on big
  data}. In \bibinfo{booktitle}{\emph{Proceedings of the 21th ACM SIGKDD
  international conference on knowledge discovery and data mining}}.
  \bibinfo{pages}{2267--2276}.
\newblock


\bibitem[Zhu et~al\mbox{.}(2021)]%
        {zhu2021graph}
\bibfield{author}{\bibinfo{person}{Jiong Zhu}, \bibinfo{person}{Ryan~A Rossi},
  \bibinfo{person}{Anup Rao}, \bibinfo{person}{Tung Mai},
  \bibinfo{person}{Nedim Lipka}, \bibinfo{person}{Nesreen~K Ahmed}, {and}
  \bibinfo{person}{Danai Koutra}.} \bibinfo{year}{2021}\natexlab{}.
\newblock \showarticletitle{Graph neural networks with heterophily}. In
  \bibinfo{booktitle}{\emph{Proceedings of the AAAI Conference on Artificial
  Intelligence}}, Vol.~\bibinfo{volume}{35}. \bibinfo{pages}{11168--11176}.
\newblock


\end{thebibliography}

\clearpage
\appendix
\section{Appendix}
In the appendix, we present the additional details of \vae\ including
\begin{itemize}
    \item Proofs of Proposition~\ref{rwr} and Theorem~\ref{temporal-theo} in Section~\ref{app:rwr} and Section~\ref{app:theo} respectively.
    \item Additional details of components in \vae\ is introduced in Section~\ref{app:details}.
    \item Reproducibility and parameter settings of baselines and the proposed \vae\ are listed in Section~\ref{app:parameter}.
    \item Additional experiments such as visualization for prediction, ablation study over self-attention in \vae\ and ablation study over RWR restart probability $c$ in \vae\ are given in Section~\ref{app:exp}.
    \item Section~\ref{app:limitation-future} discusses the limitations of our implementations of \vae\ and propose some potential future works based on NTS imputation.
\end{itemize}

\vspace{-2mm}
\subsection{Proof Proposition \ref{rwr}}\label{app:rwr}

% \hh{relative lower priority, but we should also carefully polish the writing in Appendix: the structure (e.g., put a roadmap paragraph/bullet points before Appendix A), the rigor/correctness and the language}

We prove Proposition \ref{rwr} by analyzing the properties of RWR.
\vspace{-2mm}
\begin{proposition*}
Random walk with restarts (RWR) captures information from close neighbors (local) and long-distance neighbors (global) in graph learning.
\end{proposition*}
\begin{proof}
\vspace{-2mm}
Based on Eq.~\eqref{rwr-equa}, the closed form solution for RWR can be derived as: $\mathbf{r}_i = c \cdot (\mathbf{I} - (1-c) \cdot \mathbf{\hat{A}}) ^ {-1} \mathbf{e}_i$, where $\mathbf{I}$ is the identity matrix. We could also solve this equation by power iterations: $(\mathbf{I} - (1-c) \cdot \mathbf{\hat{A}}) ^ {-1} \approx \sum_{k=0}^{\infty} ((1-c)\cdot \mathbf{\hat{A}})^k$.
% \hh{i think the relationship is exact since you have the summation to infinite}.
First of all, as the power term $k$ goes to infinity, the position embedding $\mathbf{R}$ can indeed capture \textit{global} information of graph. Second, the restart probability $c$ ensures nodes close to anchor nodes have larger values than those farther away, which encodes the \textit{local} information of graph.

{\em Remark.} Proposition \ref{rwr} holds with the assumption that the graph is connected. When graph is not connected, with proper choice of a set of anchor nodes that cover all the connected components, RWR is able to \textit{global} information within each components rather than the entire graph. An alternative to generate node position embeddings is using RWR from each node similar to~\cite{dwivedi2021graph} to get the landing probability of a node to itself for multiple steps. However, this usually increases the complexity when having a large graph and still faces similar issues as ours when graph is less connected.
\end{proof}

\subsection{Proof of Theorem \ref{temporal-theo}}\label{app:theo}
To prove Theorem~\ref{temporal-theo}, we first introduce the following proposition.
\vspace{-3mm}
\begin{proposition} \label{mutual}
For any random variables $A$, $B$, and $C$, the following inequality of the mutual information $I(\cdot;\cdot)$ holds~\cite{peng2020graph}:
\begin{align}
    I(A, B; C) \geq I(A;C)
\end{align}
% \hh{is this from the existing work? if so, cite it. also, let's demote it as a proposition}
\end{proposition}
\begin{proof}
\vspace{-3mm}
    Based on the chain rule of mutual information, we have:
    \begin{align*}
        I(A, B; C) & = H(A,B) - H(A, B | C) \\
        & = H(A) + H(B|A) - H(A|C) - H(B|A, C)\\
        & = I(A; C) + I(B; C|A)
    \end{align*}
    where $H(A)$ is the marginal entropy, $H(A|C)$ is the conditional entropy and $H(A,B)$ is the joint entropy. Since the mutual information $I(B; C|A) \geq 0$, we can conclude that $I(A, B; C) \geq I(A;C)$.
\end{proof}
Now we can prove Theorem~\ref{temporal-theo} as:
\begin{theorem*}
\vspace{-1mm}
Given a temporal graph $\mathcal{G}$, TGN with RWR-based node position embeddings $g_{\theta}$ has more expressive power than regular TGN $f_{\theta}$ in node representation learning: $\mathbb{D}(g(u),g(v)) \geq \mathbb{D}(f(u),f(v))$ where $\mathbb{D}(\cdot, \cdot)$ measures the expressiveness by counting the distinguishable node pairs $(u,v)$ in $\mathcal{G}$ based on node representations.
% \hh{do we need an equation similar to eq13 here?}
\end{theorem*}
\begin{proof}
\vspace{-2mm}
    It is natural to see that $g_{\theta}$ has at least same expressive power as $f_{\theta}$ since we add additional information with the positional embeddings for each node. By setting all the parameters of $g_{\theta}$ that handle such positional embeddings to zero, we will have a regular TGN model same as $f_{\theta}$.
    
    To prove that why the additional information brought by node position embeddings is useful for node representation learning, we provide following analysis with the help of Proposition~\ref{mutual}. Regular TGN only encodes topologically local information within $q$-hop neighbors and $q$ usually is a small number because of the over-smoothing problem~\cite{li2018deeper}, we denote the random variable for $f_{\theta}$'s node representations as $X_{local}$. Based on Proposition~\ref{rwr}, we know that the random variable for $g_{\theta}$'s node representations $X_{local+global}$ follows the joint distribution of both local and global topological information. The objective of a node representation learning task over graph $\mathcal{G}$ can be denoted as $\max I(X;Y)$ where $Y$ is the random variable follows label distributions~\cite{you2019position}. This derivation can be obtained from Information Bottleneck we discussed in Section \ref{sec:multi-task} without the constraints term. Therefore, based on Proposition~\ref{mutual}, we have $I(X_{local+global};Y) \geq I(X_{local};Y)$ which denotes that $g_{\theta}$ has more expressive power than $f_{\theta}$.
\end{proof}

\subsection{Additional Details over \vae\ }\label{app:details}
\subsubsection{Details of message-passing neural network}
The message-passing neural network (MPNN) used in \vae\ is defined as:
\begin{align}
\text{MPNN}(F_u, F_m, \mathbf{d}_{t,i}, \mathbf{A}) = F_u(\mathbf{d}_{t,i}, & \sum_{\substack{j \in \mathcal{N}(i)}} F_m(\mathbf{h}_{t,i}, \mathbf{d}_{t,j}, e_{i,j}))
\end{align}
\noindent where $F_u$ and $F_m$ are update and message functions with learnable parameters, $\mathbf{d}_{t,i}$ is the node representation for node $i$ at time step $t$, $e_{i,j}$ is the edge weight between node $i$ and $j$, and $\mathcal{N}(i)$ represents node $i$'s neighbors. The two-layer MPNNs with skip connection used in \vae\ can be defined as:
% We apply a two-layer MPNNs with skip connection to obtain the hidden node embeddings $\mathbf{H}_t^{out}$ based on the predicted graph $\mathbf{A}^{\text{out}}_t$:
% % \hh{can we change $F_{u,1}$ to $F_{u}^1$? same change for $F_{u,2}$, $F_{m,1}$, and $F_{m,2}$}
\begin{align}
\begin{split}
    & \mathbf{H}_{1,t} = \text{MPNN}(F_u^1, F_m^1, \mathbf{U}_t, \mathbf{A}_t^{\text{out}}) \\
    & \mathbf{H}_{2,t} = \text{MPNN}(F_u^2, F_m^2, \mathbf{H}_{t,1}, \mathbf{A}_t^{\text{out}}) \\
    & \mathbf{H}_{t}^{\text{graph}} = \mathbf{H}_{1,t} \oplus \mathbf{H}_{2,t}
\end{split}
\end{align}
where $\oplus$ is the element-wise addition.

\subsubsection{Details of self-attention}
The self-attention module is defined as:
\begin{align}
\begin{split}
    & \text{Attn}(\mathbf{h}_{t,i}) = \sum_{j\in \mathcal{N}(i)} \text{softmax}\big ( \frac{\mathbf{Q}_i\mathbf{K}_j^T}{\sqrt{d}} \big ) \mathbf{V}_j \\
    \text{and} \quad & \mathbf{Q}_i = \mathbf{h}_{t,i}\mathbf{W}_Q, \mathbf{K}_j = \mathbf{h}_{t,j}\mathbf{W}_K, \mathbf{V}_j = \mathbf{h}_{t,j}\mathbf{W}_V
\end{split}
\end{align}
where $\mathcal{N}(i)$ is the neighbor of node $i$, $d$ is the hidden dimension and $\mathbf{W}_Q,\mathbf{W}_K,\mathbf{W}_V$ are learnable parameters.

\subsection{Reproducibility}\label{app:parameter}
We introduce the detailed parameter settings of our models as well as baselines in this subsection. For \vae, the restart probability $c$ of RWR for position embeddings is set to the commonly used 0.15, and we picked the number of anchor nodes as $L=\log_2(N)$. We set the hidden dimension to be $64$, $\beta$ to be $0.2$ and $\gamma$ to be $0.01$.

All the experiments are based on codes from a open source library\footnote{https://github.com/iskandr/fancyimpute} \cite{fancyimpute} and those provided by corresponding authors. We modify their implementations for the NTS imputation problem and the details of parameter settings are listed as follows: for the parameters for each baseline model in the time series feature imputation, we refer to previous works for their settings \cite{cao2018brits,cini2021filling}. For BRITS\footnote{https://github.com/caow13/BRITS}, we use the same hidden dimension as \cite{cao2018brits} for AQ36 dataset, and for the traffic datasets, the hidden size is set to 256 which is aligned with the setting used in PeMS data in \cite{cini2021filling}. For rGAIN, we follow the exact same setting as in \cite{cini2021filling,marisca2022learning} in which we use 64 as the hidden size for AQ36 dataset and 256 for traffic datasets. For SAITS and TimesNet, we follow the exact parameter settings in their papers~\cite{du2023saits,wu2022timesnet}. For GRIN\footnote{https://github.com/Graph-Machine-Learning-Group/grin}, we use the same hidden size for AQ36 as \cite{cini2021filling}, while using 80 for the traffic datasets since they are much larger than AQ36. As for the hidden dimension of $\text{NET}^3$ \footnote{https://github.com/baoyujing/NET3}, we use $128$ for AQ36 dataset and $256$ for traffic datasets. For COVID-19 dataset, we use the same parameters settings of traffic datasets for all the models .

For VGAE and VGRNN in the link prediction task in NTS imputation, we use hidden size 256/128 for the AQ36 and 320/150 for the traffic datasets respectively.

We train all the models using PyTorch \cite{paszke2019pytorch} with Adam optimizer \cite{kingma2014adam}, learning rates are set to be $0.001 / 0.01$ for time series feature imputation and link prediction baselines respectively with cosine annealing scheduler \cite{loshchilov2016sgdr} to adjust the values dynamically. The batch sizes are all set to 32 and we use validation dataset for early stopping.

% \vspace{-1mm}
\subsection{Additional Experiments}\label{app:exp}
% \yc{Rephrase}
\subsubsection{Visualization of Prediction}\label{sec:pred-visual}
\begin{figure}[H]
\includegraphics[width=0.48\textwidth,trim = 2 2 2 2,clip]{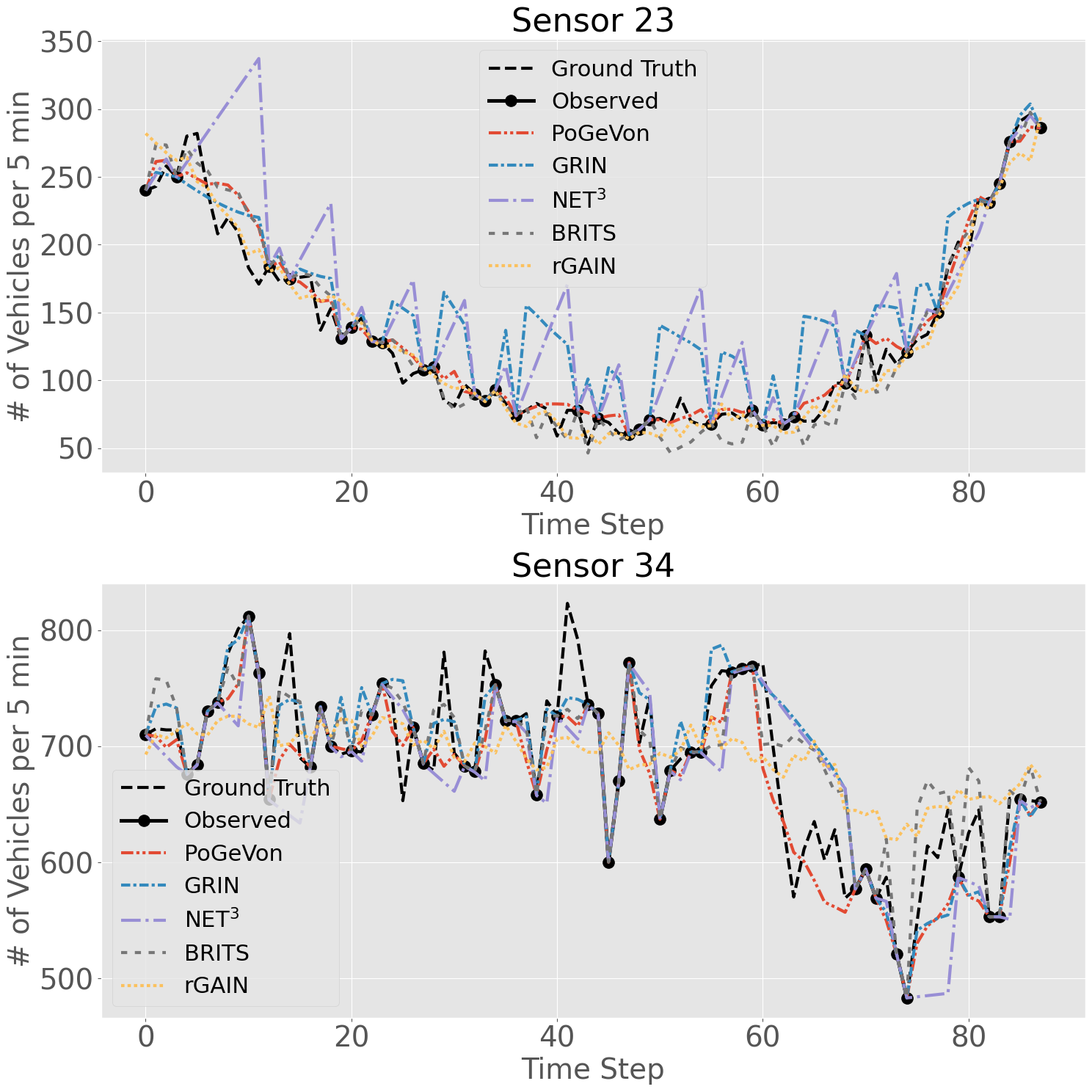}
\vspace{-6mm}
\caption{Different models' predictions of traffic flow in sensor 23 and 34 from PeMS-LA dataset. Best viewed in color.}
\label{fig:pems-pred}
\end{figure}

The prediction results of different selected baselines over PeMS-LA with $50\%$ missing rates over test data can be found in Figure \ref{fig:pems-pred}. It is clear that \vae\ can achieve better predictions results compared with other baselines. In particular, GRIN and $\text{NET}^3$ sometimes suffer a lot from fluctuations due to the missing edges in the NTS data, which result in poor performance compared to our proposed \vae. Besides, we can see that \vae\ can have finer predictions compared with all the baselines when there exist abrupt change of data values which also has high missing rates (e.g., data from time step 60 to 65 in the figure of prediction of sensor 34).

\subsubsection{Ablations on self-attention in \vae}\label{sec:attn-ablation}
We conduct ablation study over the self-attention module in \vae\ by changing it to attention mechanism used in Linformer and Reformer respectively. The results of time series imputation over COVID-19 and AQ36 datasets are shown in Tables \ref{table:attn-covid} and \ref{table:attn-aq36} respectively. It is clear that replacing vanilla self-attention with Linformer self-attention as well as LSH self-attention from Reformer can still achieve better results on COVID-19 with the strongest baseline MICE. They can also have better or comparable results over AQ36 compared with the strongest baseline BRITS as well. 
\begin{table}[ht]
\setlength{\abovecaptionskip}{3pt}
\caption{Ablation study of self-attention in \vae\ over COVID-19 datasets on time series feature imputation. Smaller is better.}
\begin{tabularx}{\columnwidth}{C|Y|Y|Y} 
\toprule
Model & MAE & MSE & MRE  \\ 
\specialrule{1pt}{1pt}{3pt}
MICE & \small{0.077} \footnotesize{$\pm 0.005$} & \small{0.013} \footnotesize{$\pm 0.002$} & \small{0.007} \footnotesize{$\pm 0.000$}\\
\midrule
\vae\  & \small{\textbf{0.007}} \footnotesize{$\pm \textbf{0.001}$} & \small{\textbf{0.000}} \footnotesize{$\pm \textbf{0.000}$} & \small{\textbf{0.001}} \footnotesize{$\pm \textbf{0.000}$} \\
\midrule
\vae\ with Linformer self-attention & \small{0.012} \footnotesize{$\pm 0.004$} & \small{0.001} \footnotesize{$\pm 0.000$} & \small{0.001} \footnotesize{$\pm 0.000$} \\ 
\midrule
% Remove 1-stage decoder & \small{60.87} \footnotesize{$\pm 0.50$} & \small{6517.95} \footnotesize{$\pm 88.11$} & \small{0.82} \footnotesize{$\pm 0.01$} \\ 
% \midrule
\vae\ with LSH self-attention & \small{0.009} \footnotesize{$\pm 0.004$} & \small{0.000} \footnotesize{$\pm 0.000$} & \small{0.001} \footnotesize{$\pm 0.000$} \\
\bottomrule
\end{tabularx}
\label{table:attn-covid}
% \vspace{-4mm}
\end{table}

\begin{table}[ht]
\setlength{\abovecaptionskip}{3pt}
\caption{Ablation study of self-attention in \vae\ over AQ36 datasets on time series feature imputation. Smaller is better.}
\begin{tabularx}{\columnwidth}{C|Y|Y|Y} 
\toprule
Model & MAE & MSE & MRE  \\ 
\specialrule{1pt}{1pt}{3pt}
BRITS & \small{23.39} \footnotesize{$\pm 0.80$} & \small{1276.23} \footnotesize{$\pm 102.92$} & \small{0.31} \footnotesize{$\pm 0.011$}\\
\midrule
\vae\  & \small{\textbf{19.49}} \footnotesize{$\pm \textbf{1.10}$} & \small{\textbf{1213.47}} \footnotesize{$\pm \textbf{125.53}$} & \small{\textbf{0.26}} \footnotesize{$\pm \textbf{0.02}$} \\
\midrule
\vae\ with Linformer self-attention & \small{21.49} \footnotesize{$\pm 1.40$} & \small{1333.61} \footnotesize{$\pm 168.71$} & \small{0.28} \footnotesize{$\pm 0.02$} \\ 
\midrule
% Remove 1-stage decoder & \small{60.87} \footnotesize{$\pm 0.50$} & \small{6517.95} \footnotesize{$\pm 88.11$} & \small{0.82} \footnotesize{$\pm 0.01$} \\ 
% \midrule
\vae\ with LSH self-attention & \small{23.69} \footnotesize{$\pm 1.35$} & \small{1422.27} \footnotesize{$\pm 123.62$} & \small{0.32} \footnotesize{$\pm 0.02$} \\ \bottomrule
\end{tabularx}
\label{table:attn-aq36}
% \vspace{-4mm}
\end{table}

\subsubsection{Ablations on RWR Restart Probability in \vae}\label{sec:c-ablation}
We have also conducted ablation studies on the RWR restart probability for our position node embeddings in \vae. In literature, the restart probability is often set to be a small number (e.g., 0.15). A high restart probability makes RWR-based node embeddings near anchor nodes be more similar to each other which increases the local information while diminishes the global information. A low restart probability sometimes results in a sparse matrix because of the deadend nodes of graphs and RWR algorithm will degenerate to a vanilla PageRank/Random Walk equations. The experiment results on this aspect by setting the restart probability to different levels: $0.1$, $0.2$, $0.4$ and $0.8$ are shown in the Table~\ref{table:c-covid}. Based on the experiment results, we do observe performance drop of \vae when choosing less effective restart probability for RWR-based node position embeddings. However, \vae demonstrates certain stability and can still outperform other baseline models.
\begin{table}[ht]
\setlength{\abovecaptionskip}{3pt}
\caption{Ablation study of RWR restart probability $c$ in \vae\ over COVID-19 datasets on time series feature imputation. Smaller is better.}
\begin{tabularx}{\columnwidth}{Z|Y|Y|Y} 
\toprule
Model & MAE & MSE & MRE  \\ 
\specialrule{1pt}{1pt}{3pt}
\vae\ w. $c=0.1$ & \small{0.00734} \footnotesize{$\pm 0.00158$} & \small{0.00017} \footnotesize{$\pm 0.00012$} & \small{0.00068} \footnotesize{$\pm 0.00015$}\\
\midrule
\vae\ w. $c=0.2$ & \small{0.00785} \footnotesize{$\pm 0.00222$} & \small{0.00021} \footnotesize{$\pm 0.00016$} & \small{0.00073} \footnotesize{$\pm 0.00021$}\\
\midrule
\vae\ w. $c=0.4$ & \small{0.00739} \footnotesize{$\pm 0.00154$} & \small{0.00015} \footnotesize{$\pm 0.00012$} & \small{0.00064} \footnotesize{$\pm 0.00014$} \\ 
\midrule
\vae\ w. $c=0.8$ & \small{0.00737} \footnotesize{$\pm 0.00167$} & \small{0.00014} \footnotesize{$\pm 0.00007$} & \small{0.00066} \footnotesize{$\pm 0.00016$} \\
\midrule
\vae\  & \small{\textbf{0.00690}} \footnotesize{$\pm \textbf{0.00085}$} & \small{\textbf{0.00013}} \footnotesize{$\pm \textbf{0.00007}$} & \small{\textbf{0.00064}} \footnotesize{$\pm \textbf{0.00008}$} \\
\end{tabularx}
\label{table:c-covid}
% \vspace{-4mm}
\end{table}

\subsection{Limitations and Future Works}\label{app:limitation-future}
%\hh{let's polish this part. the main purpose is to use it as a pro-active defense strategy in case some reviewers attack us on these points. if we do not do this carefully, it will be come a self-sabtage: the reviewers can simply use our own words to attack us. let's identify 2-3 limitations (e.g., quadratic time complexity due to the attention; the imputationabilty/impossiblity (if the missing rate keeps increases, under which condition, it becomes impoassible to impute); the negative transfer (under which condition mlt might hurt the imputation performance, maybe when negative correlation exists. in this case, we might consider gnn designed for graph heterophilly), and for each of them, let's state what the limitation is, and how we *might* be able to address it (at very high level, with 2-3 sentences).}
%\derek{For the imputationability/impossibility part, I feel it won't be a big problem as shown in sensitivity analysis section. Besides, there exist some works that can handle extremely sparse time series imputation problem. Discussing this aspect might instead bring negative effects to our paper. Thus, I only keep the other two points which are more insightful.}

One limitation of the proposed \vae\ model lies in its quadratic complexity $\mathcal{O}(N^2)$ due to %: (1) the $\mathcal{O}(N^2)$ complexity due to 
the self-attention module. As we have discussed in Section~\ref{sec:complexity}, we can reduce this complexity to either $\mathcal{O}(N)$ by Linformer~\cite{wang2020linformer} or $\mathcal{O}(N\log N)$ by Reformer~\cite{kitaev2020reformer}. Another limitation lies in the potential %The potential solutions are proposed in  and some experiment results are shown in Table~\ref{table:attn-covid} and \ref{table:attn-aq36}. (2) the 
negative transfer effect, which might happen when negative correlation exists between the time series of adjacent node pairs. Under such circumstances, directly applying multi-task learning framework in \vae\ might hurt the performance of NTS imputation. A possible solution is to resort to GNNs designed for graphs with heterophily~\cite{chien2020adaptive,zhu2021graph,lim2021large,zheng2022graph} in the decoder of the proposed \vae.
% One limitation of current implementation of \vae\ for NTS imputation is that it cannot handle larger graphs that contain thousands of nodes with tens of thousands of time steps. There are several ways to improve the scalability of \vae\ for NTS imputation, such as using AIStore~\cite{aizman2019high} for full-bandwidth disk-to-GPU data delivery, or using the sub-graph/node/edge sampling strategies from libraries such as PyTorch-Geometric~\cite{fey2019fast} and Deep Graph Library~\cite{wang2019deep}, etc.
%Regarding potential future directions for NTS imputation, 
There are several interesting aspects that are worth future study, including (1) generalizing the proposed \vae\ for detecting anomalies on NTS, forecasting NTS~\cite{wu2020connecting,jing2022retrieval} and assisting temporal graphs analysis~\cite{DBLP:conf/kdd/FuZH20,DBLP:conf/kdd/FuFMTH22}; (2) applying it to temporal knowledge graph completion~\cite{jin2020recurrent} and alignment~\cite{yan2021dynamic} as well as dynamic recommendations~\cite{li2020dynamic}.

\end{document}